\documentclass[preprint,authoryear,12pt]{elsarticle}
\usepackage{threeparttable}
\usepackage{filecontents,catchfile}
\usepackage{amsmath,amssymb,amsfonts}
\usepackage{algorithmic}
\usepackage{multirow}
\usepackage{multicol}
\usepackage{graphicx}
\usepackage{lscape} 
\usepackage[table,xcdraw]{xcolor} 
\usepackage{array}

\usepackage{float}
\usepackage{wrapfig}
\usepackage[draft]{todonotes}   % notes shown
\usepackage[labelfont=bf]{caption}
\usepackage{rotating}
\usepackage{caption}
\usepackage{mathtools} % Mathe

\usepackage{booktabs}
\usepackage{rotating}
\makeatletter
\newcommand*{\rom}[1]{\expandafter\@slowromancap\romannumeral #1@}
\makeatother
\usepackage[export]{adjustbox}

\usepackage{color}

\definecolor{azure}{rgb}{0.0, 0.5, 1.0}

\usepackage[draft]{todonotes}   % notes shown
\newcommand{\annotation}[1]{\marginpar{\small\itshape\color{azure}#1}} 
\usepackage{soul}
\newcommand{\hlc}[2][yellow]{{\sethlcolor{#1}\hl{#2}}}

\newcommand{\squeezeup}{\vspace{-2.5mm}}

\newcommand{\littlesqueezeup}{\vspace{-2.3mm}}

\definecolor{darkgreen}{rgb}{0.53, 0.66, 0.42}
\definecolor{grey}{rgb}{0.57, 0.64, 0.69}
\usepackage[colorlinks=true,linkcolor=blue,citecolor=blue]{hyperref} 

\journal{Medical Image Analysis}

\begin{document}

\begin{frontmatter}



%Multi-view Multi-layer
%\title{Fast and Accurate Feature Selection for Neurological Disorder Diagnosis by Estimating a Population-Based Representative Network Atlas of Functional Brain Connectivity}

\title{Brain Multigraph Prediction using Topology-Aware Adversarial Graph Neural Network}

%\author[BASIRA]{Anna Lisowska}
\author{Alaa Bessadok\fnref{BASIRA,ISITCOM,ENISO}}
\author{Mohamed Ali Mahjoub\fnref{ISITCOM,ENISO}}
\author{Islem Rekik\corref{cor}\fnref{BASIRA,DUNDEE}}

\address[BASIRA]{BASIRA lab, Faculty of Computer and Informatics, Istanbul Technical University, Istanbul, Turkey \ }
\address[ISITCOM]{Higher Institute of Informatics and Communication Technologies, University of Sousse, Tunisia \ }
\address[ENISO]{National Engineering School of Sousse, University of Sousse, LATIS- Laboratory of Advanced Technology and Intelligent Systems, 4023, Sousse, Tunisia \ }
\address[DUNDEE]{School of Science and Engineering, Computing, University of Dundee, UK \ }

\cortext[cor]{Corresponding author; Dr Islem Rekik (irekik@itu.edu.tr), \url{http://basira-lab.com/}, GitHub code: \url{https://github.com/basiralab/topoGAN}. Supplementary materials are available in this file.}


\begin{abstract}

Brain graphs (i.e, connectomes) constructed from medical scans such as magnetic resonance imaging (MRI) have become increasingly important tools to characterize the abnormal changes in the human brain. Due to the high acquisition cost and processing time of multimodal MRI, existing deep learning frameworks based on Generative Adversarial Network (GAN) focused on predicting the missing multimodal medical images from a few existing modalities. While brain graphs help better understand how a particular disorder can change the connectional facets of the brain, synthesizing a target brain \emph{multigraph} (i.e, multiple brain graphs) from a single source brain graph is strikingly lacking. Additionally, existing graph generation works mainly learn one model for each target domain which limits their scalability in \emph{jointly} predicting multiple target domains. Besides, while they consider the global topological scale of a graph (i.e., graph connectivity structure), they overlook the local topology at the node scale (e.g., how central a node is in the graph). %This means that the anatomical topology can be shifted in the predicted target graphs. 
To address these limitations, we introduce topology-aware graph GAN architecture (topoGAN), which {jointly} predicts multiple brain graphs from a single brain graph while preserving the topological structure of each target graph. Its three key innovations are: (i) designing a novel graph adversarial auto-encoder for predicting multiple brain graphs from a single one, (ii) clustering the encoded source graphs in order to handle the mode collapse issue of GAN and proposing a \emph{cluster-specific decoder}, (iii) introducing a \emph{topological loss} to force the prediction of topologically sound target brain graphs. The experimental results using five target domains demonstrated the outperformance of our method in brain multigraph prediction from a single graph in comparison with baseline approaches.

\end{abstract}


\end{frontmatter}

%\linenumbers

%%***************************************************************************** %%
\section{Introduction}
%%***************************************************************************** %%

{Multimodal} neuroimaging data such as magnetic resonance imaging (MRI) and positron emission tomography (PET) provides complementary information for diagnosing neurological disorders. {Nevertheless such data is not conventionally} acquired for clinical diagnosis. Therefore, predicting modalities from minimal resources becomes a fundamental task in the neuroscience field. Existing deep learning methods aiming to solve this problem can be categorized into \emph{one-target} (i.e, one-to-one) and \emph{multi-target} (i.e, one-to-many) prediction approaches. For instance, in the first category, \citep{Zeng:2019} proposed a framework based on Generative Adversarial Network (GAN) \citep{Goodfellow:2014} to predict Computed Tomography (CT) images from MRI where a cyclic reconstruction loss were introduced to improve the synthesis task. Similarly, \citep{Pan:2019} adopted the cyclic loss proposed in \citep{Zhu:2017} to predict PET from MR images for an early Alzheimer's Disease identification. While these works predicted the target image using a single source modality, \citep{Li:2019} introduced DiamondGAN, a multi-modal GAN-based framework to predict double inversion recovery (DIR) scan from three source modalities (i.e, Flair, T1 and T2). A potential limitation of such \emph{one-target} prediction frameworks, is that they are incapable of jointly predicting multiple target modalities in a single learning model (\textbf{Fig.}~\ref{1}-A). 

To this end, several attempts were made in the literature which are embedded into the second category that is \emph{multi-target} prediction frameworks. For example, \citep{Huang:2019} designed an autoencoder adversarially regularized by a discriminator to predict three target MR images (i.e, T1-weighted, T2-weighted, and FLAIR) from a single source T1 MRI scan. While it is the single \emph{multi-target} prediction work we identified in the medical imaging field, many frameworks were designed for computer vision tasks. Recently, \citep{Wu:2019} proposed a GAN-based model where the image synthesis step is first conditioned by a relative attribute vector representing the desired target domain, second it is adversarially regularized using three discriminators. To further improve the quality of the synthetic images, \citep{Cao:2019} exploited the correlation existing across multiple target domains by proposing a Wasserstein GAN-based framework. However, all the aforementioned models belonging to both categories were designed for synthesizing images, which limits their generalization to geometric data types such as graphs and manifolds \citep{Bronstein:2017}. In particular, predicting brain graphs (i.e, connectomes), which models the functional, structural, or morphological interactions between brain regions is of paramount importance for charting the brain dysconnectivity patterns \citep{van:2019,Bassett:2017}. 

%For instance, a recent work \citep{Mhiri:2020} discovered five connectional biomarkers for discriminating between autistic subjects and non disordered ones. Interestingly, some of these biomarkers are connecting a brain region from the right hemisphere and another one from the left hemisphere which means if one of these regions is affected by the disorder it will automatically affect the other one. Such discovery was not derived from neuroimaging data such as MRI. Generally, predicted discriminative biomarkers from a medical scan are only brain regions and not a connection between two brain regions.

{A brain graph is an undirected graph conventionally encoded in a symmetric connectivity matrix where each element (i.e, edge connecting two nodes) measures the connectivity strength between pairs of region of interest (ROIs). Leveraging such brain data representation for the purpose of neurological disorder diagnosis {can eventually improve prognosis \citep{Fornito:2015}. By reason of its importance for understanding normal brain function and disordered brain dysfunction, such brain representation have been used for many purposes such as brain graph integration \citep{mhiri2020supervised, yang2020unified, bessadok2018intact, gurbuz2020deep}, disease early detection \citep{song2020integrating, li2020pooling}, developmental trajectories prediction \citep{Ghribi:2019, goktas2020residual, nebli2020deep}. These studies demonstrate that leveraging different types of brain connectivities such as functional, structural and morphological ones provide more accurate results compared to neuroimaging since brain graphs represent a comprehensive mapping of neural activities.} Thus, we highly need the connectomic data types for early diagnosis of neurological diseases. However, constructing brain graphs is limited by (i) the incompleteness of existing multimodal medical datasets and (ii) the pre-processing pipeline including different steps such as cortical parcellation and the surface registration is time-consuming for a single raw MRI (e.g, T1- and T2-weighted scans) \citep{Li:2013}. Hence, these challenges dictate priorities for brain graph prediction. Especially, predicting missing target brain graphs from an existing source graph is highly required for learning the holistic brain mapping in healthy and disorder cases. In this regard, several recent studies were proposed for predicting brain graphs \citep{Bessadok:2019a, Bessadok:2019b, lgdada}. However, to predict a target brain multigraph from a single source graph using these frameworks, we need to train the model for each target brain graph independently. Thus, such one-to-one prediction frameworks have a limited robustness. Consequently, we propose topoGAN, the first geometric deep learning framework aiming to jointly predict multiple brain graphs from a single graph in an end-to-end learning architecture (\textbf{Fig.}~\ref{1}-B). We root our framework in the recently designed adversarial autoencoder model \citep{Cao:2019}, which is a multi-domain translation technique primarily designed for images. Although effective, \citep{Cao:2019} has two major limitations: (i) it fails to operate on graphs as it was primarily designed for Euclidean data, (ii) it overlooks GAN mode collapse, where the generator (i.e., decoder) produces data that mimic a few modes of the target domain. To this end, we first propose to cluster the source graphs into homogeneous groups, which helps disentangle heterogeneous source data distributions. Second, we include the topological measurements (e.g, closeness centrality) into the adversarial learning which aligns the global and local graph topology of the predicted target graphs with that of the ground-truth ones. Fundamentally, we summarize the main contributions of this paper as follows:}

\begin{enumerate}
\item \emph{Source brain graph embedding clustering.} We learn the source graph embedding using an encoder $E$ defined as a Graph Convolutional Network (GCN) \citep{Kipf:2016}. Second, we cluster the resulting embeddings of the whole training population with heterogeneous distribution into homogeneous clusters. In that way, we enforce our \emph{multi-target} prediction model to circumvent the mode collapse issue of GAN-based works \citep{Goodfellow:2014,Cao:2019}.

\item \emph{Cluster-specific multi-target graph prediction.} Given the clustered source graphs embeddings, we define a set of synergetic generators for each target domain, each representing a \emph{cluster-specific} GCN decoder. Hence, the graph prediction is learned more synergistically using our proposed cluster-specific generators, rather than using a single generator for each target domain. This generative process is regularized using one discriminator, which enforces the generated graphs to match the original target graphs.  

\item \emph{Topology-aware adversarial loss function.} In order to preserve both global and local topological properties of the original graphs, we unprecedentedly introduce a topological loss function which enforces the generated graphs to retain a centrality score of each nodes in the original target brain graph. 
\end{enumerate}

{Note that a preliminary version of this work was published in \citep{bessadok:2020}. This journal version presents the following extensions. (1) We removed the graph reconstruction loss in the adversarial loss of the generator since it had a low affect in the prediction accuracy of the graphs. (2) We carried out more experiments to show the effectiveness of our method compared with the benchmark methods. Specifically, we report the results of jointly predicting multiple brain graphs using six different source graphs (i,e. views).  (3) We further compared our framework to its variant architectures including graph attention network (GAT) and graph convolutional network (GCN). (4) We added more topology-focused evaluation metrics such as mean absolute error (i,e. MAE) between the real and predicted graphs, MAE between the real and predicted PageRank centrality, effective size and clustering coefficient of the graphs. We further reported $p$-value results using two-tailed paired $t$-test and the Kullback-Leibler divergence  between the real and predicted scores. (5) Finally, we added a visual comparison of the real and predicted graphs where we display the residual of the predicted multigraphs.}
%We also compared it to a recent state-of-the-art method \citep{bessadok:2020}.

%%***************************************************************************** %%
\section{Related work}
\begin{figure}[!htpb]
\centering
\includegraphics[width=10cm]{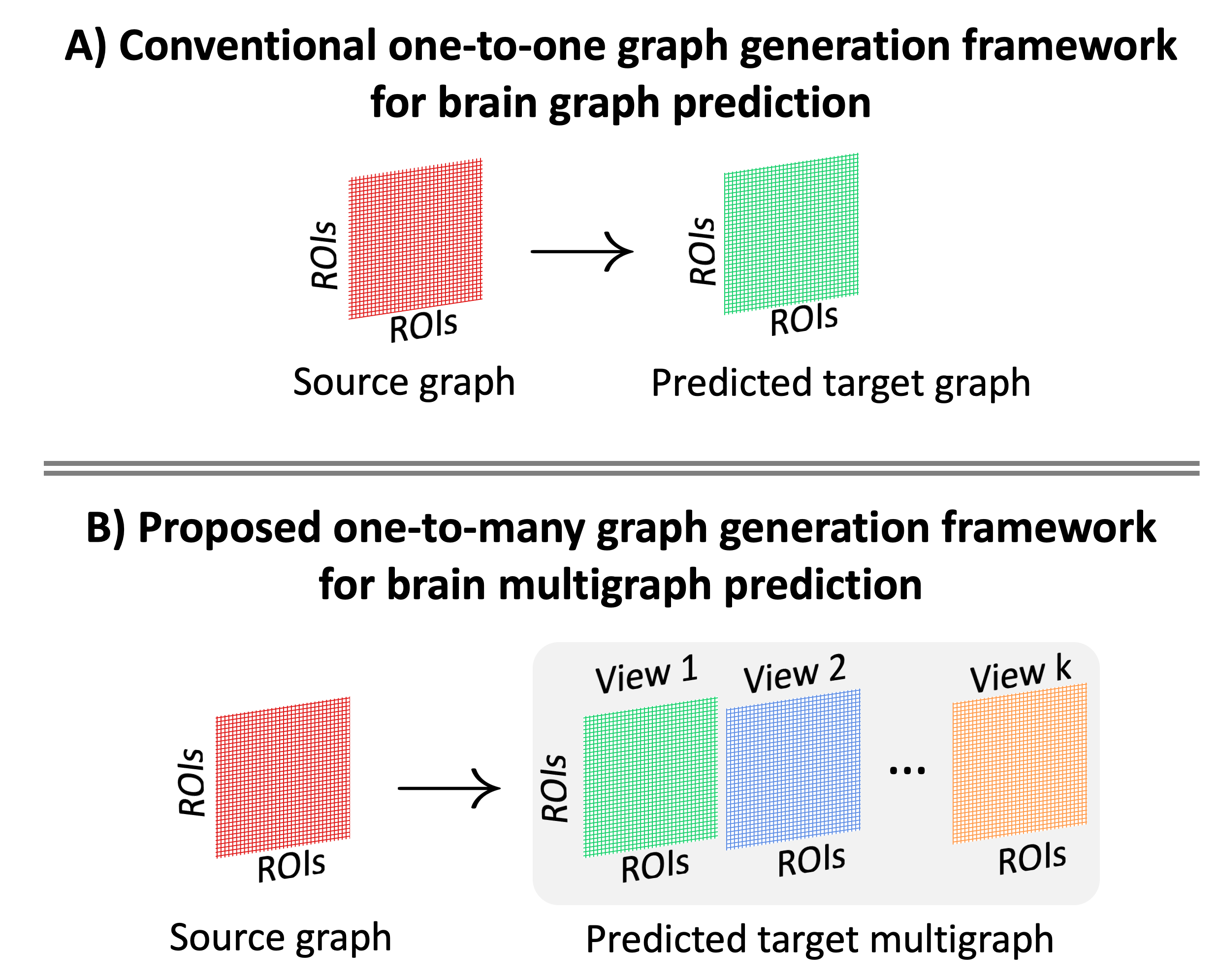}
\caption{\emph{Conventional brain graph prediction methods and the proposed brain multigraph prediction architecture.} \textbf{A)} In this illustration, we show the \emph{one-to-one} strategy adopted in recent brain graph synthesis works aiming to predict a target graph from a source one. Unfortunately, such strategies fail to simultaneously predict multiple target graphs from a single one, which limits their scalability to brain \emph{multigraph} derived from one or more magnetic resonance imaging (MRI) modalities (e.g., T1-weighted or resting-state functional MRI). \textbf{B)} To fill this gap, we propose a \emph{one-to-many} learning architecture aiming to predict a target brain multigraph from a source graph.}
\label{1}
\end{figure}

%%***************************************************************************** %%
{\textbf{Brain Graph synthesis.} A few recent papers have investigated  geometric deep learning methods for brain graph prediction  \citep{Bessadok:2019a,Bessadok:2019b,sserwadda2020,zhang2020} where in the first two works the synthesis task was partially formalized as a domain adaptation problem. }For example, \citep{Bessadok:2019a} symmetrically aligned the training source and target brain graphs and adversarially regularized the training and testing source graphs embeddings using two discriminators. The second geometric deep learning \citep{hamilton:2017} work, namely HADA \citep{Bessadok:2019b}, hierarchically aligned each source graphs to the target graphs of training subjects and optimized the whole framework using a single discriminator. Next, to predict the target brain graph of a representative subject both works averaged the target graphs of the training subjects that share similar local neighborhoods across source and target domains. Although promising, these works are not designed in an \emph{end-to-end} learning fashion (\textbf{Fig.}~\ref{1}-A). They mainly dichotomize the model into separate parts that do not co-learn which leads to relatively high accumulated errors across the learning steps. {To overcome this limitation, \citep{sserwadda2020} and \citep{zhang2020} designed end-to-end GAN-based frameworks for brain graph synthesis. In particular, \citep{sserwadda2020} adopted a cycle-consistency loss function to accurately perform the bidirectional mapping between the source and target brain graphs. Additionally, a new topological constraint was proposed to enforce the connectivity strength of the brain regions in the predicted graph to be similar to those of the ground-truth graph. On the other hand, \citep{zhang2020} parallelised multiple GCN models and fused the resulting learned representations to predict a target brain graph from a source one. Furthermore, a structure-preserving loss function was proposed to stabilize the training of both generator and discriminator models.} A shared shortcoming of all these brain graph synthesis works lies in their limited scalability for jointly predicting target brain \emph{multigraph} from a single source graph. Regarded as a holistic representation of brain connectivities, a \emph{multigraph} is a set of brain graphs stacked in a tensor where each captures a particular type of interactions between brain regions. Such brain representation plays an important role in modeling the dysfunctions in connectivity patterns existing between brain regions \citep{van:2019}. Thus, to predict a target brain \emph{multigraph} using the existing frameworks we should learn one model for each target domain. Such frameworks, however, have a limited robustness in predicting more than one target domain. {To solve this issue, our MICCAI 2020 conference paper \citep{bessadok:2020} presents the ﬁrst work that jointly predicts a set of target brain graphs (i,e. target brain \emph{multigraph}) from a single source graph. However, its experiments were restricted to only use a single source view to predict the target brain \emph{multigraph}. Even such experiments demonstrated the superiority of our model over the comparison methods, we aim in this work to further show its prediction performance using five additional source views from both hemispheres to the one we used in the conference paper.}

%Mainly, several studies approved that Autism is a heterogeneous neurological disorder which provoke hemisphere-specific abnormality changes. For example, \citep{zhao2020diagnosis} showed that  brain regions affected in the left hemisphere  are totally different from those affected in the right hemisphere. Thus, we aim to predict the \emph{multigraph} from different views of both hemispheres.}
 
\textbf{Graph generation.} Besides, plenty of efforts have been dedicated to synthesizing different types of graphs and have shown remarkable results in various applications such as road network generation \citep{belli:2019}, scene graph generation \citep{yang:2018} and biological molecules synthesis \citep{mittongraph}. Other studies \citep{Su:2019,Liao:2019} proposed to sequentially generate subgraphs consisting in a subset of nodes and their connectivities in order to generate the whole graph. Some other studies \citep{Bresson:2019,Flam:2020} proposed graph autoencoder frameworks where the encoded graph structure is decoded using a set of decoders. Recent works adopted a GAN-based solution combined with a reinforcement learning approach where an additional network is designed to further optimize the graph synthesis task \citep{de:2018,you:2018}. Despite their ineffectiveness in \emph{multi-target} prediction tasks, such graph synthesis studies fail to preserve the node-wise topological properties of the target domain. Specifically, they only learn the \emph{global graph structure} (i.e., number of nodes and edges weights). However, the brain wiring has both global and local topological properties which makes the ebb and flow of brain activity acts like the fingerprint of a subject \citep{Fornito:2015}. In fact, neurological disorders such as Alzheimer's disease alter the brain cortex and more importantly it atrophies its ROIs in varying degrees. Hence, focusing only on learning the global graph properties and overlooking the \emph{local graph structure} may undervalue the role of specific ROIs in early diagnosing the disease. Therefore, devising a scalable and accurate framework for predicting a target brain multigraph from a single source graph that preserves the global and local topological properties of the original brain graphs is of great interest \citep{Zhang:2020a,Zhang:2020}. Such local topology awareness can be defined by learning the node's influence in the graph measured using path-length based metric which is defined in graph theory as ``centrality". Several network metrics such as betweenness centrality, closeness centrality and PageRank centrality have been introduced in the graph theory literature. For instance, \citep{Huang:2019a} proposed an analysis study for Parkinson’s disease (PD) where many centrality measurements were leveraged to identify the potential biomarkers in the disease progression using functional brain graphs extracted from resting-state functional MRI data (rs-fMRI). In a follow-up work, \citep{Cid:2019} used five centrality metrics to create a graph-based subject profile for a Pulmonary tuberculosis (TB) classification purpose. To the best of our knowledge, up to now no existing works have investigated the learning of node centrality for brain graph prediction \citep{Zhou:2018}. %For instance, \citep{Yang:2020} leveraged closeness and betweenness centrality metrics to model the characteristic of directed graphs for a human action recognition task. 

% %% ***************************************************************************** %%
%%% SMALL FIGURE %%%
% %% ***************************************************************************** %%

% %% ***************************************************************************** %%
%%% MAIN FIGURE %%%
% %% ***************************************************************************** %%
\clearpage
\begin{sidewaysfigure}[ht]
\vspace{-50pt}
\includegraphics[width=\textwidth]{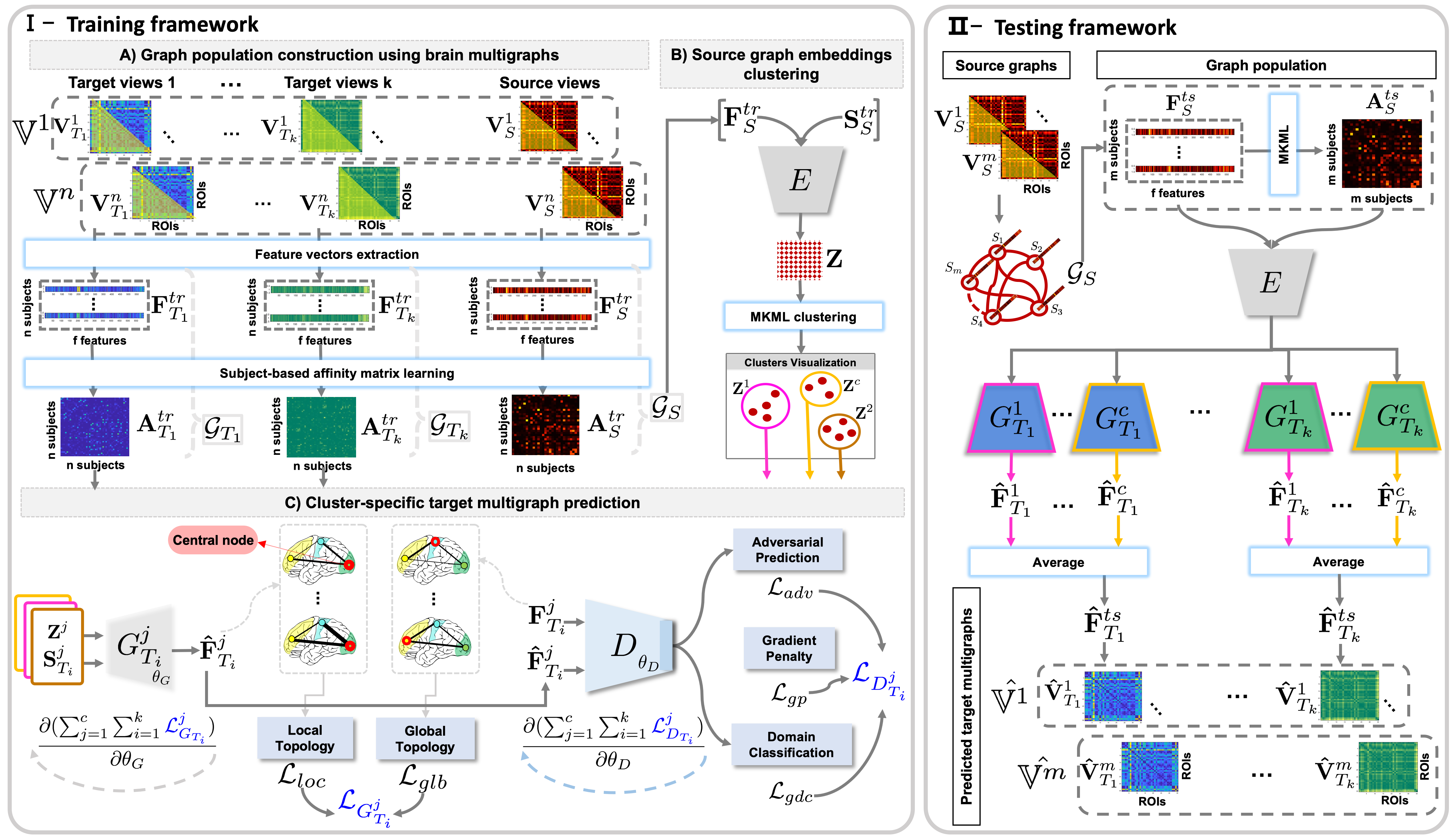}
\caption{\emph{Outline of the proposed topology-aware graph Generative Adversarial Network (topoGAN) for predicting a target brain multigraph from a single source graph.} \textbf{\rom{1}- Training pipeline.} \emph{(A) Graph population construction using brain multigraph.} For each training subject, we extract the source and $k$ target feature vectors from their brain graphs. Then, we use multiple kernel manifold learning to learn the affinity between subjects. Thus, we construct a graph population for a specific domain using the resulting extracted features and the learned affinity matrix. \emph{(B) Source graph embeddings clustering.} We learn the graph embedding of the source population and cluster it into $c$ groups. \emph{(C) Cluster-specific target multigraph prediction.} We train $c$ \emph{ cluster-specific generators} for each target domain which are adversarially regularized by a discriminator $D$ aiming to enforce the target domain classification. We further regularize each generator ${G}_{{T}_{i}}^{j}$ by proposing a local and a global topology losses which preserves the local node topology and the global graph connectivity structure, respectively. \textbf{\rom{2}-Testing pipeline.} We construct a graph population capturing the affinity between testing subjects using their brain features. Next, we learn the graph embedding of the source population and decode it using $c$ cluster-specific generators dedicated to each of the $k$ target domains. Then, we predict the testing brain multigraph's frontal views by averaging the graphs produced by our $c$ generators in a specific target domain.}
\label{2}
\end{sidewaysfigure}
\clearpage

% %% ***************************************************************************** %%
%%% Method %%%
% %% ***************************************************************************** %%
\section{Proposed Method} 

\textbf{Problem Definition.} A brain graph is defined as \( \mathcal{B}=\{ \mathcal{B}^{r}, \mathcal{B}^{e}, \mathbf{V} \}\) where \(\mathcal{B}^{r}\) is a set of nodes (i.e, ROIs) and \(\mathcal{B}^{e}\) is a set of weighted edges encoding the biological connectivity between nodes. Each training subject \(s\) in our dataset is represented by a brain \emph{multigraph} which captures the functional, structural, or morphological connectivities between brain regions. More compactly, it can be written as a tensor $\mathbb{V}^{s}$ in $\mathbb{R}^{r\times r\times v}$ where its tensor layers denote a set of multi-view brain graphs $\{ \mathbf{V}_{i}^s \}_{i=0}^v$, each view capturing a particular type of interactions between brain regions (e.g, morphological, functional and structural). \(\mathbf{V} \in \mathbb{R}^{r \times r} \) denotes the connectivity matrix measuring the pairwise edge weight between nodes using a particular view, where \({r}\) is the number of ROIs. Given a testing subject $s'$ represented by a single source graph $\mathbf{V}_{S}^{s'}$, our objective is to predict its missing target brain \emph{multigraph} $\mathbb{V}^{s'}$ in $\mathbb{R}^{r\times r\times k}$ where $k = v -1$ and its frontal views are denoted by $\{ \mathbf{{V}}_{T_{i}}^{s'} \}_{i=1}^k$.

\textbf{Fig.}~\ref{2} provides an overview of our proposed method and \textbf{Table}~\ref{tab1} summarizes the major mathematical notations we used in this paper. There are three major steps in the pipeline: 1) extraction of multi-view brain features from source and target graphs and construction of a graph population for each domain, 2) embedding and clustering of the source graphs, and 3) prediction of the target brain multigraph using cluster-specific generators.

% %% ***************************************************************************** %%
%%% Section A %%%
% %% ***************************************************************************** %%
\subsection{Graph population representation using multi-view brain graphs}
%By taking into account the similarities between samples, we are improving the graph prediction step.
To map the source brain graph of a subject to its target brain multigraph, we need to model the relationship between training samples using their connectivity features (e.g, weights). In fact, we hypothesize that samples (i.e, brain graphs) with strong affinity in the source domain will also maintain such a strong affinity in the target domains to some extent. Hence, we propose to learn the source graph embedding of each training subject using an encoder $E$ defined as a Graph Convolutional Network (GCN) \citep{Kipf:2016}. Thus, we create in this step for each of the source and target domains a graph population encoded in an affinity matrix where nodes denote subjects represented by their brain features and the edges represent the pairwise affinity between subjects. Since each brain graph is encoded in a symmetric matrix, we propose to vectorize the off-diagonal upper-triangular part which helps eliminate redundancy in the brain graphs. Then, we vertically stack the extracted feature vectors of size $f$ for $n$ training subjects which results in a feature matrix $\mathbf{F}_{v}^{tr}$ in $\mathbb{R}^{n\times f}$ where $v\in\{{S},{T}_{1},\dots,{T}_{k}\}$. Next, given these matrices we create a subject-based affinity matrix $\mathbf{A}_{v}^{tr}$ in $\mathbb{R}^{n\times n}$ by leveraging multi-kernel manifold learning (MKML) algorithm \citep{Wang:2017} which learns the affinity between training feature vectors. We choose MKML for its appealing aspect of learning multiple kernels to efficiently fit the true underlying statistical distribution of the data. Hence, for a specific view $v$, we define our graph population for a specific domain as $\mathcal{G}_{v}=\{\mathcal{G}^{n}_{v}, \mathcal{G}^{e}_{v}, \mathbf{F}_{v}\}$, $\mathcal{G}^{n}_{v}$ denotes a set of nodes (i.e, subjects), $\mathcal{G}^{e}_{v}$ a set of weighted edges encoding the affinity between subjects and $\mathbf{F}_{v}$ denotes a feature matrix. We create a set of graph populations $\{ \mathcal{G}_{i}^{tr} \}_{i=0}^v$ for the source and target domains where each is represented by a set of feature matrices $\{ \mathbf{F}_{i}^{tr} \}_{i=0}^v$ and a set of learned affinity matrices $\{ \mathbf{A}_{i}^{tr} \}_{i=0}^v$ for the training subjects (\textbf{Fig.}~\ref{2}--A). The resulting graph populations will be used in the two following steps to learn the graph embeddings of the source population (i.e, step B) and to map the source graph to the target multigraph of a training subject (i.e, step C).

% %% ***************************************************************************** %%
%%% Table %%%
% %% ***************************************************************************** %%
\begin{table}
\captionsetup{justification=centering}
\caption{Major mathematical notations used in this paper}
\centering
\begin{scriptsize}
\begin{tabular}{ >{\centering\arraybackslash}m{1in} >{\centering\arraybackslash}m{0.5in} >{\centering\arraybackslash}m{3.8in} }
	\toprule
	Notation& 
	Dimension& 
	Definition \\
	\toprule
	$r$ & $\mathbb{N}$ & number of brain regions (i.e, ROIs) \\
	$v$ & $\mathbb{N}$ & number of brain views (i.e, source and target views) \\
	$k$ & $\mathbb{N}$ & number of target views (i.e, $k = v -1$) \\
	$s$ & $\mathbb{N}$ & total number of subjects including training and testing ones\\
	$n$ & $\mathbb{N}$ & number of training subjects \\
	$m$ & $\mathbb{N}$ & number of testing subjects \\
	$f$ & $\mathbb{N}$ & number of features extracted from the original brain graph\\
	$d$ & $\mathbb{N}$ & number of features in the embedded graphs\\
	$c$ & $\mathbb{N}$ & number of clusters \\
	$e$ & $\mathbb{N}$ & number of edges in a graph population \\
	$e^{\prime}$ & $\mathbb{N}$ & number of edges in a brain graph \\
	$\mathcal{B}=\{ \mathcal{B}^{r}, \mathcal{B}^{e}, \mathbf{V} \}$ & $-$ & a brain graph of a specific subject \\
	$\mathcal{B}^{r}$ & $\mathbb{N}^{1 \times r}$ & a set of anatomical brain regions representing nodes in the brain graph \\
	$\mathcal{B}^{e}$ & $\mathbb{N}^{1 \times e^{\prime}}$ & a set of edges connecting the brain regions representing either functional, structural or morphological connectivities in the brain. \\
	$\mathbf{V}$ & $\mathbb{R}^{r \times r}$ &  connectivity matrix measuring the pairwise edge weights between nodes (i.e, ROIs)\\
	$\mathbf{V_{i}}$ & $\mathbb{R}^{r \times r}$ & brain graph constructed from a single view $i$ where $i \in \{0,\dots,v\}$ \\
	$\mathbf{\hat{V}}_{i}$ & $\mathbb{R}^{n \times n}$ & predicted brain graph for a specific target domain $i$ where $i \in \{0,\dots,k\}$ \\
	$\mathbb{V}^{s}$ & $\mathbb{R}^{r \times r \times v}$ & a brain multigraph tensor of a training subject $s$ stacking a set of source and target brain graphs\\
	$\hat{\mathbb{V}}^{s'}_{T}$ & $\mathbb{R}^{r \times r \times k}$ & predicted target brain multigraph tensor of a testing subject $s'$ stacking a set of target brain graphs \\
	$\mathcal{G}_{v}=\{\mathcal{G}^{n}_{v}, \mathcal{G}^{e}_{v}, \mathbf{F}_{v}\}$ & $-$ & graph population representing the similarity between subjects belonging to a population \\
	$\mathcal{G}^{n}_{v}$ & $\mathbb{N}^{1 \times s}$ & a set of graph population nodes or subjects in a population \\
	$\mathcal{G}^{e}_{v}$ & $\mathbb{N}^{1 \times e}$  & a set of edges connecting pairs of subjects representing the similarity between them based on their brain graphs\\
	$\mathbf{F}_{v}$ & $\mathbb{R}^{n \times s}$ & feature matrix vertically stacking feature vectors extracted from the view $v$ of $s$ subjects belonging to a specific population\\
	$\mathbf{F}_{v}^{tr}$ & $\mathbb{R}^{n \times f}$ & feature matrix vertically stacking feature vectors extracted from the view $v$ of $n$ training subjects \\
	$\mathbf{A}_{v}^{tr}$ & $\mathbb{R}^{n \times n}$ & subject-based affinity matrix between $n$ training subjects using the brain feature vectors belonging to the view $v$ \\
	$\mathbf{Z}$ & $\mathbb{R}^{n \times d}$ & learned source graph embeddings of the training subjects \\
	$(\mathbf{X}_{{T}_{i}}^{j})^{\mathcal{C}}$ & $\mathbb{R}^{n \times r}$ & centrality matrix of the real target graphs in the domain ${T}_{i}$ and the cluster $j$ of the training subjects computed using the centrality metric ${\mathcal{C}}$\\
	$(\hat{\mathbf{X}}_{{T}_{i}}^{j})^{\mathcal{C}}$ & $\mathbb{R}^{n \times r}$ & centrality matrix of the predicted target graphs in the domain ${T}_{i}$ and the cluster $j$ of the training subjects computed using the centrality metric ${\mathcal{C}}$\\
\bottomrule
\end{tabular}
\end{scriptsize}
\label{tab1}
\end{table}

% %% ***************************************************************************** %%
%%% Section B %%%
% %% ***************************************************************************** %%
\subsection{Source brain graphs embedding and clustering} 

The proposed topoGAN is a graph autoencoder comprising an encoder $E$ and a set of domain-specific decoders (i.e, generators) $\{G_{T_{i}}\}_{i=1}^k$ (i,e, a decoder for each of the $k$ target domains) regularized by a discriminator $D$. Considering such an adversarial learning-based framework causes the mode collapse problem where the generator produces very limited number of modes \citep{Goodfellow:2014}. In such case, no matter how big is our training set in terms of number of subjects because only few of them contribute to the graph synthesis. Thus, all generated graphs will look similar. We propose to solve this problem by clustering the source brain graphs which naturally have a heterogeneous statistical distribution. Since clustering samples encoded in high-dimensional feature vectors might be a complex task, we first propose to map the source brain graph into a low-dimensional space which helps reduce its dimensionality while preserving its topological structure. Consequently, the mode collapse issue is handled in two consecutive steps:

${\mathbf{(1)}}$ we first learn the population graph embedding in the source domain which maps each subject-specific features into a lower representative dimensional space. To this end, we use an encoder $E(\mathbf{F}_{S}^{tr},\mathbf{A}_{S}^{tr})$ defined as a GCN with two layers inputing the source feature matrix $\mathbf{F}_{S}^{tr}$ of the training subjects and the learned subject-based affinity matrix $\mathbf{A}_{S}^{tr}$. GCN \citep{Kipf:2016} is originally defined using convolutions in the spectral domain which is expressed by the following convolution function: 
\begin{gather}
    {f}_{\phi}(\mathbf{F}^{(l)}, \mathbf{A}_{S}^{tr} \vert \mathbf{W}^{(l)}) = {\phi}(\mathbf{\widetilde{D}}^{-\frac{1}{2}}\mathbf{\widetilde{\mathbf{A}}}_{S}^{tr}\mathbf{\widetilde{D}}^{-\frac{1}{2}}\mathbf{F}^{(l)}\mathbf{W}^{(l)})
\label{eq:1}
\end{gather}

$\phi$ denotes the $ReLU$ and $linear$ activation functions we used in the first and second layers, respectively. We define $\mathbf{F}$ in the first layer as the source feature matrix $\mathbf{F}_{S}^{tr}$ while we define it in the second layer as the resulting embeddings learned from the first layer $l$. $\mathbf{W}^{(l)}$ is a filter used to learn the convolution in the GCN in each layer $l$. $\mathbf{\widetilde{D}}_{ii} = \sum_{j}\mathbf{\widetilde{\mathbf{A}_{{S}}}}(ij)$ is a diagonal matrix and $\widetilde{\mathbf{A}}_{S}^{tr} = \mathbf{A}_{S}^{tr} + \mathbf{I}$ with $\mathbf{I}$ being an identity matrix used for regularization. Ultimately, we define the layers of our GCN encoder as follows:
\begin{gather}
	    \mathbf{Z}^{(1)} = f_{ReLU}(\mathbf{F}_{S}^{tr}, \mathbf{A}_{S}^{tr} \vert \mathbf{W}^{(0)}); \mathbf{Z}^{(2)} = f_{linear}(\mathbf{Z}^{(1)}, \mathbf{A}_{S}^{tr} \vert \mathbf{W}^{(1)})
\label{eq:2}
\end{gather}

${\mathbf{(2)}}$ We cluster the resulting source embeddings $\mathbf{Z}$ into homogeneous groups which helps disentangle the heterogeneous distribution thereby reducing the generator's risk to match a few unimodal samples of the target domain. To do so, we leverage MKML since it outperformed PCA \citep{Jolliffe:2016} and t-SNE \citep{Maaten:2008} clustering methods when dealing with biological dataset \citep{Wang:2017}. More importantly, it is widely used for brain graph analysis tasks and showed promising results in neaurological disorder diagnosis such as Autism spectrum disorder (ASD) \citep{Soussia:2018b,Bessadok:2019a,Mhiri:2020}. Specifically, it first produces a pairwise affinity matrix measuring the similarity between training subjects using their source graph embeddings $\mathbf{Z}$ and having $c$ diagonal blocks denoting the clusters. Next, the obtained source affinity matrix is projected into a lower dimension using t-SNE \citep{Maaten:2008} which results in a latent matrix in $\mathbb{R}^{n \times c}$. Last, k-means algorithm \citep{Jain:2010} is used to cluster the subjects into $c$ clusters based on the resulting learned latent matrix (\textbf{Fig.}~\ref{2}-B).

\subsection{Cluster-specific multi-target graph prediction} 

For each target view $k$, we design a set of \emph{cluster-specific generators} $\{G_{T_{i}}^{j}\}_{i=1,j=1}^{k,c}$, each learning to match the distribution of a cluster $j$ belonging to a target domain ${T_{i}}$ (\textbf{Fig.}~\ref{2}-C). Our goal is to enforce each generator to learn from all examples in the cluster $c$ thereby avoiding the mode collapse issue of GAN-based models. We define our generators as GCN decoders with similar architecture to the encoder (\textbf{Eq.}~\eqref{eq:2}). Specifically, a generator ${G}_{{T}_{i}}^{j}$ designed to predict the target graphs of the domain ${T}_{i}$ and the cluster $j$ takes as input the learned source embeddings $\mathbf{Z}^{j}$ and the subject-based affinity matrix $\mathbf{A}_{{T}_{i}}^{j}$. Specifically, we learn the affinity between feature vectors of the target domain ${T}_{i}$ belonging to the same cluster $j$. In other words, decoding the source embeddings with the affinity matrix learned using the target graphs in ${T}_{i}$ enforces the generated graphs to approximate the real target domain structure of a specific cluster $j$. We further propose to optimize the target graph prediction using a discriminator $D$ which is a GCN with three layers aiming to enforce the generated target graph to approximate the ground-truth target distribution of a specific target domain. To achieve this, we propose three loss functions which optimize the discriminator $D$. 

\textbf{Adversarial loss.} We introduce an adversarial loss function differently from the vanilla GAN \citep{Goodfellow:2014} where we compute the Wasserstein distance among all domains in order to measure the realness of the generated graphs. This distance has been widely used in the GAN literature as it stabilizes the training process of the model thereby making it less sensitive to hyperparameter regularization \citep{gulrajani:2017}. Thus, we formulate it as follows:
\begin{gather}
 \mathcal{L}_{adv}^{j} = - \mathbb{E}_{\mathbf{F}^{\prime} \sim \mathbb{P}_{{\mathbf{F}}_{S}^{j}}} \ [ D(\mathbf{F}^{\prime}) \ ] + \frac{1}{k} \sum_{i=1}^{k} \mathbb{E}_{\mathbf{F}^{\prime\prime} \sim \mathbb{P}_{{\hat{\mathbf{F}}}_{{T}_{i}}^{j}} }\ [ D(\mathbf{F}^{\prime\prime}) \ ] 
\label{eq:3}
\end{gather}

where $\mathbb{P}_{{\mathbf{F}}_{S}^{j}}$ is the real source graph distribution of the cluster $j$, and the distribution $\mathbb{P}_{{\hat{\mathbf{F}}}_{{T}_{i}}^{j}}$ is the generated distribution by $G_{T_{i}}^{j}$ in the target domain $T_{i}$.

\textbf{Graph domain classification loss.} We recall that the goal of training the cluster-specific generators $\{G_{T_{i}}^{j}\}_{i=1,j=1}^{k,c}$ is to produce graphs for the cluster $j$ which are properly classified by the discriminator to the specific target domain $T_{i}$. Thus, we define a binary classifier $D_{C}$ on top of our discriminator $D$ which classifies the synthetic graphs $\hat{\mathbf{F}}_{{T}_{i}}^{j}$ as $0$ and the real target graphs $\mathbf{F}_{{T}_{i}}^{j}$ as $1$. In detail, the former is defined as:
\begin{gather}
\mathcal{L}_{gdc}^{j} = \sum_{i=1}^{k}  \mathbb{E}_{\mathbf{F}^{\prime\prime}\sim \mathbb{P}_{\hat{\mathbf{F}}_{{T}_{i}}^{j}} \cup  \mathbb{P}_{{\mathbf{F}}_{{T}_{i}}^{j}}} \ [ \ell_{MSE}(D_{C}(\mathbf{F}^{\prime\prime}),y({\mathbf{F}^{\prime\prime}})) \ ]
\label{eq:4}
\end{gather}

$ \ell_{MSE}$ is the mean squared loss. $D_{C}(\mathbf{F}^{\prime\prime})$ and  $y({\mathbf{F}^{\prime\prime}})$ denote the predicted and ground-truth labels corresponding to the graph $\mathbf{F}^{\prime\prime}$, respectively. More specifically, we compute the $\ell_{MSE}$ for the predicted and real graphs separately (i.e, ${\hat{\mathbf{F}}_{{T}_{i}}^{j}}, {{\mathbf{F}}_{{T}_{i}}^{j}}$) then we sum both values which we mathematically denote  by the following notation for simplicity $\mathbb{E}_{\mathbf{F}^{\prime\prime}\sim \mathbb{P}_{\hat{\mathbf{F}}_{{T}_{i}}^{j}} \cup  \mathbb{P}_{{\mathbf{F}}_{{T}_{i}}^{j}}}$.

\textbf{Gradient penalty loss.} To improve the training stability of our model, we adopt the gradient penalty loss used in \citep{Cao:2019} which is formulated as follows:

\begin{gather}
    \mathcal{L}_{gp}^{j} = (max \{ 0, {\mathbb{E}_{\tilde{\mathbf{F}} \sim \mathbb{P}_{\tilde{{\mathbf{F}}}^{j}_{k}}}} \|\nabla  D(\tilde{\mathbf{F}})\|- \sigma  \} )^{2}
\label{eq:5}
\end{gather}

$\tilde{\mathbf{F}}$ is sampled between the source graph distribution $\mathbb{P}_{{\mathbf{F}}_{S}^{j}}$ and the predicted target graph distribution $\mathbb{P}_{\hat{\mathbf{F}}_{k}^{j}}$ where ${\tilde{\mathbf{F}}_{k}^{j}}$ is a matrix stacking vertically the generated target graphs for all $k$ domains in the cluster $j$. Particularly, $\tilde{\mathbf{F}} \leftarrow \alpha{\mathbf{F}}_{S}^{j} + (1-\alpha)\tilde{\mathbf{F}}_{k}^{j}$ where $\alpha \sim U\ [0,1\ ]$ and $U$ is a uniform distribution. As suggested in \citep{Cao:2019}, we set the hyper-parameter $\sigma$ to $k$. 

Ultimately, the cost function of the discriminator which helps the generators of each cluster produce brain graphs each associated with a specific target domain is formulated as follows:
\begin{equation}
 	\mathcal{L}_{D} = \sum_{j=1}^{c} ( \mathcal{L}_{adv}^{j} + \lambda_{gp} \cdot \mathcal{L}_{gp}^{j} + \lambda_{gdc} \cdot \mathcal{L}_{gdc}^{j}  ),
\label{eq:6}
\end{equation}

$\lambda_{gdc}$ and $\lambda_{gp}$ are hyper-parameters to be tuned. Hence, by maximizing the discriminator loss function defined above \textbf{Eq.}~\eqref{eq:6} the generators are optimally trained to produce graphs that belong to a specific target domain.

Moreover, brain graphs have rich topological properties including their percolation threshold, hubness and modularity \citep{Bassett:2017}. Such unique properties should be preserved when synthesizing the target brain graphs \citep{Liu:2017,Joyce:2010}. Although regarded as an efficient graph embedding model, graph autoencoder is limited to only learning the \emph{global graph structure} such as number of nodes and edges in the graph while the \emph{local graph structure} should be also learned since it reflects the node importance in the graph. To this aim, we unprecedentedly introduce a topological loss function to guide the training process of the generators and constrains each of them to preserve the local nodes properties while learning the global graph structure (\textbf{Fig.}~\ref{2}-C). Specifically, we adopt three centrality metrics to compute a score for each ROI in the brain graph. We choose three centrality metrics widely used in graph theory \citep{Borgatti:2006}: closeness centrality, betweenness centrality and eigenvector centrality.

The closeness centrality $CC$ quantifying the closeness of a node to all other nodes \citep{Freeman:1977} is defined as follows:
\begin{equation}
    CC(r^{a}) = \frac{r-1}{\sum_{r^{a}\neq r^{b}} p_{r^{a}r^{b}}}
\label{eq:7}
\end{equation}
$r$ denotes the number of nodes (i.e, ROIs) and $p_{r^{a}r^{b}}$ is the length of the shortest path between nodes $r^{a}$ and $r^{b}$.
 
The betweenness centrality $BC$ measuring the number of shortest paths which pass across a node \citep{Beauchamp:1965} can be defined as:
\begin{equation}
    BC(r^{a}) = \frac{2}{(r-1)(r-2)}\times {\sum_{r^{a}\neq r^{b}\neq r^{c}}} \frac{P_{(r^{c},r^{b})} (r^{a})}{P_{(r^{c},r^{b})}} 
\label{eq:8}
\end{equation}
$P_{(r^{c},r^{b})}(r^{a})$ denotes the number of shortest paths between two nodes $r^{c}$ and $r^{b}$ that pass through $(r^{a})$.

The eigenvector centrality $EC$ capturing the centralities of a node's neighbors \citep{Bonacich:2007} is defined as follows:
\begin{equation}
    EC(r^{a}) = \mathbf{x}^{a} = \frac{1}{\lambda} \sum_{b=1}^{r}  \mathbf{V}_{a b}\mathbf{x}^{b} 
\label{eq:9}
\end{equation}
$\mathbf{V}_{a b}$ represents the $b$ neighbor of the node $a$, $\mathbf{x}^{a}$ and $\mathbf{x}^{b}$ are the eigenvectors resulting from the eigen decomposition of the adjacency matrix $\mathbf{V}$ and $\lambda$ is a positive proportionality factor.

Next, we compute $\ell_{MAE}(.)^{\mathcal{C}}$ the absolute difference between the real and predicted centrality scores of each node in the target graph which represents our \emph{local topology loss}. In particular, given a centrality metric $\mathcal{C}$ where $\mathcal{C}\in\{CC, BC, EC\}$, a cluster $j$ and a target domain ${T}_{i}$, we define $(\mathbf{X}_{{T}_{i}}^{j})^{\mathcal{C}}$ and $(\hat{\mathbf{X}}_{{T}_{i}}^{j})^{\mathcal{C}}$ both in $\mathbb{R}^{n\times r}$ as the centralities for the real brain graphs $\mathbf{V}_{{T}_{i}}^{j}$ and the generated ones $\hat{\mathbf{V}}_{{T}_{i}}^{j}$ reconstructed from the feature matrices $\mathbf{F}_{{T}_{i}}^{j}$ and $\hat{\mathbf{F}}_{{T}_{i}}^{j}$, respectively. To preserve the relationship between brain regions in terms of number of edges and their weights we compute the absolute difference between the real and predicted feature matrices $\mathbf{F}_{{T}_{i}}$ and $\hat{\mathbf{F}}_{{T}_{i}}$, which mainly represents our \emph{global topology loss} function. Ultimately, to take advantage of both local and global losses we propose to fuse them into a single function.

\textbf{Topological loss.} This is one of the key contributions for our proposed architecture which regularizes the cluster-specific generators (\textbf{Fig.}~\ref{2}-C). It is computed as follows:

\begin{equation}
 	\mathcal{L}_{top}^{j} (\mathcal{C}) = (\underbrace{\sum_{i=1}^{k} \ell_{MAE}(\mathbf{X}_{{T}_{i}}^{j}, \hat{\mathbf{X}}_{{T}_{i}}^{j})^{\mathcal{C}}}_{\textbf{local topology loss}} + \underbrace{\sum_{i=1}^{k} \ell_{MAE}(\mathbf{F}_{{T}_{i}}^{j}, \hat{\mathbf{F}}_{{T}_{i}}^{j})}_{\textbf{global topology loss}} )
\label{eq:10}
\end{equation}

\textbf{Information maximization loss.} Since our $k$ target domains are correlated we integrate the information maximization loss term to force the \emph{cluster-specific generators} $\{G_{T_{i}}^{j}\}_{i=1,j=1}^{k,c}$ to correlate the predicted graphs with a specific target domain ${T}_{i}$. As in \citep{Cao:2019}, we define it using this formula:
\begin{equation}
    \mathcal{L}_{inf}^{j} = \sum_{i=1}^{k} \ell_{BCE}(y=1,D_{C}(\hat{\mathbf{F}}_{{T}_{i}}^{j}))
\label{eq:11}
\end{equation}

$\ell_{BCE}$ is the binary cross entropy. Given the above definitions of the topological and information maximization losses, we introduce the overall \emph{topology-aware adversarial loss function} of each generator as:
\begin{multline}
 	\mathcal{L}_{G} = \sum_{j=1}^{c} ( - \frac{1}{k} \cdot \sum_{i=1}^{k} \mathbb{E}_{\mathbf{F}^{\prime\prime} \sim {\hat{\mathbf{F}}}_{{T}_{i}}^{j}} \ [ D(\mathbf{F}^{\prime\prime}) \ ] + \lambda_{top} \cdot \mathcal{L}_{top}^{j} (\mathcal{C}) + \lambda_{inf} \cdot \mathcal{L}_{inf}^{j})
\label{eq:12}
\end{multline}

where $\lambda_{top}$ and $\lambda_{inf}$ are hyper-parameters that control the relative importance of topological loss and information maximization losses, respectively. As illustrated in (\textbf{Fig.}~\ref{2}-\rom{2}), given a testing source graph we predict each view of the target brain multigraph by averaging the graphs of a specific target domain ${T}_{i}$ produced by the \emph{cluster-specific generators}.

\begin{landscape}
\begin{table}
\begin{threeparttable}
\hspace{-90pt}
\vspace{-50pt}
\centering
\captionsetup{justification=centering}
\caption{Average prediction results of a target brain multigraph from three source views derived from the left hemisphere using different evaluation metrics.} %}
\footnotesize
\begin{tabular}{|c|c|c|c|c|c|c|c|c|}
\hline
\rowcolor[HTML]{FFFFFF} 
\multicolumn{2}{|c|}{\cellcolor[HTML]{FFFFFF}{\color[HTML]{333333} \textbf{View 1}}}                              & \textbf{MAE}                           & \textbf{MAE(CC)}                       & \textbf{MAE(BC)}                       & \textbf{MAE(EC)}                       & \textbf{MAE(PC)}                       & \textbf{MAE(EFF)}                      & \textbf{MAE(Clst)}                     \\ \hline
\rowcolor[HTML]{FFF2CC} 
\cellcolor[HTML]{FFF2CC}                                                      & \textbf{GAT} & 0.1781                                 & 0.2988                                 & 0.0134                                 & 0.022                                  & {\color[HTML]{00B0F0} \textbf{0.0035}} & 7.4249                                 & 0.429                                  \\ \cline{2-9} 
\rowcolor[HTML]{FFF2CC} 
\multirow{-2}{*}{\cellcolor[HTML]{FFF2CC}\textbf{Adapted MWGAN}}              & \textbf{GCN} & 0.192                                  & 0.2178                                 & 0.0087                                 & 0.013                                  & {\color[HTML]{ff0000} \textbf{0.0033}}    & 6.4845                                 & 0.2879                                 \\ \hline
\rowcolor[HTML]{DEEBF7} 
\cellcolor[HTML]{DEEBF7}                                                      & \textbf{GAT} & 0.2003                                 & 0.1823                                 & 0.007                                  & 0.0118                                 & 0.0071                                 & 5.6466                                 & 0.2305                                 \\ \cline{2-9} 
\rowcolor[HTML]{DEEBF7} 
\multirow{-2}{*}{\cellcolor[HTML]{DEEBF7}\textbf{Adapted MWGAN (clustering)}} & \textbf{GCN} & 0.201                                  & 0.0461                                 & 0.0015                                 & 0.0055                                 & 0.0049                                 & 1.5199                                 & 0.0495                                 \\ \hline
\rowcolor[HTML]{FBE5D6} 
\cellcolor[HTML]{FBE5D6}                                                      & \textbf{GAT} & 0.1887                                 & 0.2331                                 & 0.0096                                 & 0.0201                                 & 0.0041                                 & 6.4751                                 & 0.3002                                 \\ \cline{2-9} 
\rowcolor[HTML]{FBE5D6} 
\multirow{-2}{*}{\cellcolor[HTML]{FBE5D6}\textbf{topoGAN+CC}}                 & \textbf{GCN} & {\color[HTML]{ff0000} \textbf{0.1827}}    & 0.0581                                 & 0.0019                                 & 0.0059                                 & 0.0042                                 & 1.8975                                 & 0.0628                                 \\ \hline
\rowcolor[HTML]{E2F0D9} 
\cellcolor[HTML]{E2F0D9}                                                      & \textbf{GAT} & {\color[HTML]{00B0F0} \textbf{0.1818}} & {\color[HTML]{00B0F0} \textbf{0.1133}} & 0.0041                                 & 0.0102                                 & 0.005                                  & {\color[HTML]{00B0F0} \textbf{3.6689}} & {\color[HTML]{00B0F0} \textbf{0.1319}} \\ \cline{2-9} 
\rowcolor[HTML]{E2F0D9} 
\multirow{-2}{*}{\cellcolor[HTML]{E2F0D9}\textbf{topoGAN+BC}}                 & \textbf{GCN} & 0.2043                                 & 0.0552                                 & 0.0019                                 & 0.0064                                 & 0.0045                                 & 1.8496                                 & 0.061                                  \\ \hline
\rowcolor[HTML]{F5D4FF} 
\cellcolor[HTML]{F5D4FF}                                                      & \textbf{GAT} & 0.2152                                 & 0.1678                                 & {\color[HTML]{00B0F0} \textbf{0.0063}} & {\color[HTML]{00B0F0} \textbf{0.0098}} & 0.0055                                 & 5.3983                                 & 0.2137                                 \\ \cline{2-9} 
\rowcolor[HTML]{F5D4FF} 
\multirow{-2}{*}{\cellcolor[HTML]{F5D4FF}\textbf{topoGAN+EC}}                 & \textbf{GCN} & 0.1903                                 & {\color[HTML]{ff0000} \textbf{0.0262}}    & {\color[HTML]{ff0000} \textbf{0.0007}}    & {\color[HTML]{ff0000} \textbf{0.0041}}    & 0.005                                  & {\color[HTML]{ff0000} \textbf{0.7618}}    & {\color[HTML]{ff0000} \textbf{0.0242}}    \\ \hline
\rowcolor[HTML]{FFD8CE} 
\textbf{MultiGraphGAN+CC}                                                     & \textbf{GCN} & 0.1927                                 & 0.1103                                 & 0.004                                  & 0.0089                                 & 0.0055                                 & 3.6406                                 & 0.1301                                 \\ \hline
\rowcolor[HTML]{F6F9D4} 
\textbf{MultiGraphGAN+BC}                                                     & \textbf{GCN} & 0.192                                  & 0.0567                                 & 0.0018                                 & 0.0055                                 & 0.0039                                 & 1.905                                  & 0.063                                  \\ \hline
\rowcolor[HTML]{DEDCE6} 
\textbf{MultiGraphGAN+EC}                                                     & \textbf{GCN} & 0.1889                                 & 0.0374                                 & 0.0012                                 & 0.0048                                 & 0.0045                                 & 1.2506                                 & 0.0403                                 \\ \hline
\rowcolor[HTML]{FFFFFF} 
\multicolumn{2}{|c|}{\cellcolor[HTML]{FFFFFF}\textbf{View 2}}                                   & \textbf{MAE}                           & \textbf{MAE(CC)}                       & \textbf{MAE(BC)}                       & \textbf{MAE(EC)}                       & \textbf{MAE(PC)}                       & \textbf{MAE(EFF)}                      & \textbf{MAE(Clst)}                     \\ \hline
\rowcolor[HTML]{FFF2CC} 
\cellcolor[HTML]{FFF2CC}                                                      & \textbf{GAT} & 0.2317                                 & 0.293                                  & 0.0131                                 & 0.0253                                 & 0.0059                                 & 7.4492                                 & 0.4206                                 \\ \cline{2-9} 
\rowcolor[HTML]{FFF2CC} 
\multirow{-2}{*}{\cellcolor[HTML]{FFF2CC}\textbf{Adapted MWGAN}}              & \textbf{GCN} & 0.2152                                 & 0.2665                                 & 0.0115                                 & 0.02                                   & {\color[HTML]{ff0000} \textbf{0.0048}}    & 7.0436                                 & 0.3604                                 \\ \hline
\rowcolor[HTML]{DEEBF7} 
\cellcolor[HTML]{DEEBF7}                                                      & \textbf{GAT} & {\color[HTML]{00B0F0} \textbf{0.2243}} & 0.1804                                 & 0.0069                                 & 0.012                                  & 0.0045                                 & 5.6719                                 & 0.231                                  \\ \cline{2-9} 
\rowcolor[HTML]{DEEBF7} 
\multirow{-2}{*}{\cellcolor[HTML]{DEEBF7}\textbf{Adapted MWGAN (clustering)}} & \textbf{GCN} & 0.2252                                 & 0.2257                                 & 0.0093                                 & 0.0193                                 & 0.0052                                 & 6.3599                                 & 0.2881                                 \\ \hline
\rowcolor[HTML]{FBE5D6} 
\cellcolor[HTML]{FBE5D6}                                                      & \textbf{GAT} & 0.2247                                 & 0.2581                                 & 0.011                                  & 0.0212                                 & {\color[HTML]{00B0F0} \textbf{0.0042}} & 6.7694                                 & 0.3355                                 \\ \cline{2-9} 
\rowcolor[HTML]{FBE5D6} 
\multirow{-2}{*}{\cellcolor[HTML]{FBE5D6}\textbf{topoGAN+CC}}                 & \textbf{GCN} & 0.2266                                 & 0.2322                                 & 0.0096                                 & 0.0162                                 & 0.0057                                 & 6.6855                                 & 0.3096                                 \\ \hline
\rowcolor[HTML]{E2F0D9} 
\cellcolor[HTML]{E2F0D9}                                                      & \textbf{GAT} & 0.2302                                 & 0.186                                  & 0.0072                                 & 0.0116                                 & 0.0054                                 & 5.7792                                 & 0.2381                                 \\ \cline{2-9} 
\rowcolor[HTML]{E2F0D9} 
\multirow{-2}{*}{\cellcolor[HTML]{E2F0D9}\textbf{topoGAN+BC}}                 & \textbf{GCN} & 0.2231                                 & {\color[HTML]{ff0000} \textbf{0.1744}}    & {\color[HTML]{ff0000} \textbf{0.0067}}    & {\color[HTML]{ff0000} \textbf{0.0117}}    & 0.0051                                 & {\color[HTML]{ff0000} \textbf{5.4989}}    & {\color[HTML]{ff0000} \textbf{0.2211}}    \\ \hline
\rowcolor[HTML]{F5D4FF} 
\cellcolor[HTML]{F5D4FF}                                                      & \textbf{GAT} & 0.2282                                 & {\color[HTML]{00B0F0} \textbf{0.1414}} & {\color[HTML]{00B0F0} \textbf{0.0052}} & {\color[HTML]{00B0F0} \textbf{0.0107}} & 0.0058                                 & {\color[HTML]{00B0F0} \textbf{4.4982}} & {\color[HTML]{00B0F0} \textbf{0.1694}} \\ \cline{2-9} 
\rowcolor[HTML]{F5D4FF} 
\multirow{-2}{*}{\cellcolor[HTML]{F5D4FF}\textbf{topoGAN+EC}}                 & \textbf{GCN} & 0.2239                                 & 0.2468                                 & 0.0103                                 & 0.0148                                 & 0.0056                                 & 7.1952                                 & 0.3464                                 \\ \hline
\rowcolor[HTML]{FFD8CE} 
\textbf{MultiGraphGAN+CC}                                                     & \textbf{GCN} & 0.2228                                 & 0.1961                                 & 0.0077                                 & 0.0145                                 & 0.0051                                 & 5.9207                                 & 0.25                                   \\ \hline
\rowcolor[HTML]{F6F9D4} 
\textbf{MultiGraphGAN+BC}                                                     & \textbf{GCN} & 0.224                                  & 0.2318                                 & 0.0095                                 & 0.0184                                 & 0.0058                                 & 6.6384                                 & 0.3067                                 \\ \hline
\rowcolor[HTML]{DEDCE6} 
\textbf{MultiGraphGAN+EC}                                                     & \textbf{GCN} & {\color[HTML]{ff0000} \textbf{0.2218}}    & 0.1777                                 & 0.0068                                 & 0.0126                                 & 0.0052                                 & 5.513                                  & 0.2233                                 \\ \hline
\rowcolor[HTML]{FFFFFF} 
\multicolumn{2}{|c|}{\cellcolor[HTML]{FFFFFF}\textbf{View 3}}                                   & \textbf{MAE}                           & \textbf{MAE(CC)}                       & \textbf{MAE(BC)}                       & \textbf{MAE(EC)}                       & \textbf{MAE(PC)}                       & \textbf{MAE(EFF)}                      & \textbf{MAE(Clst)}                     \\ \hline
\rowcolor[HTML]{FFF2CC} 
\cellcolor[HTML]{FFF2CC}                                                      & \textbf{GAT} & 0.1483                                 & 0.2404                                 & 0.01                                   & 0.0179                                 & 0.0034                                 & 6.7071                                 & 0.3171                                 \\ \cline{2-9} 
\rowcolor[HTML]{FFF2CC} 
\multirow{-2}{*}{\cellcolor[HTML]{FFF2CC}\textbf{Adapted MWGAN}}              & \textbf{GCN} & 0.1156                                 & 0.1806                                 & 0.0069                                 & 0.011                                  & 0.0028                                 & 5.635                                  & 0.2296                                 \\ \hline
\rowcolor[HTML]{DEEBF7} 
\cellcolor[HTML]{DEEBF7}                                                      & \textbf{GAT} & 0.1556                                 & 0.1732                                 & 0.0066                                 & 0.012                                  & 0.0044                                 & 5.4467                                 & 0.2183                                 \\ \cline{2-9} 
\rowcolor[HTML]{DEEBF7} 
\multirow{-2}{*}{\cellcolor[HTML]{DEEBF7}\textbf{Adapted MWGAN (clustering)}} & \textbf{GCN} & 0.1206                                 & 0.0548                                 & {\color[HTML]{ff0000} \textbf{0.0018}}    & {\color[HTML]{ff0000} \textbf{0.0052}}    & 0.0032                                 & 1.8435                                 & 0.0608                                 \\ \hline
\rowcolor[HTML]{FBE5D6} 
\cellcolor[HTML]{FBE5D6}                                                      & \textbf{GAT} & 0.1935                                 & 0.1784                                 & 0.0068                                 & 0.0123                                 & 0.0063                                 & 5.566                                  & 0.2257                                 \\ \cline{2-9} 
\rowcolor[HTML]{FBE5D6} 
\multirow{-2}{*}{\cellcolor[HTML]{FBE5D6}\textbf{topoGAN+CC}}                 & \textbf{GCN} & {\color[HTML]{ff0000} \textbf{0.1087}}    & 0.0626                                 & 0.0021                                 & 0.0065                                 & 0.0028                                 & 2.0644                                 & 0.0687                                 \\ \hline
\rowcolor[HTML]{E2F0D9} 
\cellcolor[HTML]{E2F0D9}                                                      & \textbf{GAT} & 0.1234                                 & {\color[HTML]{00B0F0} \textbf{0.1303}} & {\color[HTML]{00B0F0} \textbf{0.0047}} & {\color[HTML]{00B0F0} \textbf{0.0098}} & {\color[HTML]{00B0F0} \textbf{0.0033}} & {\color[HTML]{00B0F0} \textbf{4.187}}  & {\color[HTML]{00B0F0} \textbf{0.1548}} \\ \cline{2-9} 
\rowcolor[HTML]{E2F0D9} 
\multirow{-2}{*}{\cellcolor[HTML]{E2F0D9}\textbf{topoGAN+BC}}                 & \textbf{GCN} & 0.1324                                 & 0.0594                                 & 0.002                                  & 0.0063                                 & 0.0028                                 & 1.9572                                 & 0.0648                                 \\ \hline
\rowcolor[HTML]{F5D4FF} 
\cellcolor[HTML]{F5D4FF}                                                      & \textbf{GAT} & {\color[HTML]{00B0F0} \textbf{0.1221}} & 0.1445                                 & 0.0053                                 & 0.0102                                 & {\color[HTML]{00B0F0} \textbf{0.0033}} & 4.5529                                 & 0.1726                                 \\ \cline{2-9} 
\rowcolor[HTML]{F5D4FF} 
\multirow{-2}{*}{\cellcolor[HTML]{F5D4FF}\textbf{topoGAN+EC}}                 & \textbf{GCN} & 0.1323                                 & {\color[HTML]{ff0000} \textbf{0.0545}}    & {\color[HTML]{ff0000} \textbf{0.0018}}    & {\color[HTML]{ff0000} \textbf{0.0052}}    & 0.0037                                 & {\color[HTML]{ff0000} \textbf{1.8202}}    & {\color[HTML]{ff0000} \textbf{0.0601}}    \\ \hline
\rowcolor[HTML]{FFD8CE} 
\textbf{MultiGraphGAN+CC}                                                     & \textbf{GCN} & 0.1336                                 & 0.0769                                 & 0.0026                                 & 0.0078                                 & 0.0029                                 & 2.502                                  & 0.085                                  \\ \hline
\rowcolor[HTML]{F6F9D4} 
\textbf{MultiGraphGAN+BC}                                                     & \textbf{GCN} & 0.1195                                 & 0.0686                                 & 0.0023                                 & 0.0062                                 & {\color[HTML]{ff0000} \textbf{0.0027}}    & 2.2984                                 & 0.0772                                 \\ \hline
\rowcolor[HTML]{DEDCE6} 
\textbf{MultiGraphGAN+EC}                                                     & \textbf{GCN} & 0.1545                                 & 0.0799                                 & 0.0027                                 & 0.0082                                 & 0.0031                                 & 2.5771                                 & 0.0877                                 \\ \hline
%\multicolumn{9}{p{750}}{View 1: maximum principal curvature (LH). View 2: average curvature (LH). View 3: mean sulcal depth (LH). MAE: mean absolute error. CC: closeness centrality. BC: betweenness centrality. EC: eigenvector centrality. PC: PageRank centrality. EFF: effective size. Clst: clustering coefficient. We highlight in red and blue colors the lowest MAE resulting from a particular evaluation metric for the GCN and GAT versions of topoGAN, respectively.}
\end{tabular}
\begin{tablenotes}[para,flushleft]
      \footnotesize
      \item View 1: maximum principal curvature (LH). View 2: average curvature (LH). View 3: mean sulcal depth (LH). MAE: mean absolute error. CC: closeness centrality. BC: betweenness centrality. EC: eigenvector centrality. PC: PageRank centrality. EFF: effective size. Clst: clustering coefficient. We highlight in red and blue colors the lowest MAE resulting from a particular evaluation metric for the GCN and GAT versions of topoGAN, respectively.
\end{tablenotes}
\end{threeparttable}
\label{MAE1}
\end{table}
\end{landscape}

\begin{landscape}
\begin{table}
\begin{threeparttable}
\hspace{-90pt}
\vspace{-50pt}
\centering
\captionsetup{justification=centering}
\caption{Average prediction results of a target brain multigraph from three source views derived from the right hemisphere using different evaluation metrics.} %}
\footnotesize
\begin{tabular}{|c|c|c|c|c|c|c|c|c|}
\hline
\rowcolor[HTML]{FFFFFF} 
\multicolumn{2}{|c|}{\cellcolor[HTML]{FFFFFF}\textbf{View 4}}                                   & \textbf{MAE}                           & \textbf{MAE(CC)}                       & \textbf{MAE(BC)}                       & \textbf{MAE(EC)}                       & \textbf{MAE(PC)}                       & \textbf{MAE(EFF)}                      & \textbf{MAE(Clst)}                     \\ \hline
\rowcolor[HTML]{FFF2CC} 
\cellcolor[HTML]{FFF2CC}                                                      & \textbf{GAT} & 0.2096                                 & 0.3242                                 & 0.015                                  & 0.0245                                 & 0.0076                                 & 7.8095                                 & 0.5078                                 \\ \cline{2-9} 
\rowcolor[HTML]{FFF2CC} 
\multirow{-2}{*}{\cellcolor[HTML]{FFF2CC}\textbf{Adapted MWGAN}}              & \textbf{GCN} & 0.1844                                 & 0.2193                                 & 0.0088                                 & 0.014                                  & {\color[HTML]{ff0000} \textbf{0.0033}}    & 6.6041                                 & 0.2949                                 \\ \hline
\rowcolor[HTML]{DEEBF7} 
\cellcolor[HTML]{DEEBF7}                                                      & \textbf{GAT} & 0.1977                                 & 0.1878                                 & 0.0073                                 & 0.0132                                 & 0.0074                                 & 5.6507                                 & 0.2333                                 \\ \cline{2-9} 
\rowcolor[HTML]{DEEBF7} 
\multirow{-2}{*}{\cellcolor[HTML]{DEEBF7}\textbf{Adapted MWGAN (clustering)}} & \textbf{GCN} & 0.1886                                 & 0.0456                                 & 0.0015                                 & 0.0049                                 & 0.0042                                 & 1.5334                                 & {\color[HTML]{ff0000} \textbf{0.05}}      \\ \hline
\rowcolor[HTML]{FBE5D6} 
\cellcolor[HTML]{FBE5D6}                                                      & \textbf{GAT} & 0.1978                                 & 0.175                                  & 0.0067                                 & 0.0117                                 & 0.0063                                 & 5.4843                                 & 0.2201                                 \\ \cline{2-9} 
\rowcolor[HTML]{FBE5D6} 
\multirow{-2}{*}{\cellcolor[HTML]{FBE5D6}\textbf{topoGAN+CC}}                 & \textbf{GCN} & 0.1878                                 & 0.0558                                 & 0.0019                                 & 0.0069                                 & 0.0047                                 & 1.7895                                 & 0.0591                                 \\ \hline
\rowcolor[HTML]{E2F0D9} 
\cellcolor[HTML]{E2F0D9}                                                      & \textbf{GAT} & {\color[HTML]{00B0F0} \textbf{0.1906}} & {\color[HTML]{00B0F0} \textbf{0.1448}} & {\color[HTML]{00B0F0} \textbf{0.0053}} & {\color[HTML]{00B0F0} \textbf{0.0103}} & {\color[HTML]{00B0F0} \textbf{0.004}}  & {\color[HTML]{00B0F0} \textbf{4.5976}} & {\color[HTML]{00B0F0} \textbf{0.1747}} \\ \cline{2-9} 
\rowcolor[HTML]{E2F0D9} 
\multirow{-2}{*}{\cellcolor[HTML]{E2F0D9}\textbf{topoGAN+BC}}                 & \textbf{GCN} & {\color[HTML]{ff0000} \textbf{0.1758}}    & 0.0672                                 & 0.0022                                 & 0.0076                                 & 0.0044                                 & 2.1657                                 & 0.0725                                 \\ \hline
\rowcolor[HTML]{F5D4FF} 
\cellcolor[HTML]{F5D4FF}                                                      & \textbf{GAT} & 0.231                                  & 0.1535                                 & 0.0057                                 & 0.0104                                 & 0.0054                                 & 4.8375                                 & 0.1864                                 \\ \cline{2-9} 
\rowcolor[HTML]{F5D4FF} 
\multirow{-2}{*}{\cellcolor[HTML]{F5D4FF}\textbf{topoGAN+EC}}                 & \textbf{GCN} & 0.1875                                 & 0.0569                                 & 0.0018                                 & 0.0072                                 & 0.0046                                 & 1.7222                                 & 0.0566                                 \\ \hline
\rowcolor[HTML]{FFD8CE} 
\textbf{MultiGraphGAN+CC}                                                     & \textbf{GCN} & 0.1816                                 & 0.04621                                & 0.0015                                 & {\color[HTML]{ff0000} \textbf{0.0046}}    & 0.0043                                 & 1.5496                                 & 0.0506                                 \\ \hline
\rowcolor[HTML]{F6F9D4} 
\textbf{MultiGraphGAN+BC}                                                     & \textbf{GCN} & 0.2008                                 & 0.0845                                 & 0.0029                                 & 0.0082                                 & 0.0048                                 & 2.7354                                 & 0.0939                                 \\ \hline
\rowcolor[HTML]{DEDCE6} 
\textbf{MultiGraphGAN+EC}                                                     & \textbf{GCN} & 0.195                                  & {\color[HTML]{ff0000} \textbf{0.0435}}    & {\color[HTML]{ff0000} \textbf{0.0014}}    & 0.0054                                 & 0.0046                                 & {\color[HTML]{ff0000} \textbf{1.4378}}    & {\color[HTML]{ff0000} \textbf{0.05}}      \\ \hline
\rowcolor[HTML]{FFFFFF} 
\multicolumn{2}{|c|}{\cellcolor[HTML]{FFFFFF}\textbf{View 5}}                                   & \textbf{MAE}                           & \textbf{MAE(CC)}                       & \textbf{MAE(BC)}                       & \textbf{MAE(EC)}                       & \textbf{MAE(PC)}                       & \textbf{MAE(EFF)}                      & \textbf{MAE(Clst)}                     \\ \hline
\rowcolor[HTML]{FFF2CC} 
\cellcolor[HTML]{FFF2CC}                                                      & \textbf{GAT} & 0.2267                                 & 0.3022                                 & 0.0136                                 & 0.0245                                 & 0.0058                                 & 7.5284                                 & 0.4402                                 \\ \cline{2-9} 
\rowcolor[HTML]{FFF2CC} 
\multirow{-2}{*}{\cellcolor[HTML]{FFF2CC}\textbf{Adapted MWGAN}}              & \textbf{GCN} & {\color[HTML]{ff0000} \textbf{0.2175}}    & 0.2583                                 & 0.0109                                 & 0.0176                                 & 0.0042                                 & 7.1924                                 & 0.3595                                 \\ \hline
\rowcolor[HTML]{DEEBF7} 
\cellcolor[HTML]{DEEBF7}                                                      & \textbf{GAT} & 0.2256                                 & {\color[HTML]{00B0F0} \textbf{0.1475}} & {\color[HTML]{00B0F0} \textbf{0.0055}} & 0.0133                                 & {\color[HTML]{00B0F0} \textbf{0.0029}} & 4.5758                                 & {\color[HTML]{00B0F0} \textbf{0.1747}} \\ \cline{2-9} 
\rowcolor[HTML]{DEEBF7} 
\multirow{-2}{*}{\cellcolor[HTML]{DEEBF7}\textbf{Adapted MWGAN (clustering)}} & \textbf{GCN} & 0.2198                                 & {\color[HTML]{ff0000} \textbf{0.169}}     & {\color[HTML]{ff0000} \textbf{0.0064}}    & 0.0122                                 & 0.0042                                 & 5.3084                                 & 0.2118                                 \\ \hline
\rowcolor[HTML]{FBE5D6} 
\cellcolor[HTML]{FBE5D6}                                                      & \textbf{GAT} & 0.2223                                 & 0.1542                                 & 0.0058                                 & 0.0101                                 & 0.0033                                 & 4.887                                  & 0.1884                                 \\ \cline{2-9} 
\rowcolor[HTML]{FBE5D6} 
\multirow{-2}{*}{\cellcolor[HTML]{FBE5D6}\textbf{topoGAN+CC}}                 & \textbf{GCN} & 0.2215                                 & 0.1691                                 & 0.0065                                 & 0.0126                                 & {\color[HTML]{ff0000} \textbf{0.0041}}    & {\color[HTML]{ff0000} \textbf{5.23}}      & {\color[HTML]{ff0000} \textbf{0.2083}}    \\ \hline
\rowcolor[HTML]{E2F0D9} 
\cellcolor[HTML]{E2F0D9}                                                      & \textbf{GAT} & 0.2317                                 & 0.1982                                 & 0.0078                                 & 0.016                                  & 0.0035                                 & 5.8761                                 & 0.2491                                 \\ \cline{2-9} 
\rowcolor[HTML]{E2F0D9} 
\multirow{-2}{*}{\cellcolor[HTML]{E2F0D9}\textbf{topoGAN+BC}}                 & \textbf{GCN} & 0.2232                                 & 0.1993                                 & 0.0079                                 & 0.0155                                 & 0.0051                                 & 5.9041                                 & 0.2511                                 \\ \hline
\rowcolor[HTML]{F5D4FF} 
\cellcolor[HTML]{F5D4FF}                                                      & \textbf{GAT} & {\color[HTML]{00B0F0} \textbf{0.2218}} & 0.1673                                 & 0.0063                                 & {\color[HTML]{00B0F0} \textbf{0.0083}} & 0.0035                                 & {\color[HTML]{00B0F0} \textbf{5.3628}} & 0.2118                                 \\ \cline{2-9} 
\rowcolor[HTML]{F5D4FF} 
\multirow{-2}{*}{\cellcolor[HTML]{F5D4FF}\textbf{topoGAN+EC}}                 & \textbf{GCN} & 0.2224                                 & 0.2179                                 & 0.0088                                 & 0.0168                                 & 0.0047                                 & 6.3244                                 & 0.2816                                 \\ \hline
\rowcolor[HTML]{FFD8CE} 
\textbf{MultiGraphGAN+CC}                                                     & \textbf{GCN} & 0.2206                                 & 0.1748                                 & 0.0067                                 & {\color[HTML]{ff0000} \textbf{0.0117}}    & 0.0045                                 & 5.4439                                 & 0.2193                                 \\ \hline
\rowcolor[HTML]{F6F9D4} 
\textbf{MultiGraphGAN+BC}                                                     & \textbf{GCN} & 0.2243                                 & 0.2385                                 & 0.0099                                 & 0.0199                                 & 0.0043                                 & 6.7307                                 & 0.3176                                 \\ \hline
\rowcolor[HTML]{DEDCE6} 
\textbf{MultiGraphGAN+EC}                                                     & \textbf{GCN} & 0.2189                                 & 0.2044                                 & 0.0081                                 & 0.0161                                 & 0.0049                                 & 5.8971                                 & 0.2537                                 \\ \hline
\rowcolor[HTML]{FFFFFF} 
\multicolumn{2}{|c|}{\cellcolor[HTML]{FFFFFF}\textbf{View 6}}                                   & \textbf{MAE}                           & \textbf{MAE(CC)}                       & \textbf{MAE(BC)}                       & \textbf{MAE(EC)}                       & \textbf{MAE(PC)}                       & \textbf{MAE(EFF)}                      & \textbf{MAE(Clst)}                     \\ \hline
\rowcolor[HTML]{FFF2CC} 
\cellcolor[HTML]{FFF2CC}                                                      & \textbf{GAT} & 0.1409                                 & 0.2466                                 & 0.0103                                 & 0.0176                                 & {\color[HTML]{00B0F0} \textbf{0.0032}} & 6.9177                                 & 0.3326                                 \\ \cline{2-9} 
\rowcolor[HTML]{FFF2CC} 
\multirow{-2}{*}{\cellcolor[HTML]{FFF2CC}\textbf{Adapted MWGAN}}              & \textbf{GCN} & 0.1096                                 & 0.3107                                 & 0.0141                                 & 0.0167                                 & 0.0046                                 & 7.5565                                 & 0.4564                                 \\ \hline
\rowcolor[HTML]{DEEBF7} 
\cellcolor[HTML]{DEEBF7}                                                      & \textbf{GAT} & 0.1288                                 & 0.2056                                 & 0.0082                                 & 0.0144                                 & 0.0057                                 & 6.1356                                 & 0.2652                                 \\ \cline{2-9} 
\rowcolor[HTML]{DEEBF7} 
\multirow{-2}{*}{\cellcolor[HTML]{DEEBF7}\textbf{Adapted MWGAN (clustering)}} & \textbf{GCN} & 0.1121                                 & 0.1351                                 & 0.0049                                 & 0.0105                                 & 0.0051                                 & 4.349                                  & 0.1627                                 \\ \hline
\rowcolor[HTML]{FBE5D6} 
\cellcolor[HTML]{FBE5D6}                                                      & \textbf{GAT} & 0.1919                                 & 0.1854                                 & 0.0072                                 & 0.0138                                 & 0.0049                                 & 5.6401                                 & 0.2332                                 \\ \cline{2-9} 
\rowcolor[HTML]{FBE5D6} 
\multirow{-2}{*}{\cellcolor[HTML]{FBE5D6}\textbf{topoGAN+CC}}                 & \textbf{GCN} & 0.115                                  & 0.1004                                 & 0.0035                                 & 0.0084                                 & 0.0039                                 & 3.2629                                 & 0.1149                                 \\ \hline
\rowcolor[HTML]{E2F0D9} 
\cellcolor[HTML]{E2F0D9}                                                      & \textbf{GAT} & {\color[HTML]{00B0F0} \textbf{0.0805}} & {\color[HTML]{00B0F0} \textbf{0.1597}} & {\color[HTML]{00B0F0} \textbf{0.006}}  & {\color[HTML]{00B0F0} \textbf{0.0097}} & 0.0068                                 & 5.151                                  & 0.2007                                 \\ \cline{2-9} 
\rowcolor[HTML]{E2F0D9} 
\multirow{-2}{*}{\cellcolor[HTML]{E2F0D9}\textbf{topoGAN+BC}}                 & \textbf{GCN} & 0.1028                                 & {\color[HTML]{ff0000} \textbf{0.0722}}    & {\color[HTML]{ff0000} \textbf{0.0025}}    & 0.0081                                 & {\color[HTML]{ff0000} \textbf{0.0037}}    & {\color[HTML]{ff0000} \textbf{2.2983}}    & {\color[HTML]{ff0000} \textbf{0.0776}}    \\ \hline
\rowcolor[HTML]{F5D4FF} 
\cellcolor[HTML]{F5D4FF}                                                      & \textbf{GAT} & 0.1008                                 & 0.1617                                 & 0.0061                                 & 0.0127                                 & 0.0048                                 & {\color[HTML]{00B0F0} \textbf{5.0954}} & {\color[HTML]{00B0F0} \textbf{0.1999}} \\ \cline{2-9} 
\rowcolor[HTML]{F5D4FF} 
\multirow{-2}{*}{\cellcolor[HTML]{F5D4FF}\textbf{topoGAN+EC}}                 & \textbf{GCN} & 0.0772                                 & 0.086                                  & 0.003                                  & {\color[HTML]{ff0000} \textbf{0.0073}}    & 0.0045                                 & 2.8036                                 & 0.0966                                 \\ \hline
\rowcolor[HTML]{FFD8CE} 
\textbf{MultiGraphGAN+CC}                                                     & \textbf{GCN} & 0.1825                                 & 0.1793                                 & 0.0069                                 & 0.0144                                 & 0.0054                                 & 5.5751                                 & 0.2273                                 \\ \hline
\rowcolor[HTML]{F6F9D4} 
\textbf{MultiGraphGAN+BC}                                                     & \textbf{GCN} & 0.1587                                 & 0.1671                                 & 0.0063                                 & 0.0109                                 & 0.0052                                 & 5.1932                                 & 0.2061                                 \\ \hline
\rowcolor[HTML]{DEDCE6} 
\textbf{MultiGraphGAN+EC}                                                     & \textbf{GCN} & {\color[HTML]{ff0000} \textbf{0.0767}}    & 0.1178                                 & 0.0042                                 & 0.0101                                 & 0.0052                                 & 3.7543                                 & 0.1357                                 \\ \hline
%\multicolumn{9}{p{750}}{View 4: maximum principal curvature (RH). View 5: average curvature (RH). View 6: mean sulcal depth (RH). MAE: mean absolute error. CC: closeness centrality. BC: betweenness centrality. EC: eigenvector centrality. PC: PageRank centrality. EFF: effective size. Clst: clustering coefficient. We highlight in red and blue colors the lowest MAE resulting from a particular evaluation metric for the GCN and GAT versions of topoGAN, respectively.}
\end{tabular}
\begin{tablenotes}[para,flushleft]
      \footnotesize
      \item View 4: maximum principal curvature (RH). View 5: average curvature (RH). View 6: mean sulcal depth (RH). MAE: mean absolute error. CC: closeness centrality. BC: betweenness centrality. EC: eigenvector centrality. PC: PageRank centrality. EFF: effective size. Clst: clustering coefficient. We highlight in red and blue colors the lowest MAE resulting from a particular evaluation metric for the GCN and GAT versions of topoGAN, respectively.
\end{tablenotes}
\end{threeparttable}
\label{MAE2}
\end{table}
\end{landscape}

% %% ***************************************************************************** %%
\section{Experiments and Discussion}
% %% ***************************************************************************** %%

\subsection{Multi-view brain graph dataset}

For evaluation, we use the Autism Brain Imaging Data Exchange (ABIDE\footnote{\underline{http://fcon\_1000.projects.nitrc.org/indi/abide/}}) including 310 subjects (i.e, structural T1-w MRI). We reconstruct both cortical hemispheres by FreeSurfer \citep{Fischl:2012} and then parcellate each cortical hemisphere into 35 cortical regions using Desikan-Killiany Atlas. For each hemisphere, we extract three morphological brain graphs (i.e, MBG), which means each subject in our dataset is represented by six MBGs. We use the following cortical measurements to extract a pair of graphs for left and right hemispheres: maximum principal curvature (i.e, view 1 and view 4), average curvature (i.e, view 2 and view 5) and mean sulcal depth (i.e, view 3 and view 6). Recently introduced in \citep{Mahjoub:2018}, morphological brain networks (MBNs) quantify the dissimilarity in morphology between pairs of ROIs in the cortex. Essentially, given a particular cortical attribute, (e.g, mean sulcal depth), we compute its average across all vertices in a given brain ROI. Next, we define the edge weight between two ROIs as the absolute difference between their average cortical attributes. MBNs have been widely used for healthy \citep{Dhifallah:2020,Nebli:2019,Mhiri:2020} and disordered brain connectivity analysis \citep{Khelifa:2019,Banka:2019,Soussia:2018b}.

\subsection{Model architecture and parameter setting}

As shown in (\textbf{Fig.}~\ref{2}-\rom{1}), we use $(c\times k + 2)$ trainable neural networks: an encoder $E$, a discriminator $D$ and $(c\times k)$ cluster-specific generators $\{G_{T_{i}}^{j}\}_{i=1,j=1}^{k,c}$. We construct our encoder with a hidden layer comprising 32 neurons and an embedding layer with 16 neurons. Conversely, we define all generators with two layers each comprising 16 and 32 neurons. The discriminator comprises three layers each has 32, 16 and 1 neurons, respectively. We add to its last layer a softmax activation function representing our domain classifier. We train our model on 1000 iterations using batch size 70, a learning rate of 0.0001, $\beta_{1} = 0.5$ and $\beta{2} = 0.999$ for Adam optimizer. We use a grid search to set our hyper-parameters $\lambda_{gdc}=1$, $\lambda_{gp}=0.1$, $\lambda_{top}=0.1$, $\lambda_{rec}=0.01$ and $\lambda_{inf}=1$. Specifically, we use the parameter setting that produced the best performance for each method independently. We also train our discriminator five times and the generators one time in order to have better models. By doing so, we are improving their learning performances. For MKML parameters \citep{Wang:2017}, we fix the number of kernels to 10. We vary the number of clusters $c$ between 2 and 4 then we choose the one which gave the best performance $c=2$.

\subsection{Evaluation and comparison methods}

We train our model using two different training strategies: (i) a random split where 90\% of the dataset represents the training set and 10\% of it is used for testing, and (ii) a 3-fold cross-validation where we train the model on two folds and test it on the left fold. To quantitatively evaluate the topoGAN's performance, we use the mean absolute error (MAE) \citep{Lin:1990} computed between ground-truth target graphs and the generated ones and the MAE between the ground-truth centrality scores and the predicted ones. To comprehensively evaluate our model, we further incorporate an additional centrality evaluation including PageRank centrality (PC), effective size (Eff) and the clustering coefficient (Clst). In particular, PageRank centrality introduced in \citep{Brin:1998} assumes that a central node is likely to receive more edges from other nodes. The effective size metric introduced in \citep{Burt:2009} assume that the efficiency of a node is related to the number of non-redundant links in a graph and the clustering coefficient \citep{Saramaki:2007} measures the degree to which nodes in a graph tend to cluster together. Lastly, we consider the average of the resulting MAEs computed for each of the $k$ target domains as the final measures to evaluate our framework. The lower is the MAE the better is the performance of our framework in predicting the target multigraph. Moreover, we compute the Kullback–Leibler divergence \citep{Kullback:1959} between the real and predicted topology scores (i.e, CC, BC, EC, EFF and Clst). Then, we average the KL-divergence results of all $k$ domains and report the results for each source brain graph. Originated in probability theory and information theory, KL-divergence measures the information lost when the predicted distribution approximates the original distribution. Thus, a lower value indicates that the predicted distribution is almost identical to the real one.

We compare our topoGAN framework with two baseline methods and three ablated versions of our model where we use three different centrality metrics. Essentially, each method is implemented with two graph neural network architectures (i.e, GCN \citep{Kipf:2016} and graph attention network (GAT) \citep{Velivckovic:2017}): 
\begin{enumerate}
\item \textbf{Adapted MWGAN:} we use the same architecture proposed in \citep{Cao:2019} that we adapted to graph data types where we neither include the clustering step nor our proposed topological loss term (\textbf{Eq.}~\eqref{eq:10}).
\item \textbf{Adapted MWGAN (clustering):} it is a variant of the first method where we add the MKML clustering of the source graph embeddings
\citep{Wang:2017}. 
\item \textbf{topoGAN+CC}: it is an ablated version of our framework which adopts a clustering strategy and a topological loss function that includes the closeness centrality \citep{Freeman:1977} in the topological loss function (i.e, $\mathcal{C} = CC$ in \textbf{Eq.}~\eqref{eq:10}).
\item \textbf{topoGAN+BC}: in this method we incorporate the betweenness centrality \citep{Beauchamp:1965} in the topological loss function (i.e, $\mathcal{C} = BC$ in \textbf{Eq.}~\eqref{eq:10}).
\item \textbf{topoGAN+EC}: in this method we include the eigenvector centrality \citep{Bonacich:2007} in the topological loss function (i.e, $\mathcal{C} = EC$ in \textbf{Eq.}~\eqref{eq:10}).
\end{enumerate}

\subsection{Results and benchmarking}

We conducted six different experiments, each taking one of the six MBGs as a source graph and the five remaining ones as views stacked in the target brain multigraph to predict. Clearly, supplementary \textbf{Table}~\ref{p-value-1}, \textbf{Table}~\ref{p-value-2} and \textbf{Table}~\ref{p-value-3} show that our topoGAN consistently and significantly ($p$-value $\prec$ 0.05 using two-tailed paired $t$-test) outperformed benchmark methods in predicting brain multigraph from a single source graph.

\subsubsection{Impact of the clustering and global topology loss function}

\begin{figure}[!htpb]
\centering
\vspace{-60pt}
\centerline{\includegraphics[width=13cm]{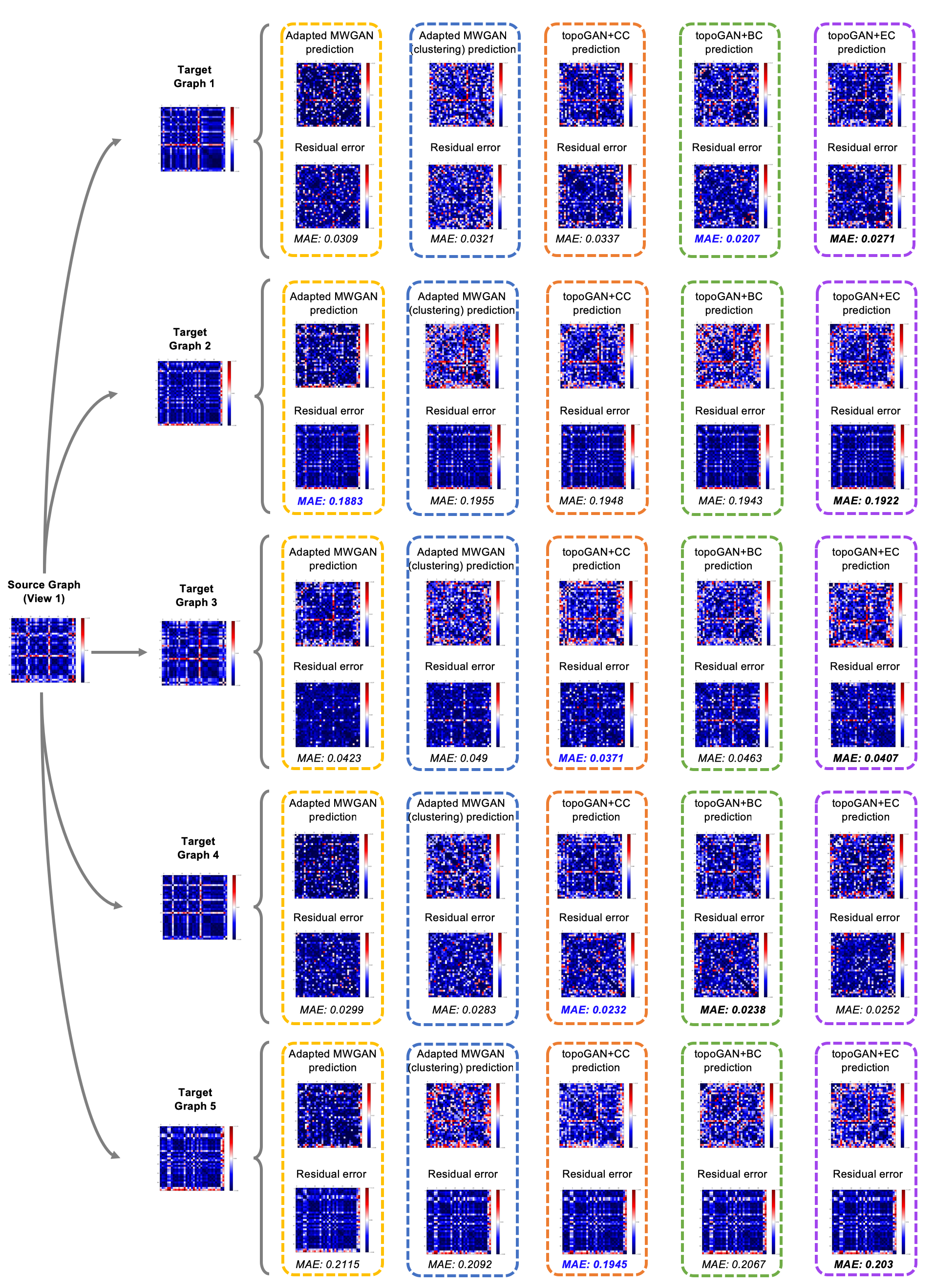}}
\caption{\emph{Visual comparison between the ground-truth and the predicted five views stacked in the target brain multigraph of a representative testing subject using different methods: Adapted MWGAN \cite{Cao:2019}, Adapted MWGAN (clustering), and topoGAN using different centrality metrics.} We display the residual error computed using mean absolute error (MAE) metric between the original brain graph and the predicted graph. Source brain graph is derived from maximum principal curvature (LH) (i.e, view1). Target brain graphs are derived from the average curvature (LH) (Target graph 1), mean sulcal depth (LH) (Target graph 2), maximum principal curvature (RH) (Target graph 3), average curvature (RH) (Target graph 4) and mean sulcal depth (RH) (Target graph 5).}
\label{residu}
\end{figure}

To evaluate the effectiveness of clustering the source graph embeddings and learning the global topological structures of the target graphs, we report in \textbf{Table}~\ref{MAE1} and \textbf{Table}~\ref{MAE2} the average MAE (i.e, first column) computed between the real and predicted graphs of five target domains. These results show that three variants of topoGAN outperformed two baseline methods. Although it produced a slightly higher MAE using one source graph (i.e, view 5), our framework achieved the lowest MAE results using five source graphs (i.e, views 1,2,3,4 and 6). Specifically, methods adopting the CC and BC both ranked first best in six experiments while the method adopting EC ranked second best using the MAE metric. Notably, these results show that our topoGAN using the proposed topological loss function significantly outperforms the baseline methods in preserving the global topology of the original target graphs. We also show that our framework using the \emph{global topology loss} term is better then GCN-based methods which simply leverage graph convolution to learn the graph structure. Additionally, \textbf{Fig.}~\ref{residu} displays the source graph, the ground-truth target graphs and the predicted ones by topoGAN using three centrality measures and the two baseline methods (Adapted MWGAN and Adapted MWGAN (clustering)) for a representative subject. We display below each predicted graph its residual graph, representing the absolute difference between the ground-truth and predicted target graph. We observe that the residual was noticeably reduced by our method using both CC and BC measures. When predicting the target graph 2 (i.e, mean sulcal depth (LH)), Adapted MWGAN is marginally better than three variants of topoGAN. This shows that while the average results of topoGAN outperform benchmark methods, our framework might lag behind for a few particular subjects. However, we argue that although our framework works relatively poorly in predicting the target graph 2, it still achieved better results than the Adapted MWGAN (clustering) in predicting four target graphs for the same representative subject (\textbf{Fig.}~\ref{residu}). This demonstrates the advantage of our \emph{cluster-specific generators} in eventually avoiding the mode collapse problem thereby boosting the performance of topoGAN in the target brain multigraph prediction.

% %% ***************************************************************************** %%
%%% Residual %%%
% %% ***************************************************************************** %%

\subsubsection{Impact of the local topology loss and the reconstruction loss evaluation}

To further evaluate the effectiveness of learning the local topological structures of the original target graphs, we compute the MAE between the ground-truth and the predicted centrality scores, effective size and clustering coefficient. Results reported in \textbf{Table}~\ref{MAE1} and \textbf{Table}~\ref{MAE2} show that our method consistently achieved the best topology-preserving predictions compared with the baseline methods using the six graph topology evaluation measures. Notably, four out of six experiments (i.e, views 1,2,3 and 6) results highlight the importance of using BC and EC which both gave the lowest MAE using different evaluation metrics. This is explicable since considering the node neighborhoods (i.e., EC) and the frequency of being on the shortest path between nodes in the graph (i.e., BC) have much impact on identifying the most influential node rather than focusing on the average shortest path existing between two nodes. As for predicting five target graphs from the source view 4 (derived from the maximum principal curvature of the right hemisphere), topoGAN outperformed both baseline methods (i.e, Adapted MWGAN and Adapted MWGAN (clustering)) in terms of MAE computed using the whole graph, however it produced a slightly higher MAE results computed using the topological measurements (i.e, CC, BC, EC, EFF and Clst). We can see similar results for the view 5 (average curvature of the right hemisphere). We included this example in our results to show that while the average results of topoGAN were remarkable and outperformed comparison methods, our method might lag behind for a few particular source views. Still the Adapted MWGAN (clustering) which is the clustering-based baseline method achieved better results than the Adapted MWGAN method. This clearly demonstrates the advantage of our \emph{cluster-specific generators} for avoiding the mode collapse problem. We can conclusively confirm that our proposed \emph{local topology loss} term optimally improves the learning of the local graph structure when predicting the target graphs. On the other hand, the results reported in \textbf{Table}~\ref{MAE1} and \textbf{Table}~\ref{MAE2} show that the fact of removing the graph reconstruction loss term introduced in \citep{bessadok:2020} from the adversarial loss remarkably improves the learning of our model. Mainly, five out of six experiments (i.e, views 1,2,3,5 and 6) results proved that the graph reconstruction loss does not impact the accuracy of predicting the target multigraph. This is explicable because our aim is to make our prediction similar to the real target multigraph so there is no need to enforce the model to preserve the topology of the source domain in the predicted graphs. This further demonstrates that our proposed framework outperformed the existing state-of-the-art method named MultiGraphGAN \citep{bessadok:2020}. Interestingly, our model achieved better results than comparison methods using two different training strategies: random split and cross-validation. Specifically, we report in  supplementary \textbf{Table}~\ref{CV} results for three GCN-based comparison methods and our GCN-based topoGAN learned using eigenvector centrality in our loss function. More results can be found in the supplementary material.

\subsubsection{Impact of the graph representation learning using GCN}

To further evaluate the effectiveness of GCN in learning the brain representation, we implemented a GAT-based version with 8 heads for each benchmark method. Results reported in \textbf{Table}~\ref{MAE1} and \textbf{Table}~\ref{MAE2} demonstrate that our GCN-based framework outperformed the GAT version for six experiments. This can be explained by the fact that assigning larger weights to the most important nodes in brain graphs is not effective since each ROI have a distinct role in the graph thereby equally learning the nodes' representations is highly needed. Originally, the graph convolution operation updates the local deep features by aggregating the features in a local neighborhood of a particular sample (i.e, node) \citep{Zhang:2019}. Thus, our GCN version of topoGAN combined with the topological constraint perfectly models the larger contextual region within the brain which is important for learning the correct anatomical structures in connectomes. We observe in the \textbf{Table}~\ref{KL1} and \textbf{Table}~\ref{KL2} that the average KL-divergence over five target views was noticeably reduced by our method. Still, our framework produces higher KL-distance using the source views 4 and 5, which is similar to the MAE results reported in \textbf{Table}~\ref{MAE2}. Essentially, views 1, 2, 3 and 6 show that betweenness and eigenvector centralities present better choices to boost the topological property preservation in jointly predicting multiple target graphs. Such good results demonstrate the advantage of our GCN-based framework in learning the source brain graph representation and preserving the structure of the original target graphs which is in line with recent GCN-based brain analysis works \citep{Banka:2019,Bessadok:2019a}.

\subsubsection{Comparison using cross-validation}

For a thorough comparison, we also compare our topoGAN with two baseline methods (i.e., Adapted MWGAN and Adapted MWGAN (clustering)) and MultiGraphGAN method using 3-fold cross-validation strategy. Since GCN-based variants along with the inclusion of EC in our loss function outperformed the GAT-based ones and GCN-based methods that include CC and BC during the learning process, we only include the experimental results of the GCN-based methods. Supplementary \textbf{Table 4} reports the average results over three folds on the same dataset. Our topoGAN achieved better performance than comparison methods when predicting five target graphs from the source views 1, 4 and 5 in terms of MAE computed using the topological measurements (i.e, CC, BC, EFF and Clst), however it produced a slightly higher MAE results computed using the EC and PC evaluation measurements. This is in line with previous results reported in \textbf{Table}~\ref{MAE1} and \textbf{Table}~\ref{MAE2} for the topoGAN+EC method. On the other hand, our topoGAN ranked second best when predicting five target graphs using the source views 2 and 3. This is because the alignment of a source domain to multiple target domains is very challenging, even a one-to-one domain alignment has been recently shown to be a hard task when performed on brain graphs \citep{wang2020discriminative, lgdada}. Thus, integrating a domain alignment module in our topoGAN is our future avenue which will better learn the adaptation of the source distribution to multiple target distributions. Still, these results reported in the supplementary \textbf{Table 4} indicate the superiority of our topoGAN over three experiments (i.e, view 1, 4 and 5) which demonstrate the advantage of our proposed cluster-specific generators and our topological loss function compared to its ablated versions as well as the state-of-the-art method.

\begin{table}
\begin{threeparttable}
\centering
\captionsetup{justification=centering}
\caption{Prediction results for three source views derived from the left hemisphere using KL-divergence evaluation metric.} %}
\footnotesize

\begin{tabular}{|c|c|c|c|c|c|c|c|}
\hline
\rowcolor[HTML]{FFFFFF} 
\multicolumn{2}{|c|}{\cellcolor[HTML]{FFFFFF}\textbf{View 1}}                                   & \textbf{CC}                            & \textbf{BC}                            & \textbf{EC}                            & \textbf{PC}                            & \textbf{EFF}                           & \textbf{Clst}                          \\ \hline
\rowcolor[HTML]{FFF2CC} 
\cellcolor[HTML]{FFF2CC}                                                      & \textbf{GAT} & 0.0023                                 & 0.088                                  & 0.0203                                 & {\color[HTML]{00B0F0} \textbf{0.0124}} & 0.0213                                 & 0.0025                                 \\ \cline{2-8} 
\rowcolor[HTML]{FFF2CC} 
\multirow{-2}{*}{\cellcolor[HTML]{FFF2CC}\textbf{Adapted MWGAN}}              & \textbf{GCN} & 0.0019                                 & 0.0461                                 & 0.0063                                 & {\color[HTML]{ff0000} \textbf{0.0093}}    & 0.0083                                 & 0.0003                                 \\ \hline
\rowcolor[HTML]{DEEBF7} 
\cellcolor[HTML]{DEEBF7}                                                      & \textbf{GAT} & 0.0022                                 & 0.0501                                 & 0.0061                                 & 0.0702                                 & 0.0104                                 & 0.0003                                 \\ \cline{2-8} 
\rowcolor[HTML]{DEEBF7} 
\multirow{-2}{*}{\cellcolor[HTML]{DEEBF7}\textbf{Adapted MWGAN (clustering)}} & \textbf{GCN} & 0.0009                                 & 0.0226                                 & 0.001                                  & 0.027                                  & 0.0038                                 & 1.6E-05                                \\ \hline
\rowcolor[HTML]{FBE5D6} 
\cellcolor[HTML]{FBE5D6}                                                      & \textbf{GAT} & 0.0035                                 & 0.1064                                 & 0.0144                                 & 0.0194                                 & 0.0222                                 & 0.0021                                 \\ \cline{2-8} 
\rowcolor[HTML]{FBE5D6} 
\multirow{-2}{*}{\cellcolor[HTML]{FBE5D6}\textbf{topoGAN+CC}}                 & \textbf{GCN} & 0.001                                  & 0.0209                                 & 0.0011                                 & 0.0206                                 & 0.0031                                 & 1.6E-05                                \\ \hline
\rowcolor[HTML]{E2F0D9} 
\cellcolor[HTML]{E2F0D9}                                                      & \textbf{GAT} & 0.0024                                 & {\color[HTML]{00B0F0} \textbf{0.0369}} & 0.0039                                 & 0.0309                                 & 0.0072                                 & {\color[HTML]{00B0F0} \textbf{0.0001}} \\ \cline{2-8} 
\rowcolor[HTML]{E2F0D9} 
\multirow{-2}{*}{\cellcolor[HTML]{E2F0D9}\textbf{topoGAN+BC}}                 & \textbf{GCN} & 0.0011                                 & 0.027                                  & 0.0014                                 & 0.0223                                 & 0.004                                  & 2.1E-05                                \\ \hline
\rowcolor[HTML]{F5D4FF} 
\cellcolor[HTML]{F5D4FF}                                                      & \textbf{GAT} & {\color[HTML]{00B0F0} \textbf{0.0017}} & 0.0374                                 & {\color[HTML]{00B0F0} \textbf{0.0034}} & 0.0383                                 & {\color[HTML]{00B0F0} \textbf{0.0056}} & {\color[HTML]{00B0F0} \textbf{0.0001}} \\ \cline{2-8} 
\rowcolor[HTML]{F5D4FF} 
\multirow{-2}{*}{\cellcolor[HTML]{F5D4FF}\textbf{topoGAN+EC}}                 & \textbf{GCN} & {\color[HTML]{ff0000} \textbf{0.0006}}    & 0.0215                                 & {\color[HTML]{ff0000} \textbf{0.0006}}    & 0.0282                                 & {\color[HTML]{ff0000} \textbf{0.0025}}    & {\color[HTML]{ff0000} \textbf{0,0000072}} \\ \hline
\rowcolor[HTML]{FFD8CE} 
\textbf{MultiGraphGAN+CC}                                                     & \textbf{GCN} & 0.00167                                & 0.0351                                 & 0.0026                                 & 0.0394                                 & 0.0055                                 & 0,000071                               \\ \hline
\rowcolor[HTML]{F6F9D4} 
\textbf{MultiGraphGAN+BC}                                                     & \textbf{GCN} & 0.0007                                 & {\color[HTML]{ff0000} \textbf{0.0178}}    & 0.0009                                 & 0.0155                                 & {\color[HTML]{ff0000} \textbf{0.0017}}    & 0,0000086                              \\ \hline
\rowcolor[HTML]{DEDCE6} 
\textbf{MultiGraphGAN+EC}                                                     & \textbf{GCN} & 0.0007                                 & 0.0211                                 & 0.0008                                 & 0.0229                                 & 0.0033                                 & 0,000013                               \\ \hline
\rowcolor[HTML]{FFFFFF} 
\multicolumn{2}{|c|}{\cellcolor[HTML]{FFFFFF}\textbf{View 2}}                                   & \textbf{CC}                            & \textbf{BC}                            & \textbf{EC}                            & \textbf{PC}                            & \textbf{EFF}                           & \textbf{Clst}                          \\ \hline
\rowcolor[HTML]{FFF2CC} 
\cellcolor[HTML]{FFF2CC}                                                      & \textbf{GAT} & 0.0038                                 & 0.1718                                 & 0.0235                                 & 0.0415                                 & 0.0314                                 & 0.0045                                 \\ \cline{2-8} 
\rowcolor[HTML]{FFF2CC} 
\multirow{-2}{*}{\cellcolor[HTML]{FFF2CC}\textbf{Adapted MWGAN}}              & \textbf{GCN} & 0.0034                                 & 0.1336                                 & 0.0179                                 & {\color[HTML]{ff0000} \textbf{0.0283}}    & 0.026                                  & 0.0014                                 \\ \hline
\rowcolor[HTML]{DEEBF7} 
\cellcolor[HTML]{DEEBF7}                                                      & \textbf{GAT} & {\color[HTML]{00B0F0} \textbf{0.002}}  & {\color[HTML]{00B0F0} \textbf{0.0366}} & 0.0051                                 & 0.0241                                 & {\color[HTML]{00B0F0} \textbf{0.0067}} & 0.0002                                 \\ \cline{2-8} 
\rowcolor[HTML]{DEEBF7} 
\multirow{-2}{*}{\cellcolor[HTML]{DEEBF7}\textbf{Adapted MWGAN (clustering)}} & \textbf{GCN} & 0.0041                                 & 0.1092                                 & 0.0135                                 & 0.0374                                 & 0.0237                                 & 0.0009                                 \\ \hline
\rowcolor[HTML]{FBE5D6} 
\cellcolor[HTML]{FBE5D6}                                                      & \textbf{GAT} & 0.0039                                 & 0.1646                                 & 0.0196                                 & {\color[HTML]{00B0F0} \textbf{0.021}}  & 0.0398                                 & 0.0033                                 \\ \cline{2-8} 
\rowcolor[HTML]{FBE5D6} 
\multirow{-2}{*}{\cellcolor[HTML]{FBE5D6}\textbf{topoGAN+CC}}                 & \textbf{GCN} & 0.003                                  & 0.068                                  & 0.0128                                 & 0.0415                                 & 0.0132                                 & 0.0006                                 \\ \hline
\rowcolor[HTML]{E2F0D9} 
\cellcolor[HTML]{E2F0D9}                                                      & \textbf{GAT} & 0.0022                                 & 0.0481                                 & 0.0055                                 & 0.031                                  & 0.0084                                 & 0.0003                                 \\ \cline{2-8} 
\rowcolor[HTML]{E2F0D9} 
\multirow{-2}{*}{\cellcolor[HTML]{E2F0D9}\textbf{topoGAN+BC}}                 & \textbf{GCN} & {\color[HTML]{ff0000} \textbf{0.0018}}    & {\color[HTML]{ff0000} \textbf{0.0324}}    & {\color[HTML]{ff0000} \textbf{0.0045}}    & 0.0317                                 & 0.0061                                 & {\color[HTML]{ff0000} \textbf{0.0002}}    \\ \hline
\rowcolor[HTML]{F5D4FF} 
\cellcolor[HTML]{F5D4FF}                                                      & \textbf{GAT} & 0.0023                                 & 0.0501                                 & {\color[HTML]{00B0F0} \textbf{0.0042}} & 0.0371                                 & 0.0086                                 & {\color[HTML]{00B0F0} \textbf{0.0001}} \\ \cline{2-8} 
\rowcolor[HTML]{F5D4FF} 
\multirow{-2}{*}{\cellcolor[HTML]{F5D4FF}\textbf{topoGAN+EC}}                 & \textbf{GCN} & {\color[HTML]{ff0000} \textbf{0.0018}}    & 0.0371                                 & 0.0079                                 & 0.0352                                 & {\color[HTML]{ff0000} \textbf{0.0078}}    & 0.0009                                 \\ \hline
\rowcolor[HTML]{FFD8CE} 
\textbf{MultiGraphGAN+CC}                                                     & \textbf{GCN} & 0.0029                                 & 0.0557                                 & 0.0074                                 & 0.0312                                 & 0.0116                                 & 0.0005                                 \\ \hline
\rowcolor[HTML]{F6F9D4} 
\textbf{MultiGraphGAN+BC}                                                     & \textbf{GCN} & 0.0033                                 & 0.0873                                 & 0.0122                                 & 0.0394                                 & 0.0181                                 & 0.0008                                 \\ \hline
\rowcolor[HTML]{DEDCE6} 
\textbf{MultiGraphGAN+EC}                                                     & \textbf{GCN} & 0.0024                                 & 0.0441                                 & 0.0059                                 & 0.0358                                 & 0.0083                                 & {\color[HTML]{ff0000} \textbf{0.0002}}    \\ \hline
\rowcolor[HTML]{FFFFFF} 
\multicolumn{2}{|c|}{\cellcolor[HTML]{FFFFFF}\textbf{View 3}}                                   & \textbf{CC}                            & \textbf{BC}                            & \textbf{EC}                            & \textbf{PC}                            & \textbf{EFF}                           & \textbf{Clst}                          \\ \hline
\rowcolor[HTML]{FFF2CC} 
\cellcolor[HTML]{FFF2CC}                                                      & \textbf{GAT} & 0.0032                                 & 0.0893                                 & 0.0134                                 & 0.0109                                 & 0.0189                                 & 0.0012                                 \\ \cline{2-8} 
\rowcolor[HTML]{FFF2CC} 
\multirow{-2}{*}{\cellcolor[HTML]{FFF2CC}\textbf{Adapted MWGAN}}              & \textbf{GCN} & 0.0017                                 & 0.0332                                 & 0.0045                                 & {\color[HTML]{ff0000} \textbf{0.0052}}    & 0.0063                                 & 0.0002                                 \\ \hline
\rowcolor[HTML]{DEEBF7} 
\cellcolor[HTML]{DEEBF7}                                                      & \textbf{GAT} & {\color[HTML]{00B0F0} \textbf{0.0022}} & {\color[HTML]{00B0F0} \textbf{0.0362}} & 0.0052                                 & 0.021                                  & {\color[HTML]{00B0F0} \textbf{0.0077}} & 0.0003                                 \\ \cline{2-8} 
\rowcolor[HTML]{DEEBF7} 
\multirow{-2}{*}{\cellcolor[HTML]{DEEBF7}\textbf{Adapted MWGAN (clustering)}} & \textbf{GCN} & {\color[HTML]{ff0000} \textbf{0.0007}}    & 0.0147                                 & {\color[HTML]{ff0000} \textbf{0.0009}}    & 0.0088                                 & 0.0024                                 & 0,000014                               \\ \hline
\rowcolor[HTML]{FBE5D6} 
\cellcolor[HTML]{FBE5D6}                                                      & \textbf{GAT} & 0.0026                                 & 0.0491                                 & 0.0063                                 & 0.0431                                 & 0.0082                                 & 0.0002                                 \\ \cline{2-8} 
\rowcolor[HTML]{FBE5D6} 
\multirow{-2}{*}{\cellcolor[HTML]{FBE5D6}\textbf{topoGAN+CC}}                 & \textbf{GCN} & 0.0012                                 & 0.0282                                 & 0.0015                                 & 0.0064                                 & 0.0064                                 & 0,000047                               \\ \hline
\rowcolor[HTML]{E2F0D9} 
\cellcolor[HTML]{E2F0D9}                                                      & \textbf{GAT} & 0.0024                                 & 0.0508                                 & {\color[HTML]{00B0F0} \textbf{0.0045}} & 0.0123                                 & 0.0086                                 & {\color[HTML]{00B0F0} \textbf{0.0001}} \\ \cline{2-8} 
\rowcolor[HTML]{E2F0D9} 
\multirow{-2}{*}{\cellcolor[HTML]{E2F0D9}\textbf{topoGAN+BC}}                 & \textbf{GCN} & 0.0012                                 & 0.0265                                 & 0.0015                                 & 0.0065                                 & 0.0057                                 & 0,000035                               \\ \hline
\rowcolor[HTML]{F5D4FF} 
\cellcolor[HTML]{F5D4FF}                                                      & \textbf{GAT} & 0.0024                                 & 0.0409                                 & 0.0047                                 & {\color[HTML]{00B0F0} \textbf{0.0105}} & 0.0092                                 & 0.0003                                 \\ \cline{2-8} 
\rowcolor[HTML]{F5D4FF} 
\multirow{-2}{*}{\cellcolor[HTML]{F5D4FF}\textbf{topoGAN+EC}}                 & \textbf{GCN} & 0.0008                                 & {\color[HTML]{ff0000} \textbf{0.012}}     & {\color[HTML]{ff0000} \textbf{0.0009}}    & 0.0113                                 & {\color[HTML]{ff0000} \textbf{0.0016}}    & {\color[HTML]{ff0000} \textbf{0,0000098}} \\ \hline
\rowcolor[HTML]{FFD8CE} 
\textbf{MultiGraphGAN+CC}                                                     & \textbf{GCN} & 0.0015                                 & 0.0247                                 & 0.0021                                 & 0.0068                                 & 0.0054                                 & 0,000048                               \\ \hline
\rowcolor[HTML]{F6F9D4} 
\textbf{MultiGraphGAN+BC}                                                     & \textbf{GCN} & 0.0017                                 & 0.0266                                 & 0.0013                                 & 0.0061                                 & 0.0037                                 & 0,000024                               \\ \hline
\rowcolor[HTML]{DEDCE6} 
\textbf{MultiGraphGAN+EC}                                                     & \textbf{GCN} & 0.0019                                 & 0.0465                                 & 0.0028                                 & 0.0079                                 & 0.0103                                 & 0,000097                               \\ \hline

%\multicolumn{8}{p{540}}{View 1: maximum principal curvature (LH). View 2: average curvature (LH). View 3: mean sulcal depth (LH). MAE: mean absolute error. CC: closeness centrality. BC: betweenness centrality. EC: eigenvector centrality. PC: PageRank centrality. EFF: effective size. Clst: clustering coefficient. We highlight in red and blue colors the lowest KL-distance resulting from a particular evaluation metric for the GCN and GAT versions of topoGAN, respectively.}
\end{tabular}
\begin{tablenotes}[para,flushleft]
      \footnotesize
      \item View 1: maximum principal curvature (LH). View 2: average curvature (LH). View 3: mean sulcal depth (LH). MAE: mean absolute error. CC: closeness centrality. BC: betweenness centrality. EC: eigenvector centrality. PC: PageRank centrality. EFF: effective size. Clst: clustering coefficient. We highlight in red and blue colors the lowest KL-distance resulting from a particular evaluation metric for the GCN and GAT versions of topoGAN, respectively.
\end{tablenotes}
\end{threeparttable}
\label{KL1}
\end{table}

\begin{table}
\begin{threeparttable}
\centering
\captionsetup{justification=centering}
\caption{Prediction results for three source views derived from the right hemisphere using KL-divergence evaluation metric.} %}
\footnotesize
\begin{tabular}{|c|c|c|c|c|c|c|c|}
\hline
\rowcolor[HTML]{FFFFFF} 
\multicolumn{2}{|c|}{\cellcolor[HTML]{FFFFFF}\textbf{View 4}}                                   & \textbf{CC}                            & \textbf{BC}                            & \textbf{EC}                            & \textbf{PC}                            & \textbf{EFF}                           & \textbf{Clst}                          \\ \hline
\rowcolor[HTML]{FFF2CC} 
\cellcolor[HTML]{FFF2CC}                                                      & \textbf{GAT} & 0.0028                                 & 0.1416                                 & 0.0268                                 & 0.0661                                 & 0.0223                                 & 0.0062                                 \\ \cline{2-8} 
\rowcolor[HTML]{FFF2CC} 
\multirow{-2}{*}{\cellcolor[HTML]{FFF2CC}\textbf{Adapted MWGAN}}              & \textbf{GCN} & 0.0021                                 & 0.0552                                 & 0.0075                                 & {\color[HTML]{ff0000} \textbf{0.0102}}    & 0.0095                                 & 0.0003                                 \\ \hline
\rowcolor[HTML]{DEEBF7} 
\cellcolor[HTML]{DEEBF7}                                                      & \textbf{GAT} & 0.0029                                 & 0.0773                                 & 0.0085                                 & 0.0766                                 & 0.0165                                 & 0.0005                                 \\ \cline{2-8} 
\rowcolor[HTML]{DEEBF7} 
\multirow{-2}{*}{\cellcolor[HTML]{DEEBF7}\textbf{Adapted MWGAN (clustering)}} & \textbf{GCN} & 0.0008                                 & 0.0167                                 & 0.001                                  & 0.0188                                 & {\color[HTML]{ff0000} \textbf{0.002}}     & 0,000009                               \\ \hline
\rowcolor[HTML]{FBE5D6} 
\cellcolor[HTML]{FBE5D6}                                                      & \textbf{GAT} & 0.0023                                 & 0.061                                  & 0.0052                                 & 0.0455                                 & 0.0115                                 & 0.0004                                 \\ \cline{2-8} 
\rowcolor[HTML]{FBE5D6} 
\multirow{-2}{*}{\cellcolor[HTML]{FBE5D6}\textbf{topoGAN+CC}}                 & \textbf{GCN} & 0.0014                                 & 0.034                                  & 0.0018                                 & 0.0235                                 & 0.0055                                 & 0,00003                                \\ \hline
\rowcolor[HTML]{E2F0D9} 
\cellcolor[HTML]{E2F0D9}                                                      & \textbf{GAT} & 0.0025                                 & {\color[HTML]{00B0F0} \textbf{0.0417}} & 0.0049                                 & {\color[HTML]{00B0F0} \textbf{0.0178}} & {\color[HTML]{00B0F0} \textbf{0.0067}} & {\color[HTML]{00B0F0} \textbf{0.0002}} \\ \cline{2-8} 
\rowcolor[HTML]{E2F0D9} 
\multirow{-2}{*}{\cellcolor[HTML]{E2F0D9}\textbf{topoGAN+BC}}                 & \textbf{GCN} & 0.0015                                 & 0.0349                                 & 0.0019                                 & 0.0198                                 & 0.005                                  & 0,000022                               \\ \hline
\rowcolor[HTML]{F5D4FF} 
\cellcolor[HTML]{F5D4FF}                                                      & \textbf{GAT} & {\color[HTML]{00B0F0} \textbf{0.0021}} & 0.0421                                 & {\color[HTML]{00B0F0} \textbf{0.0042}} & 0.034                                  & 0.0075                                 & {\color[HTML]{00B0F0} \textbf{0.0002}} \\ \cline{2-8} 
\rowcolor[HTML]{F5D4FF} 
\multirow{-2}{*}{\cellcolor[HTML]{F5D4FF}\textbf{topoGAN+EC}}                 & \textbf{GCN} & 0.0015                                 & 0.0418                                 & 0.0018                                 & 0.0236                                 & 0.006                                  & 0,000026                               \\ \hline
\rowcolor[HTML]{FFD8CE} 
\textbf{MultiGraphGAN+CC}                                                     & \textbf{GCN} & {\color[HTML]{ff0000} \textbf{0.0006}}    & {\color[HTML]{ff0000} \textbf{0.0117}}    & {\color[HTML]{ff0000} \textbf{0.0007}}    & 0.0185                                 & 0.0014                                 & {\color[HTML]{ff0000} \textbf{0,000008}}  \\ \hline
\rowcolor[HTML]{F6F9D4} 
\textbf{MultiGraphGAN+BC}                                                     & \textbf{GCN} & 0.0019                                 & 0.0461                                 & 0.0029                                 & 0.0267                                 & 0.0088                                 & 0,000085                               \\ \hline
\rowcolor[HTML]{DEDCE6} 
\textbf{MultiGraphGAN+EC}                                                     & \textbf{GCN} & 0.0008                                 & 0.0153                                 & 0.001                                  & 0.0255                                 & 0.0029                                 & 0,000015                               \\ \hline
\rowcolor[HTML]{FFFFFF} 
\multicolumn{2}{|c|}{\cellcolor[HTML]{FFFFFF}\textbf{View 5}}                                   & \textbf{CC}                            & \textbf{BC}                            & \textbf{EC}                            & \textbf{PC}                            & \textbf{EFF}                           & \textbf{Clst}                          \\ \hline
\rowcolor[HTML]{FFF2CC} 
\cellcolor[HTML]{FFF2CC}                                                      & \textbf{GAT} & 0.0034                                 & 0.1783                                 & 0.023                                  & 0.0426                                 & 0.0337                                 & 0.0045                                 \\ \cline{2-8} 
\rowcolor[HTML]{FFF2CC} 
\multirow{-2}{*}{\cellcolor[HTML]{FFF2CC}\textbf{Adapted MWGAN}}              & \textbf{GCN} & 0.0024                                 & 0.0688                                 & 0.0116                                 & 0.0195                                 & 0.0123                                 & 0.0015                                 \\ \hline
\rowcolor[HTML]{DEEBF7} 
\cellcolor[HTML]{DEEBF7}                                                      & \textbf{GAT} & 0.0032                                 & 0.0543                                 & 0.0066                                 & {\color[HTML]{00B0F0} \textbf{0.0074}} & 0.0116                                 & 0.0003                                 \\ \cline{2-8} 
\rowcolor[HTML]{DEEBF7} 
\multirow{-2}{*}{\cellcolor[HTML]{DEEBF7}\textbf{Adapted MWGAN (clustering)}} & \textbf{GCN} & 0.0025                                 & {\color[HTML]{ff0000} \textbf{0.0386}}    & 0.0056                                 & 0.0198                                 & {\color[HTML]{ff0000} \textbf{0.006}}     & {\color[HTML]{ff0000} \textbf{0.0003}}    \\ \hline
\rowcolor[HTML]{FBE5D6} 
\cellcolor[HTML]{FBE5D6}                                                      & \textbf{GAT} & 0.0021                                 & {\color[HTML]{00B0F0} \textbf{0.0374}} & 0.0042                                 & 0.0154                                 & 0.0087                                 & 0.0002                                 \\ \cline{2-8} 
\rowcolor[HTML]{FBE5D6} 
\multirow{-2}{*}{\cellcolor[HTML]{FBE5D6}\textbf{topoGAN+CC}}                 & \textbf{GCN} & 0.0026                                 & 0.052                                  & 0.0065                                 & {\color[HTML]{ff0000} \textbf{0.018}}     & 0.0116                                 & 0.0005                                 \\ \hline
\rowcolor[HTML]{E2F0D9} 
\cellcolor[HTML]{E2F0D9}                                                      & \textbf{GAT} & 0.0036                                 & 0.0738                                 & 0.0105                                 & 0.014                                  & 0.0156                                 & 0.0006                                 \\ \cline{2-8} 
\rowcolor[HTML]{E2F0D9} 
\multirow{-2}{*}{\cellcolor[HTML]{E2F0D9}\textbf{topoGAN+BC}}                 & \textbf{GCN} & 0.0035                                 & 0.0725                                 & 0.01                                   & 0.0332                                 & 0.0157                                 & 0.0005                                 \\ \hline
\rowcolor[HTML]{F5D4FF} 
\cellcolor[HTML]{F5D4FF}                                                      & \textbf{GAT} & {\color[HTML]{00B0F0} \textbf{0.0013}} & 0.0405                                 & {\color[HTML]{00B0F0} \textbf{0.0026}} & 0.0111                                 & {\color[HTML]{00B0F0} \textbf{0.0065}} & {\color[HTML]{00B0F0} \textbf{0.0001}} \\ \cline{2-8} 
\rowcolor[HTML]{F5D4FF} 
\multirow{-2}{*}{\cellcolor[HTML]{F5D4FF}\textbf{topoGAN+EC}}                 & \textbf{GCN} & 0.0035                                 & 0.0845                                 & 0.0114                                 & 0.0253                                 & 0.0171                                 & 0.001                                  \\ \hline
\rowcolor[HTML]{FFD8CE} 
\textbf{MultiGraphGAN+CC}                                                     & \textbf{GCN} & {\color[HTML]{ff0000} \textbf{0.0022}}    & 0.0431                                 & {\color[HTML]{ff0000} \textbf{0.0052}}    & 0.0244                                 & 0.0069                                 & {\color[HTML]{ff0000} \textbf{0.0003}}    \\ \hline
\rowcolor[HTML]{F6F9D4} 
\textbf{MultiGraphGAN+BC}                                                     & \textbf{GCN} & 0.0038                                 & 0.1026                                 & 0.015                                  & 0.0232                                 & 0.0185                                 & 0.0012                                 \\ \hline
\rowcolor[HTML]{DEDCE6} 
\textbf{MultiGraphGAN+EC}                                                     & \textbf{GCN} & 0.0036                                 & 0.0746                                 & 0.0099                                 & 0.0295                                 & 0.0175                                 & 0.0008                                 \\ \hline
\rowcolor[HTML]{FFFFFF} 
\multicolumn{2}{|c|}{\cellcolor[HTML]{FFFFFF}\textbf{View 6}}                                   & \textbf{CC}                            & \textbf{BC}                            & \textbf{EC}                            & \textbf{PC}                            & \textbf{EFF}                           & \textbf{Clst}                          \\ \hline
\rowcolor[HTML]{FFF2CC} 
\cellcolor[HTML]{FFF2CC}                                                      & \textbf{GAT} & 0.0027                                 & 0.0883                                 & 0.0117                                 & {\color[HTML]{00B0F0} \textbf{0.0099}} & 0.018                                  & 0.0011                                 \\ \cline{2-8} 
\rowcolor[HTML]{FFF2CC} 
\multirow{-2}{*}{\cellcolor[HTML]{FFF2CC}\textbf{Adapted MWGAN}}              & \textbf{GCN} & {\color[HTML]{ff0000} \textbf{0.0014}}    & 0.0791                                 & 0.0095                                 & 0.026                                  & 0.0169                                 & 0.002                                  \\ \hline
\rowcolor[HTML]{DEEBF7} 
\cellcolor[HTML]{DEEBF7}                                                      & \textbf{GAT} & 0.003                                  & 0.0708                                 & 0.0098                                 & 0.0414                                 & 0.0122                                 & 0.0006                                 \\ \cline{2-8} 
\rowcolor[HTML]{DEEBF7} 
\multirow{-2}{*}{\cellcolor[HTML]{DEEBF7}\textbf{Adapted MWGAN (clustering)}} & \textbf{GCN} & 0.0018                                 & 0.0257                                 & 0.004                                  & 0.031                                  & {\color[HTML]{ff0000} \textbf{0.0045}}    & 0.0002                                 \\ \hline
\rowcolor[HTML]{FBE5D6} 
\cellcolor[HTML]{FBE5D6}                                                      & \textbf{GAT} & 0.0026                                 & {\color[HTML]{00B0F0} \textbf{0.0386}} & 0.0068                                 & 0.0314                                 & {\color[HTML]{00B0F0} \textbf{0.006}}  & 0.0003                                 \\ \cline{2-8} 
\rowcolor[HTML]{FBE5D6} 
\multirow{-2}{*}{\cellcolor[HTML]{FBE5D6}\textbf{topoGAN+CC}}                 & \textbf{GCN} & 0.0017                                 & 0.037                                  & 0.0026                                 & 0.0202                                 & 0.0065                                 & {\color[HTML]{ff0000} \textbf{0.0001}}    \\ \hline
\rowcolor[HTML]{E2F0D9} 
\cellcolor[HTML]{E2F0D9}                                                      & \textbf{GAT} & {\color[HTML]{00B0F0} \textbf{0.0017}} & 0.0416                                 & {\color[HTML]{00B0F0} \textbf{0.0034}} & 0.0581                                 & 0.0076                                 & {\color[HTML]{00B0F0} \textbf{0.0002}} \\ \cline{2-8} 
\rowcolor[HTML]{E2F0D9} 
\multirow{-2}{*}{\cellcolor[HTML]{E2F0D9}\textbf{topoGAN+BC}}                 & \textbf{GCN} & 0.0019                                 & 0.0295                                 & 0.0028                                 & {\color[HTML]{ff0000} \textbf{0.0189}}    & 0.0065                                 & {\color[HTML]{ff0000} \textbf{0.0001}}    \\ \hline
\rowcolor[HTML]{F5D4FF} 
\cellcolor[HTML]{F5D4FF}                                                      & \textbf{GAT} & 0.0026                                 & 0.0481                                 & 0.0054                                 & 0.0304                                 & 0.0096                                 & 0.0003                                 \\ \cline{2-8} 
\rowcolor[HTML]{F5D4FF} 
\multirow{-2}{*}{\cellcolor[HTML]{F5D4FF}\textbf{topoGAN+EC}}                 & \textbf{GCN} & 0.0015                                 & {\color[HTML]{ff0000} \textbf{0.0254}}    & {\color[HTML]{ff0000} \textbf{0.0022}}    & 0.027                                  & 0.0052                                 & {\color[HTML]{ff0000} \textbf{0.0001}}    \\ \hline
\rowcolor[HTML]{FFD8CE} 
\textbf{MultiGraphGAN+CC}                                                     & \textbf{GCN} & 0.0027                                 & 0.0491                                 & 0.0066                                 & 0.0426                                 & 0.0077                                 & 0.0003                                 \\ \hline
\rowcolor[HTML]{F6F9D4} 
\textbf{MultiGraphGAN+BC}                                                     & \textbf{GCN} & 0.0018                                 & 0.029                                  & 0.0056                                 & 0.0342                                 & 0.0075                                 & 0.0003                                 \\ \hline
\rowcolor[HTML]{DEDCE6} 
\textbf{MultiGraphGAN+EC}                                                     & \textbf{GCN} & 0.0026                                 & 0.0548                                 & 0.0043                                 & 0.0351                                 & 0.0124                                 & 0.0002                                 \\ \hline

%\multicolumn{8}{p{540}}{View 4: maximum principal curvature (RH). View 5: average curvature (RH). View 6: mean sulcal depth (RH). MAE: mean absolute error. CC: closeness centrality. BC: betweenness centrality. EC: eigenvector centrality. PC: PageRank centrality. EFF: effective size. Clst: clustering coefficient. We highlight in red and blue colors the lowest KL-distance resulting from a particular evaluation metric for the GCN and GAT versions of topoGAN, respectively.}
\end{tabular}
\begin{tablenotes}[para,flushleft]
      \footnotesize
      \item View 4: maximum principal curvature (RH). View 5: average curvature (RH). View 6: mean sulcal depth (RH). MAE: mean absolute error. CC: closeness centrality. BC: betweenness centrality. EC: eigenvector centrality. PC: PageRank centrality. EFF: effective size. Clst: clustering coefficient. We highlight in red and blue colors the lowest KL-distance resulting from a particular evaluation metric for the GCN and GAT versions of topoGAN, respectively.
\end{tablenotes}
\end{threeparttable}
\label{KL2}
\end{table}

\subsection{Discussion and future recommendations} 

In this paper, we introduced the first geometric deep learning framework designed for jointly predicting multiple brain graphs from a single brain graph. Our model (i) is a novel graph adversarial auto-encoder that includes an encoder and a set of decoders (i.e., generators) all regularized by a single discriminator, (ii) clusters the source embeddings and defines a set of \emph{cluster-specific decoder} to overcome the mode collapse issue of GAN, (iii) includes a \emph{topological loss} to preserve both global and local topological properties of the original graphs. For the first time, we take the connectomics field one step further into predicting missing brain graphs from existing minimal resources (i.e., a single brain graph). Our topoGAN achieved significantly better performances than the ablated versions and the state-of-the-art method --namely MultiGraphGAN \citep{bessadok:2020}. Moreover, it consistently outperformed all comparison methods not only when evaluated using random split (i.e., 90/10\%) but also when using cross-validation strategy. Although the proposed target brain multigraph prediction framework is generic and can also be applied to any type of brain graph (e.g, functional or structural), there are still some issues that can affect its performance. One major issue is that the more clusters we introduce, the more easily the model can overfit the training set. This might be circumvented by monitoring the training and testing loss functions \citep{rice2020overfitting}. % Thus, we aim in our future work to leverage “Elbow” method which is widely used for selecting the best number of clusters.

A second major issue is considering a single discriminator to regularize the encoder and our set of \emph{clustering-specific} generators while recent works demonstrate that multiple discriminators improve the learning process even using small datasets \citep{Durugkar:2016,Neyshabur:2017}. Hence, we will improve the cluster-specific multi-target graph prediction step (\textbf{Fig.}~\ref{2}-C) by leveraging a recent multi-objective optimization framework \citep{Albuquerque:2019} that ensures an accurate data generation using multiple adversarial regularizers. However, this method was originally designed for image generation so we aim to adopt it to geometric data. To further improve the GCN learning on brain graphs, we aim to use variational GCN \citep{Tiao} in combination with a recent adversarial graph embedding technique \citep{Pan:2018}. We hypothesize that such architecture improvement will boost the source graph embedding. As another research direction, we will evaluate our framework using a larger dataset including functional and structural brain graphs. Since in this work we only focused on graph synthesis task we aim in the future to work on early disease identification. Essentially, we will combine both graph prediction and disease classification tasks in a single and unified geometric deep learning framework trained in an end-to-end manner which will be used for diagnosing different neurological disorders. Ultimately, this model can serve as a stepping stone to develop more holistic brain prediction models such as predicting spatiotempral trajectory \citep{Ghribi:2019,Vohryzek:2020} and super-resolution brain graphs \citep{Cengiz:2019,Mhiri:2020b}.

% %% ***************************************************************************** %%
\section{Conclusion}
% %% ***************************************************************************** %%
Very few models exist for predicting a single target brain graph from a source graph (i.e, one-to-one prediction task). This is a recently emerging field with high-level meaningful implications in neurological disorder diagnosis. We presented in this paper the first geometric deep learning framework, namely topoGAN, for jointly predicting multiple brain views represented by a \emph{target multigraph} from a single source graph both derived from MRI (i.e, one-to-many prediction task). Our architecture has two compelling strengths: (i) clustering the learned source graphs embeddings then training a set of \emph{cluster-specific generators} which synergistically predict the target brain graphs, (ii) introducing a \emph{topological loss function} using a centrality measure which enforces the generators to preserve both local and global topologies of the original target graphs. Our proposed brain multigraph prediction framework can be further tailored to predict the evolution of target brain multigraph over time from a single source brain graph \citep{Ghribi:2019,Vohryzek:2020}. Such framework harbors a powerful tool to examine how alterations in the connectome may lead to a progressive brain dysfunction in neurological disorder. Eventually, we envision to evaluate our framework on larger connectomic datasets and cover a diverse range of brain graphs such structural and functional networks \citep{Wen:2017,Mhiri:2020}.

% %% ***************************************************************************** %%
\section{Acknowledgements}
% %% ***************************************************************************** %%

This work was funded by generous grants from the European H2020 Marie Sklodowska-Curie action (grant no. 101003403, \url{http://basira-lab.com/normnets/}) to I.R. and the Scientific and Technological Research Council of Turkey to I.R. under the TUBITAK 2232 Fellowship for Outstanding Researchers (no. 118C288, \url{http://basira-lab.com/reprime/}). However, all scientific contributions made in this project are owned and approved solely by the authors. A.B. is supported by the same the TUBITAK 2232 Fellowship for Outstanding Researchers.

%%%%%%%%%%%%%%%%%%%%%%%%%%%%%%%%%%%%%%%%%%%%%%%%%%%%%%%%%%%%%%%%%%%%%%%%%%%%%%%%%%%%%%%%%%%%%%%%%%%%%%%%%%%%
\newpage

\begin{table}
\begin{threeparttable}
\centering
\vspace{-30pt}
\captionsetup{justification=centering}
\caption{P-value results using two-tailed paired t-test for predicting a target brain multigraph using two source views derived from the left hemisphere.}% }
\scriptsize
\begin{tabular}{|c|c|c|c|c|}
\hline
\multicolumn{2}{|c|}{\cellcolor[HTML]{FFFFFF}}                                                  & \cellcolor[HTML]{FCE4D6}                                      & \cellcolor[HTML]{E2EFDA}                                      & \cellcolor[HTML]{F5D4FF}                                      \\
\multicolumn{2}{|c|}{\multirow{-2}{*}{\cellcolor[HTML]{FFFFFF}\textbf{View 1}}}                 & \multirow{-2}{*}{\cellcolor[HTML]{FCE4D6}\textbf{topoGAN+CC}} & \multirow{-2}{*}{\cellcolor[HTML]{E2EFDA}\textbf{topoGAN+BC}} & \multirow{-2}{*}{\cellcolor[HTML]{F5D4FF}\textbf{topoGAN+EC}} \\ \hline
\rowcolor[HTML]{FFF2CC} 
\cellcolor[HTML]{FFF2CC}                                                      & \textbf{View 1} & \textbf{1.1661}$\times10^{-10}$                               & \textbf{1.7835}$\times 10^{-5} $                         & \textbf{4.8519} $\times10^{-10}$                              \\ \cline{2-5} 
\rowcolor[HTML]{FFF2CC} 
\cellcolor[HTML]{FFF2CC}                                                      & \textbf{View 2} & \textbf{0.0354}                                               & 0.0911                                                        & 0.1138                                                        \\ \cline{2-5} 
\rowcolor[HTML]{FFF2CC} 
\cellcolor[HTML]{FFF2CC}                                                      & \textbf{View 3} & 0.1259                                                        & \textbf{0.0321}                                               & 0.0516                                               \\ \cline{2-5} 
\rowcolor[HTML]{FFF2CC} 
\cellcolor[HTML]{FFF2CC}                                                      & \textbf{View 4} & \textbf{9.8979} $\times10^{-7}$                               & \textbf{0.0404}                                               & 0.0717                                                        \\ \cline{2-5} 
\rowcolor[HTML]{FFF2CC} 
\multirow{-5}{*}{\cellcolor[HTML]{FFF2CC}\textbf{Adapted MWGAN}}              & \textbf{View 5} & \textbf{0.0369}                                               & 0.0694                                                        & 0.0808                                                        \\ \hline
\rowcolor[HTML]{DDEBF7} 
\cellcolor[HTML]{DDEBF7}                                                      & \textbf{View 1} & \textbf{0.016}                                                & 0.0924                                                        & 0.1008                                                        \\ \cline{2-5} 
\rowcolor[HTML]{DDEBF7} 
\cellcolor[HTML]{DDEBF7}                                                      & \textbf{View 2} & 0.0643                                                        & 0.0679                                                        & \textbf{0.0281}                                               \\ \cline{2-5} 
\rowcolor[HTML]{DDEBF7} 
\cellcolor[HTML]{DDEBF7}                                                      & \textbf{View 3} & \textbf{0.0146}                                               & \textbf{0.0007}                                               & \textbf{0.0014}                                               \\ \cline{2-5} 
\rowcolor[HTML]{DDEBF7} 
\cellcolor[HTML]{DDEBF7}                                                      & \textbf{View 4} & \textbf{7.6946} $\times10^{-7}$                               & \textbf{0.0045}                                               & 0.0553                                                        \\ \cline{2-5} 
\rowcolor[HTML]{DDEBF7} 
\multirow{-5}{*}{\cellcolor[HTML]{DDEBF7}\textbf{Adapted MWGAN (clustering)}} & \textbf{View 5} & \textbf{0.0014}                                               & 0.1216                                                        & \textbf{0.0454}                                               \\ \hline

\rowcolor[HTML]{FFD8CE} 
\cellcolor[HTML]{FFD8CE}                                            & \textbf{View 1} & \textbf{0.0001}         & \textbf{0.0012}     & \textbf{0.0023 }         \\ \cline{2-5} 
\rowcolor[HTML]{FFD8CE} 
\cellcolor[HTML]{FFD8CE}                                            & \textbf{View 2} & \textbf{2.3} $\times10^{-20}$       & 0.0001     & 0.0041          \\ \cline{2-5} 
\rowcolor[HTML]{FFD8CE} 
\cellcolor[HTML]{FFD8CE}                                            & \textbf{View 3} & \textbf{0.0299 }        & \textbf{2.7} $\times10^{-7}$ & 0.0018          \\ \cline{2-5} 
\rowcolor[HTML]{FFD8CE} 
\cellcolor[HTML]{FFD8CE}                                            & \textbf{View 4} & \textbf{0.0403}         & \textbf{0.0011}     & \textbf{2.02} $\times10^{-8}$     \\ \cline{2-5} 
\rowcolor[HTML]{FFD8CE} 
\multirow{-5}{*}{\cellcolor[HTML]{FFD8CE}\textbf{MultiGraphGAN+CC}} & \textbf{View 5} & \textbf{0.0003}         & \textbf{2.7} $\times10^{-7}$ & \textbf{2.5} $\times10^{-12}$ \\ \hline
\rowcolor[HTML]{F6F9D4} 
\cellcolor[HTML]{F6F9D4}                                            & \textbf{View 1} & 0.0789         & 0.0649     & \textbf{0.0306 }         \\ \cline{2-5} 
\rowcolor[HTML]{F6F9D4} 
\cellcolor[HTML]{F6F9D4}                                            & \textbf{View 2} & 0.0976         & \textbf{0.0245}     & \textbf{0.0002}          \\ \cline{2-5} 
\rowcolor[HTML]{F6F9D4} 
\cellcolor[HTML]{F6F9D4}                                            & \textbf{View 3} & 0.086          & \textbf{0.0015 }    & \textbf{0.0002 }         \\ \cline{2-5} 
\rowcolor[HTML]{F6F9D4} 
\cellcolor[HTML]{F6F9D4}                                            & \textbf{View 4} & 0.0599         & \textbf{0.0115 }    & \textbf{0.0211  }        \\ \cline{2-5} 
\rowcolor[HTML]{F6F9D4} 
\multirow{-5}{*}{\cellcolor[HTML]{F6F9D4}\textbf{MultiGraphGAN+BC}} & \textbf{View 5} & 0.0805         & 0.0549     & \textbf{0.0102  }        \\ \hline
\rowcolor[HTML]{DEDCE6} 
\cellcolor[HTML]{DEDCE6}                                            & \textbf{View 1} & 0.0679         & 0.1076     & \textbf{0.0059 }         \\ \cline{2-5} 
\rowcolor[HTML]{DEDCE6} 
\cellcolor[HTML]{DEDCE6}                                            & \textbf{View 2} & \textbf{1.4} $\times10^{-7}$     & 0.1027     & 0.0723          \\ \cline{2-5} 
\rowcolor[HTML]{DEDCE6} 
\cellcolor[HTML]{DEDCE6}                                            & \textbf{View 3} & \textbf{0.0177 }        & 0.0917     & \textbf{0.0264  }        \\ \cline{2-5} 
\rowcolor[HTML]{DEDCE6} 
\cellcolor[HTML]{DEDCE6}                                            & \textbf{View 4} & \textbf{8.5} $\times10^{-11}$ & \textbf{0.0054 }    & 0.0609          \\ \cline{2-5} 
\rowcolor[HTML]{DEDCE6} 
\multirow{-5}{*}{\cellcolor[HTML]{DEDCE6}\textbf{MultiGraphGAN+EC}} & \textbf{View 5} & \textbf{0.0161  }       & 0.0548     & 0.086           \\ \hline

\multicolumn{2}{|c|}{\cellcolor[HTML]{FFFFFF}}                                                  & \cellcolor[HTML]{FCE4D6}                                      & \cellcolor[HTML]{E2EFDA}                                      & \cellcolor[HTML]{F5D4FF}                                      \\
\multicolumn{2}{|c|}{\multirow{-2}{*}{\cellcolor[HTML]{FFFFFF}\textbf{View 2}}}                 & \multirow{-2}{*}{\cellcolor[HTML]{FCE4D6}\textbf{topoGAN+CC}} & \multirow{-2}{*}{\cellcolor[HTML]{E2EFDA}\textbf{topoGAN+BC}} & \multirow{-2}{*}{\cellcolor[HTML]{F5D4FF}\textbf{topoGAN+EC}} \\ \hline
\rowcolor[HTML]{FFF2CC} 
\cellcolor[HTML]{FFF2CC}                                                      & \textbf{View 1} & \textbf{1.7883} $\times10^{-15}$                              & \textbf{8.9525}$\times10^{-12}$                               & \textbf{0.0008}                                               \\ \cline{2-5} 
\rowcolor[HTML]{FFF2CC} 
\cellcolor[HTML]{FFF2CC}                                                      & \textbf{View 2} & \textbf{2.4751}$\times10^{-15}$                               & \textbf{7.0026} $\times10^{-10}$                             & \textbf{0.0007}                                               \\ \cline{2-5} 
\rowcolor[HTML]{FFF2CC} 
\cellcolor[HTML]{FFF2CC}                                                      & \textbf{View 3} & \textbf{4.6296}$\times10^{-20}$                            & \textbf{9.9692} $\times10^{-20}$                              & \textbf{9.9618} $\times10^{-11}$                              \\ \cline{2-5} 
\rowcolor[HTML]{FFF2CC} 
\cellcolor[HTML]{FFF2CC}                                                      & \textbf{View 4} & \textbf{4.6198}$\times10^{-7}$                             & 0.0603                                                        & 0.0567                                                        \\ \cline{2-5} 
\rowcolor[HTML]{FFF2CC} 
\multirow{-5}{*}{\cellcolor[HTML]{FFF2CC}\textbf{Adapted MWGAN}}              & \textbf{View 5} & \textbf{1.0612}$\times10^{-12}$                           & \textbf{0.0001}                                               & \textbf{0.0003}                                               \\ \hline
\rowcolor[HTML]{DDEBF7} 
\cellcolor[HTML]{DDEBF7}                                                      & \textbf{View 1} & 0.2857                                                        & 0.1867                                                        & \textbf{0.0294}                                               \\ \cline{2-5} 
\rowcolor[HTML]{DDEBF7} 
\cellcolor[HTML]{DDEBF7}                                                      & \textbf{View 2} & \textbf{0.0433}                                               & 0.2578                                                        & 0.1832                                                        \\ \cline{2-5} 
\rowcolor[HTML]{DDEBF7} 
\cellcolor[HTML]{DDEBF7}                                                      & \textbf{View 3} & 0.0599                                                        & 0.0553                                                        & 0.4429                                                        \\ \cline{2-5} 
\rowcolor[HTML]{DDEBF7} 
\cellcolor[HTML]{DDEBF7}                                                      & \textbf{View 4} & 0.4283                                                        & \textbf{0.0005}                                               & \textbf{0.0292}                                               \\ \cline{2-5} 
\rowcolor[HTML]{DDEBF7} 
\multirow{-5}{*}{\cellcolor[HTML]{DDEBF7}\textbf{Adapted MWGAN (clustering)}} & \textbf{View 5} & 0.1386                                                        & \textbf{0.0361}                                               & 0.2063                                                        \\ \hline
                         
\rowcolor[HTML]{FFD8CE} 
\cellcolor[HTML]{FFD8CE}                                            & \textbf{View 1} & 0.1786   & 0.3806   & \textbf{0.0246}  \\ \cline{2-5} 
\rowcolor[HTML]{FFD8CE} 
\cellcolor[HTML]{FFD8CE}                                            & \textbf{View 2} & \textbf{0.0361}   & 0.1927   & 0.263   \\ \cline{2-5} 
\rowcolor[HTML]{FFD8CE} 
\cellcolor[HTML]{FFD8CE}                                            & \textbf{View 3} & 0.4158   & 0.4336   & 0.1761  \\ \cline{2-5} 
\rowcolor[HTML]{FFD8CE} 
\cellcolor[HTML]{FFD8CE}                                            & \textbf{View 4} & 0.14     & \textbf{0.0002}   & \textbf{0.0003}  \\ \cline{2-5} 
\rowcolor[HTML]{FFD8CE} 
\multirow{-5}{*}{\cellcolor[HTML]{FFD8CE}\textbf{MultiGraphGAN+CC}} & \textbf{View 5} & \textbf{1.01}$\times10^{-5}$  & 0.2049   & 0.1333  \\ \hline
\rowcolor[HTML]{F6F9D4} 
\cellcolor[HTML]{F6F9D4}                                            & \textbf{View 1} & \textbf{0.0129 }  & 0.1223   & 0.3667  \\ \cline{2-5} 
\rowcolor[HTML]{F6F9D4} 
\cellcolor[HTML]{F6F9D4}                                            & \textbf{View 2} & \textbf{7.8}$\times10^{-7}$   & \textbf{0.0359}   & 0.2009  \\ \cline{2-5} 
\rowcolor[HTML]{F6F9D4} 
\cellcolor[HTML]{F6F9D4}                                            & \textbf{View 3} & \textbf{0.0099}   & \textbf{0.0372}   & 0.2489  \\ \cline{2-5} 
\rowcolor[HTML]{F6F9D4} 
\cellcolor[HTML]{F6F9D4}                                            & \textbf{View 4} & \textbf{5.8}$\times10^{-24}$  & \textbf{6.98}$\times10^{-7}$ & \textbf{1.8}$\times10^{-10}$  \\ \cline{2-5} 
\rowcolor[HTML]{F6F9D4} 
\multirow{-5}{*}{\cellcolor[HTML]{F6F9D4}\textbf{MultiGraphGAN+BC}} & \textbf{View 5} & \textbf{0.0184}   & 0.3352   & 0.5021  \\ \hline
\rowcolor[HTML]{DEDCE6} 
\cellcolor[HTML]{DEDCE6}                                            & \textbf{View 1} & 0.1926   & 0.3907   & \textbf{0.0498}  \\ \cline{2-5} 
\rowcolor[HTML]{DEDCE6} 
\cellcolor[HTML]{DEDCE6}                                            & \textbf{View 2} & 0.3544   & 0.3772   & \textbf{0.0046}  \\ \cline{2-5} 
\rowcolor[HTML]{DEDCE6} 
\cellcolor[HTML]{DEDCE6}                                            & \textbf{View 3} & \textbf{0.01}     & \textbf{0.031}    & 0.2065  \\ \cline{2-5} 
\rowcolor[HTML]{DEDCE6} 
\cellcolor[HTML]{DEDCE6}                                            & \textbf{View 4} & \textbf{3.3}$\times10^{-6}$   & \textbf{0.0407}   & \textbf{0.0343}  \\ \cline{2-5} 
\rowcolor[HTML]{DEDCE6} 
\multirow{-5}{*}{\cellcolor[HTML]{DEDCE6}\textbf{MultiGraphGAN+EC}} & \textbf{View 5} & \textbf{0.0014}   & 0.3154   & 0.2411  \\ \hline

%\multicolumn{5}{p{550}}{Adapted MWGAN: the graph-based architecture of the method introduced in \citep{Cao:2019}. Adapted MWGAN (clustering): a variant of the adapted method \citep{Cao:2019} with a clustering step. topoGAN+CC, topoGAN+BC, topoGAN+EC: the proposed method that includes closeness, betweenness and eigenvector. MultiGraphGAN+CC, MultiGraphGAN+BC, MultiGraphGAN+EC: the state-of-the-art method \citep{bessadok:2020} that includes the reconstruction loss and the closeness, betweenness and eigenvector centralities. View 1: maximum principal curvature (LH). View 2: average curvature (LH). We highlight in bold the $p$-value $\prec$ $0.05$ using two-tailed paired $t$-test between row-wise and column-wise methods.}\\
\end{tabular}
\begin{tablenotes}[para,flushleft]
      \footnotesize
      \item Adapted MWGAN: the graph-based architecture of the method introduced in \citep{Cao:2019}. Adapted MWGAN (clustering): a variant of the adapted method \citep{Cao:2019} with a clustering step. topoGAN+CC, topoGAN+BC, topoGAN+EC: the proposed method that includes closeness, betweenness and eigenvector. MultiGraphGAN+CC, MultiGraphGAN+BC, MultiGraphGAN+EC: the state-of-the-art method \citep{bessadok:2020} that includes the reconstruction loss and the closeness, betweenness and eigenvector centralities. View 1: maximum principal curvature (LH). View 2: average curvature (LH). We highlight in bold the $p$-value $\prec$ $0.05$ using two-tailed paired $t$-test between row-wise and column-wise methods.
\end{tablenotes}
\end{threeparttable}
\label{p-value-1}
\end{table}

\begin{table}
\begin{threeparttable}
\centering
\vspace{-30pt}
\captionsetup{justification=centering}
\caption{P-value results using two-tailed paired t-test for predicting a target brain multigraph using six source views.}% }
\scriptsize
\begin{tabular}{|c|c|c|c|c|}
\hline
\multicolumn{2}{|c|}{\cellcolor[HTML]{FFFFFF}}                                                  & \cellcolor[HTML]{FCE4D6}                                      & \cellcolor[HTML]{E2EFDA}                                      & \cellcolor[HTML]{F5D4FF}                                      \\\multicolumn{2}{|c|}{\multirow{-2}{*}{\cellcolor[HTML]{FFFFFF}\textbf{View 3}}}                 & \multirow{-2}{*}{\cellcolor[HTML]{FCE4D6}\textbf{topoGAN+CC}} & \multirow{-2}{*}{\cellcolor[HTML]{E2EFDA}\textbf{topoGAN+BC}} & \multirow{-2}{*}{\cellcolor[HTML]{F5D4FF}\textbf{topoGAN+EC}} \\ \hline
\rowcolor[HTML]{FFF2CC} 
\cellcolor[HTML]{FFF2CC}                                                      & \textbf{View 1} & \textbf{0.0252}                                               & 0.1635                                                        & 0.0961                                                        \\ \cline{2-5} 
\rowcolor[HTML]{FFF2CC} 
\cellcolor[HTML]{FFF2CC}                                                      & \textbf{View 2} & \textbf{4.2923}$\times10^{-13}$                             & \textbf{0.0265}                                               & \textbf{0.005}                                                \\ \cline{2-5} 
\rowcolor[HTML]{FFF2CC} 
\cellcolor[HTML]{FFF2CC}                                                      & \textbf{View 3} & 0.1796                                                        & \textbf{0.0043}                                               & 0.0611                                                        \\ \cline{2-5} 
\rowcolor[HTML]{FFF2CC} 
\cellcolor[HTML]{FFF2CC}                                                      & \textbf{View 4} & \textbf{0.024}                                                & \textbf{0.0006}                                               & \textbf{0.0059}                                               \\ \cline{2-5} 
\rowcolor[HTML]{FFF2CC} 
\multirow{-5}{*}{\cellcolor[HTML]{FFF2CC}\textbf{Adapted MWGAN}}              & \textbf{View 5} & 0.0583                                                        & 0.0547                                                        & \textbf{0.0479}                                               \\ \hline
\rowcolor[HTML]{DDEBF7} 
\cellcolor[HTML]{DDEBF7}                                                      & \textbf{View 1} & \textbf{0.0016}                                               & 0.179                                                         & \textbf{0.0437}                                               \\ \cline{2-5} 
\rowcolor[HTML]{DDEBF7} 
\cellcolor[HTML]{DDEBF7}                                                      & \textbf{View 2} & \textbf{6.1182}$\times10^{-7}$                               & 0.0858                                                        & 0.0527                                                        \\ \cline{2-5} 
\rowcolor[HTML]{DDEBF7} 
\cellcolor[HTML]{DDEBF7}                                                      & \textbf{View 3} & \textbf{0.0217}                                               & 0.091                                                         & \textbf{0.0292}                                               \\ \cline{2-5} 
\rowcolor[HTML]{DDEBF7} 
\cellcolor[HTML]{DDEBF7}                                                      & \textbf{View 4} & \textbf{0.0204}                                               & \textbf{0.0068}                                               & \textbf{0.0317}                                               \\ \cline{2-5} 
\rowcolor[HTML]{DDEBF7} 
\multirow{-5}{*}{\cellcolor[HTML]{DDEBF7}\textbf{Adapted MWGAN (clustering)}} & \textbf{View 5} & \textbf{0.0202}                                               & 0.0595                                                        & \textbf{0.023}                                                \\ \hline

\rowcolor[HTML]{FFD8CE} 
\cellcolor[HTML]{FFD8CE}                                            & \textbf{View 1} & \textbf{0.0001}   & \textbf{0.0012}  & \textbf{0.0023 }  \\ \cline{2-5} 
\rowcolor[HTML]{FFD8CE} 
\cellcolor[HTML]{FFD8CE}                                            & \textbf{View 2} & \textbf{2.3}$\times10^{-20}$  & \textbf{0.0001}  & \textbf{0.0041}   \\ \cline{2-5} 
\rowcolor[HTML]{FFD8CE} 
\cellcolor[HTML]{FFD8CE}                                            & \textbf{View 3} & \textbf{0.0299 }  & \textbf{2.7}$\times10^{-7}$  & \textbf{0.0018}   \\ \cline{2-5} 
\rowcolor[HTML]{FFD8CE} 
\cellcolor[HTML]{FFD8CE}                                            & \textbf{View 4} & \textbf{0.0403 }  & \textbf{0.0011}  & \textbf{2.02}$\times10^{-8}$  \\ \cline{2-5} 
\rowcolor[HTML]{FFD8CE} 
\multirow{-5}{*}{\cellcolor[HTML]{FFD8CE}\textbf{MultiGraphGAN+CC}} & \textbf{View 5} & \textbf{0.0003}   & \textbf{2.7}$\times10^{-7}$  & \textbf{2.5}$\times10^{-12}$   \\ \hline
\rowcolor[HTML]{F6F9D4} 
\cellcolor[HTML]{F6F9D4}                                            & \textbf{View 1} & 0.0789   & 0.0649  & \textbf{0.0306}   \\ \cline{2-5} 
\rowcolor[HTML]{F6F9D4} 
\cellcolor[HTML]{F6F9D4}                                            & \textbf{View 2} & 0.0976   & \textbf{0.0245}  & \textbf{0.0002}   \\ \cline{2-5} 
\rowcolor[HTML]{F6F9D4} 
\cellcolor[HTML]{F6F9D4}                                            & \textbf{View 3} & 0.086    & \textbf{0.0015}  & \textbf{0.0002}   \\ \cline{2-5} 
\rowcolor[HTML]{F6F9D4} 
\cellcolor[HTML]{F6F9D4}                                            & \textbf{View 4} & 0.0599   & \textbf{0.0115 } & \textbf{0.0211 }  \\ \cline{2-5} 
\rowcolor[HTML]{F6F9D4} 
\multirow{-5}{*}{\cellcolor[HTML]{F6F9D4}\textbf{MultiGraphGAN+BC}} & \textbf{View 5} & 0.0805   & 0.0549  & \textbf{0.0102 }  \\ \hline
\rowcolor[HTML]{DEDCE6} 
\cellcolor[HTML]{DEDCE6}                                            & \textbf{View 1} & 0.0679   & 0.1076  & \textbf{0.0059}   \\ \cline{2-5} 
\rowcolor[HTML]{DEDCE6} 
\cellcolor[HTML]{DEDCE6}                                            & \textbf{View 2} & \textbf{1.4}$\times10^{-7}$   & 0.1027  & 0.0723   \\ \cline{2-5} 
\rowcolor[HTML]{DEDCE6} 
\cellcolor[HTML]{DEDCE6}                                            & \textbf{View 3} & \textbf{0.0177}   & 0.0917  & \textbf{0.0264 }  \\ \cline{2-5} 
\rowcolor[HTML]{DEDCE6} 
\cellcolor[HTML]{DEDCE6}                                            & \textbf{View 4} & \textbf{8.5}$\times10^{-11}$  & \textbf{0.0054}  & 0.0609   \\ \cline{2-5} 
\rowcolor[HTML]{DEDCE6} 
\multirow{-5}{*}{\cellcolor[HTML]{DEDCE6}\textbf{MultiGraphGAN+EC}} & \textbf{View 5} & \textbf{0.0161}   & 0.0548  & 0.086    \\ \hline

\multicolumn{2}{|c|}{\cellcolor[HTML]{FFFFFF}}                                                  & \cellcolor[HTML]{FCE4D6}                                      & \cellcolor[HTML]{E2EFDA}                                      & \cellcolor[HTML]{F5D4FF}                                     \\

\multicolumn{2}{|c|}{\multirow{-2}{*}{\cellcolor[HTML]{FFFFFF}\textbf{View 4}}}                 & \multirow{-2}{*}{\cellcolor[HTML]{FCE4D6}\textbf{topoGAN+CC}} & \multirow{-2}{*}{\cellcolor[HTML]{E2EFDA}\textbf{topoGAN+BC}} & \multirow{-2}{*}{\cellcolor[HTML]{F5D4FF}\textbf{topoGAN+EC}} \\ \hline
\rowcolor[HTML]{FFF2CC} 
\cellcolor[HTML]{FFF2CC}                                                      & \textbf{View 1} & \textbf{0.0437}                                               & 0.0762                                                        & \textbf{0.0008}                                               \\ \cline{2-5} 
\rowcolor[HTML]{FFF2CC} 
\cellcolor[HTML]{FFF2CC}                                                      & \textbf{View 2} & 0.2806                                                        & 0.2209                                                        & 0.0847                                                        \\ \cline{2-5} 
\rowcolor[HTML]{FFF2CC} 
\cellcolor[HTML]{FFF2CC}                                                      & \textbf{View 3} & 0.1332                                                        & 0.1404                                                        & \textbf{8.6744}$\times10^{-7}$                                \\ \cline{2-5} 
\rowcolor[HTML]{FFF2CC} 
\cellcolor[HTML]{FFF2CC}                                                      & \textbf{View 4} & 0.1761                                                        & 0.0564                                                        & \textbf{0.0316}                                               \\ \cline{2-5} 
\rowcolor[HTML]{FFF2CC} 
\multirow{-5}{*}{\cellcolor[HTML]{FFF2CC}\textbf{Adapted MWGAN}}              & \textbf{View 5} & 0.166                                                         & \textbf{0.0047}                                               & 0.1258                                                        \\ \hline
\rowcolor[HTML]{DDEBF7} 
\cellcolor[HTML]{DDEBF7}                                                      & \textbf{View 1} & 0.1384                                                        & \textbf{1.2571}$\times10^{-16}$                               & 0.1709                                                        \\ \cline{2-5} 
\rowcolor[HTML]{DDEBF7} 
\cellcolor[HTML]{DDEBF7}                                                      & \textbf{View 2} & \textbf{0.0002}                                               & \textbf{3.1156}$\times10^{-11}$                               & 0.373                                                         \\ \cline{2-5} 
\rowcolor[HTML]{DDEBF7} 
\cellcolor[HTML]{DDEBF7}                                                      & \textbf{View 3} & \textbf{0.0462}                                               & \textbf{6.7905}$\times10^{-7}$                                & \textbf{0.0099}                                               \\ \cline{2-5} 
\rowcolor[HTML]{DDEBF7} 
\cellcolor[HTML]{DDEBF7}                                                      & \textbf{View 4} & \textbf{0.0269}                                               & \textbf{6.8363}$\times10^{-11}$                               & \textbf{4.0492}$\times10^{-37}$                               \\ \cline{2-5} 
\rowcolor[HTML]{DDEBF7} 
\multirow{-5}{*}{\cellcolor[HTML]{DDEBF7}\textbf{Adapted MWGAN (clustering)}} & \textbf{View 5} & 0.1468                                                        & \textbf{2.1769}$\times10^{-6}$                               & 0.2629                                                        \\ \hline

\rowcolor[HTML]{FFE5CA} 
\cellcolor[HTML]{FFE5CA}                                            & \textbf{View 1} & 0.0835  & \textbf{2.1}$\times10^{-14}$  & 0.1468   \\ \cline{2-5} 
\rowcolor[HTML]{FFE5CA} 
\cellcolor[HTML]{FFE5CA}                                            & \textbf{View 2} & 0.0793  & \textbf{0.0015}   & \textbf{0.0321}   \\ \cline{2-5} 
\rowcolor[HTML]{FFE5CA} 
\cellcolor[HTML]{FFE5CA}                                            & \textbf{View 3} & 0.1515  & \textbf{0.0023}   & \textbf{0.0044}   \\ \cline{2-5} 
\rowcolor[HTML]{FFE5CA} 
\cellcolor[HTML]{FFE5CA}                                            & \textbf{View 4} & 0.1276  & \textbf{0.0076}   & \textbf{0.0007}   \\ \cline{2-5} 
\rowcolor[HTML]{FFE5CA} 
\multirow{-5}{*}{\cellcolor[HTML]{FFE5CA}\textbf{MultiGraphGAN+CC}} & \textbf{View 5} & \textbf{0.0231}  & \textbf{3.1}$\times10^{-10}$  & \textbf{0.0276}   \\ \hline
\rowcolor[HTML]{DFCCE4} 
\cellcolor[HTML]{DFCCE4}                                            & \textbf{View 1} & \textbf{0.0122}  & \textbf{1.96}$\times10^{-6}$  & \textbf{0.0105}   \\ \cline{2-5} 
\rowcolor[HTML]{DFCCE4} 
\cellcolor[HTML]{DFCCE4}                                            & \textbf{View 2} & 0.1649  & 0.1468   & \textbf{0.0384}   \\ \cline{2-5} 
\rowcolor[HTML]{DFCCE4} 
\cellcolor[HTML]{DFCCE4}                                            & \textbf{View 3} & \textbf{0.0287}  & \textbf{3.97}$\times10^{-13}$  & \textbf{0.0211}   \\ \cline{2-5} 
\rowcolor[HTML]{DFCCE4} 
\cellcolor[HTML]{DFCCE4}                                            & \textbf{View 4} & \textbf{1.3}$\times10^{-5}$  & \textbf{0.0389 }  & 0.1389   \\ \cline{2-5} 
\rowcolor[HTML]{DFCCE4} 
\multirow{-5}{*}{\cellcolor[HTML]{DFCCE4}\textbf{MultiGraphGAN+BC}} & \textbf{View 5} & \textbf{0.0006}  & \textbf{1.3}$\times10^{-16}$  & \textbf{5.65}$\times10^{-8}$ \\ \hline
\rowcolor[HTML]{E0EFD4} 
\cellcolor[HTML]{E0EFD4}                                            & \textbf{View 1} & 0.1068  & \textbf{0.0314}   & 0.1585   \\ \cline{2-5} 
\rowcolor[HTML]{E0EFD4} 
\cellcolor[HTML]{E0EFD4}                                            & \textbf{View 2} & \textbf{0.0181 } & \textbf{0.027}    & 0.2697   \\ \cline{2-5} 
\rowcolor[HTML]{E0EFD4} 
\cellcolor[HTML]{E0EFD4}                                            & \textbf{View 3} & 0.0994  & \textbf{0.0048}   & \textbf{0.0004}   \\ \cline{2-5} 
\rowcolor[HTML]{E0EFD4} 
\cellcolor[HTML]{E0EFD4}                                            & \textbf{View 4} & \textbf{0.0001}  & \textbf{0.0464}   & 0.1533   \\ \cline{2-5} 
\rowcolor[HTML]{E0EFD4} 
\multirow{-5}{*}{\cellcolor[HTML]{E0EFD4}\textbf{MultiGraphGAN+EC}} & \textbf{View 5} & 0.1148  & \textbf{0.0444}   & 0.1643   \\ \hline

%\multicolumn{5}{p{550}}{Adapted MWGAN: the graph-based architecture of the method introduced in \citep{Cao:2019}. Adapted MWGAN (clustering): a variant of the adapted method \citep{Cao:2019} with a clustering step. topoGAN+CC, topoGAN+BC, topoGAN+EC: the proposed method that includes closeness, betweenness and eigenvector. MultiGraphGAN+CC, MultiGraphGAN+BC, MultiGraphGAN+EC: the state-of-the-art method \citep{bessadok:2020} that includes the reconstruction loss and the closeness, betweenness and eigenvector centralities.  View 3: mean sulcal depth (LH). View 4: maximum principal curvature (RH). We highlight in bold the $p-$value $\prec$ $0.05$ using two-tailed paired $t-$test between row-wise and column-wise methods.}\\
\end{tabular}
\begin{tablenotes}[para,flushleft]
      \footnotesize
      \item Adapted MWGAN: the graph-based architecture of the method introduced in \citep{Cao:2019}. Adapted MWGAN (clustering): a variant of the adapted method \citep{Cao:2019} with a clustering step. topoGAN+CC, topoGAN+BC, topoGAN+EC: the proposed method that includes closeness, betweenness and eigenvector. MultiGraphGAN+CC, MultiGraphGAN+BC, MultiGraphGAN+EC: the state-of-the-art method \citep{bessadok:2020} that includes the reconstruction loss and the closeness, betweenness and eigenvector centralities.  View 3: mean sulcal depth (LH). View 4: maximum principal curvature (RH). We highlight in bold the $p-$value $\prec$ $0.05$ using two-tailed paired $t-$test between row-wise and column-wise methods.
\end{tablenotes}
\end{threeparttable}
\label{p-value-2}
\end{table}

\begin{table}
\begin{threeparttable}
\centering
\vspace{-30pt}
\captionsetup{justification=centering}
\caption{P-value results using two-tailed paired t-test for predicting a target brain multigraph using six source views.}% }
\scriptsize
\begin{tabular}{|c|c|c|c|c|}
\hline
\multicolumn{2}{|c|}{\cellcolor[HTML]{FFFFFF}}                                                  & \cellcolor[HTML]{FCE4D6}                                      & \cellcolor[HTML]{E2EFDA}                                      & \cellcolor[HTML]{F5D4FF}                                      \\
\multicolumn{2}{|c|}{\multirow{-2}{*}{\cellcolor[HTML]{FFFFFF}\textbf{View 5}}}                 & \multirow{-2}{*}{\cellcolor[HTML]{FCE4D6}\textbf{topoGAN+CC}} & \multirow{-2}{*}{\cellcolor[HTML]{E2EFDA}\textbf{topoGAN+BC}} & \multirow{-2}{*}{\cellcolor[HTML]{F5D4FF}\textbf{topoGAN+EC}} \\ \hline
\rowcolor[HTML]{FFF2CC} 
\cellcolor[HTML]{FFF2CC}                                                      & \textbf{View 1} & \textbf{0.03}                                                 & \textbf{1.9211}$\times10^{-15}$                               & \textbf{0.0002}                                               \\ \cline{2-5} 
\rowcolor[HTML]{FFF2CC} 
\cellcolor[HTML]{FFF2CC}                                                      & \textbf{View 2} & 0.0967                                                        & \textbf{2.9925}$\times10^{-6}$                                & 0.2356                                                        \\ \cline{2-5} 
\rowcolor[HTML]{FFF2CC} 
\cellcolor[HTML]{FFF2CC}                                                      & \textbf{View 3} & \textbf{7.8964}$\times10^{-6}$                                & \textbf{2.9087}$\times10^{-11}$                               & \textbf{0.0006}                                               \\ \cline{2-5} 
\rowcolor[HTML]{FFF2CC} 
\cellcolor[HTML]{FFF2CC}                                                      & \textbf{View 4} & \textbf{1.3136}$\times10^{-7}$                                & \textbf{9.1108}$\times10^{-9}$                               & \textbf{4.8735}$\times10^{-10}$                               \\ \cline{2-5} 
\rowcolor[HTML]{FFF2CC} 
\multirow{-5}{*}{\cellcolor[HTML]{FFF2CC}\textbf{Adapted MWGAN}}              & \textbf{View 5} & \textbf{0.029}                                                & \textbf{0.0002}                                               & \textbf{0.0037}                                               \\ \hline
\rowcolor[HTML]{DDEBF7} 
\cellcolor[HTML]{DDEBF7}                                                      & \textbf{View 1} & 0.0997                                                        & \textbf{1.3734}$\times10^{-9}$                                & \textbf{0.0489}                                               \\ \cline{2-5} 
\rowcolor[HTML]{DDEBF7} 
\cellcolor[HTML]{DDEBF7}                                                      & \textbf{View 2} & \textbf{0.046}                                                & \textbf{0.0002}                                               & 0.2289                                                        \\ \cline{2-5} 
\rowcolor[HTML]{DDEBF7} 
\cellcolor[HTML]{DDEBF7}                                                      & \textbf{View 3} & 0.1573                                                        & \textbf{0.0319}                                               & 0.2678                                                        \\ \cline{2-5} 
\rowcolor[HTML]{DDEBF7} 
\cellcolor[HTML]{DDEBF7}                                                      & \textbf{View 4} & 0.1849                                                        & 0.2604                                                        & 0.108                                                         \\ \cline{2-5} 
\rowcolor[HTML]{DDEBF7} 
\multirow{-5}{*}{\cellcolor[HTML]{DDEBF7}\textbf{Adapted MWGAN (clustering)}} & \textbf{View 5} & 0.1197                                                        & \textbf{0.012}                                                & 0.0808                                                        \\ \hline

\rowcolor[HTML]{FFE5CA} 
\cellcolor[HTML]{FFE5CA}                                            & \textbf{View 1} & 0.2674 & \textbf{0.0074} & 0.1941 \\ \cline{2-5} 
\rowcolor[HTML]{FFE5CA} 
\cellcolor[HTML]{FFE5CA}                                            & \textbf{View 2} & 0.3214 & \textbf{0.0206} & 0.4486 \\ \cline{2-5} 
\rowcolor[HTML]{FFE5CA} 
\cellcolor[HTML]{FFE5CA}                                            & \textbf{View 3} & 0.39   & 0.1531 & 0.0641 \\ \cline{2-5} 
\rowcolor[HTML]{FFE5CA} 
\cellcolor[HTML]{FFE5CA}                                            & \textbf{View 4} & 0.1074 & 0.0933 & 0.0707 \\ \cline{2-5} 
\rowcolor[HTML]{FFE5CA} 
\multirow{-5}{*}{\cellcolor[HTML]{FFE5CA}\textbf{MultiGraphGAN+CC}} & \textbf{View 5} & 0.099  & 0.0514 & 0.095  \\ \hline
\rowcolor[HTML]{DFCCE4} 
\cellcolor[HTML]{DFCCE4}                                            & \textbf{View 1} & 0.3093 & \textbf{0.034}  & 0.1881 \\ \cline{2-5} 
\rowcolor[HTML]{DFCCE4} 
\cellcolor[HTML]{DFCCE4}                                            & \textbf{View 2} & 0.1106 & \textbf{0.013}  & 0.1803 \\ \cline{2-5} 
\rowcolor[HTML]{DFCCE4} 
\cellcolor[HTML]{DFCCE4}                                            & \textbf{View 3} & 0.2799 & 0.208  & 0.0787 \\ \cline{2-5} 
\rowcolor[HTML]{DFCCE4} 
\cellcolor[HTML]{DFCCE4}                                            & \textbf{View 4} & 0.1296 & 0.0765 & 0.1991 \\ \cline{2-5} 
\rowcolor[HTML]{DFCCE4} 
\multirow{-5}{*}{\cellcolor[HTML]{DFCCE4}\textbf{MultiGraphGAN+BC}} & \textbf{View 5} & 0.0954 & 0.3384 & 0.1493 \\ \hline
\rowcolor[HTML]{E0EFD4} 
\cellcolor[HTML]{E0EFD4}                                            & \textbf{View 1} & 0.3618 & \textbf{0.0036} & 0.1621 \\ \cline{2-5} 
\rowcolor[HTML]{E0EFD4} 
\cellcolor[HTML]{E0EFD4}                                            & \textbf{View 2} & 0.2133 & 0.27   & 0.1087 \\ \cline{2-5} 
\rowcolor[HTML]{E0EFD4} 
\cellcolor[HTML]{E0EFD4}                                            & \textbf{View 3} & \textbf{0.0444} & \textbf{0.0326} & 0.1248 \\ \cline{2-5} 
\rowcolor[HTML]{E0EFD4} 
\cellcolor[HTML]{E0EFD4}                                            & \textbf{View 4} & 0.0509 & \textbf{0.0295} & \textbf{0.0323} \\ \cline{2-5} 
\rowcolor[HTML]{E0EFD4} 
\multirow{-5}{*}{\cellcolor[HTML]{E0EFD4}\textbf{MultiGraphGAN+EC}} & \textbf{View 5} & 0.2233 & 0.0551 & 0.1527 \\ \hline
\multicolumn{2}{|c|}{\cellcolor[HTML]{FFFFFF}}                                                  & \cellcolor[HTML]{FCE4D6}                                      & \cellcolor[HTML]{E2EFDA}                                      & \cellcolor[HTML]{F5D4FF}                                      \\

\multicolumn{2}{|c|}{\multirow{-2}{*}{\cellcolor[HTML]{FFFFFF}\textbf{View 6}}}                 & \multirow{-2}{*}{\cellcolor[HTML]{FCE4D6}\textbf{topoGAN+CC}} & \multirow{-2}{*}{\cellcolor[HTML]{E2EFDA}\textbf{topoGAN+BC}} & \multirow{-2}{*}{\cellcolor[HTML]{F5D4FF}\textbf{topoGAN+EC}} \\ \hline
\rowcolor[HTML]{FFF2CC} 
\cellcolor[HTML]{FFF2CC}                                                      & \textbf{View 1} & \textbf{0.047}                                                & \textbf{0.0035}                                               & \textbf{0.0396}                                               \\ \cline{2-5} 
\rowcolor[HTML]{FFF2CC} 
\cellcolor[HTML]{FFF2CC}                                                      & \textbf{View 2} & 0.1528                                                        & 0.0824                                                        & 0.0838                                                        \\ \cline{2-5} 
\rowcolor[HTML]{FFF2CC} 
\cellcolor[HTML]{FFF2CC}                                                      & \textbf{View 3} & \textbf{0.0129}                                               & \textbf{0.0019}                                               & \textbf{0.0044}                                               \\ \cline{2-5} 
\rowcolor[HTML]{FFF2CC} 
\cellcolor[HTML]{FFF2CC}                                                      & \textbf{View 4} & 0.1271                                                        & 0.0515                                                        & 0.0626                                                        \\ \cline{2-5} 
\rowcolor[HTML]{FFF2CC} 
\multirow{-5}{*}{\cellcolor[HTML]{FFF2CC}\textbf{Adapted MWGAN}}              & \textbf{View 5} & 0.0904                                                        & 0.0647                                                        & 0.1407                                                        \\ \hline
\rowcolor[HTML]{DDEBF7} 
\cellcolor[HTML]{DDEBF7}                                                      & \textbf{View 1} & \textbf{0.034}                                                & \textbf{0.0399}                                               & 0.0522                                                        \\ \cline{2-5} 
\rowcolor[HTML]{DDEBF7} 
\cellcolor[HTML]{DDEBF7}                                                      & \textbf{View 2} & \textbf{0.034}                                                & \textbf{0.0335}                                               & \textbf{0.0019}                                               \\ \cline{2-5} 
\rowcolor[HTML]{DDEBF7} 
\cellcolor[HTML]{DDEBF7}                                                      & \textbf{View 3} & \textbf{0.0152}                                               & \textbf{0.0295}                                               & 0.0916                                                        \\ \cline{2-5} 
\rowcolor[HTML]{DDEBF7} 
\cellcolor[HTML]{DDEBF7}                                                      & \textbf{View 4} & \textbf{0.0134}                                               & 0.0592                                                        & 0.075                                                         \\ \cline{2-5} 
\rowcolor[HTML]{DDEBF7} 
\multirow{-5}{*}{\cellcolor[HTML]{DDEBF7}\textbf{Adapted MWGAN (clustering)}} & \textbf{View 5} & 0.0908                                                        & 0.0732                                                        & \textbf{0.0014}                                               \\ \hline 

\rowcolor[HTML]{FFE5CA} 
\cellcolor[HTML]{FFE5CA}                                            & \textbf{View 1} & 0.0631 & \textbf{0.0466} & 0.0562   \\ \cline{2-5} 
\rowcolor[HTML]{FFE5CA} 
\cellcolor[HTML]{FFE5CA}                                            & \textbf{View 2} & 0.0805 & \textbf{0.0273} & \textbf{0.015}    \\ \cline{2-5} 
\rowcolor[HTML]{FFE5CA} 
\cellcolor[HTML]{FFE5CA}                                            & \textbf{View 3} & 0.0866 & \textbf{0.0304} & 0.0751   \\ \cline{2-5} 
\rowcolor[HTML]{FFE5CA} 
\cellcolor[HTML]{FFE5CA}                                            & \textbf{View 4} & 0.0584 & 0.0619 & 0.0694   \\ \cline{2-5} 
\rowcolor[HTML]{FFE5CA} 
\multirow{-5}{*}{\cellcolor[HTML]{FFE5CA}\textbf{MultiGraphGAN+CC}} & \textbf{View 5} & \textbf{0.028}  & \textbf{0.0217} & 0.0526   \\ \hline
\rowcolor[HTML]{DFCCE4} 
\cellcolor[HTML]{DFCCE4}                                            & \textbf{View 1} & \textbf{0.0113} & 0.0601 & 8.95E-06 \\ \cline{2-5} 
\rowcolor[HTML]{DFCCE4} 
\cellcolor[HTML]{DFCCE4}                                            & \textbf{View 2} & \textbf{0.0006} & 0.1182 & \textbf{0.0014 }  \\ \cline{2-5} 
\rowcolor[HTML]{DFCCE4} 
\cellcolor[HTML]{DFCCE4}                                            & \textbf{View 3} & 0.1567 & 0.0928 & 0.0771   \\ \cline{2-5} 
\rowcolor[HTML]{DFCCE4} 
\cellcolor[HTML]{DFCCE4}                                            & \textbf{View 4} & \textbf{0.0308} & 0.0644 & \textbf{0.0073}   \\ \cline{2-5} 
\rowcolor[HTML]{DFCCE4} 
\multirow{-5}{*}{\cellcolor[HTML]{DFCCE4}\textbf{MultiGraphGAN+BC}} & \textbf{View 5} & \textbf{0.0011} & \textbf{0.0286} & \textbf{0.0423 }  \\ \hline
\rowcolor[HTML]{E0EFD4} 
\cellcolor[HTML]{E0EFD4}                                            & \textbf{View 1} & \textbf{0.0005} & 0.0063 & \textbf{0.0325}   \\ \cline{2-5} 
\rowcolor[HTML]{E0EFD4} 
\cellcolor[HTML]{E0EFD4}                                            & \textbf{View 2} & \textbf{0.0009} & 0.1067 & \textbf{0.0039}   \\ \cline{2-5} 
\rowcolor[HTML]{E0EFD4} 
\cellcolor[HTML]{E0EFD4}                                            & \textbf{View 3} & \textbf{0.0109 }& \textbf{0.0402 }& \textbf{0.0224}   \\ \cline{2-5} 
\rowcolor[HTML]{E0EFD4} 
\cellcolor[HTML]{E0EFD4}                                            & \textbf{View 4} & \textbf{0.0225} & \textbf{0.043}  & \textbf{0.0247 }  \\ \cline{2-5} 
\rowcolor[HTML]{E0EFD4} 
\multirow{-5}{*}{\cellcolor[HTML]{E0EFD4}\textbf{MultiGraphGAN+EC}} & \textbf{View 5} & \textbf{0.0072} & 0.0057 & \textbf{0.0174}   \\ \hline

%\multicolumn{5}{p{550}}{Adapted MWGAN: the graph-based architecture of the method introduced in \citep{Cao:2019}. Adapted MWGAN (clustering): a variant of the adapted method \citep{Cao:2019} with a clustering step. topoGAN+CC, topoGAN+BC, topoGAN+EC: the proposed method that includes closeness, betweenness and eigenvector.  MultiGraphGAN+CC, MultiGraphGAN+BC, MultiGraphGAN+EC: the state-of-the-art method \citep{bessadok:2020} that includes the reconstruction loss and the closeness, betweenness and eigenvector centralities. View 5: average curvature (RH). View 6: mean sulcal depth (RH). We highlight in bold the $p$-value $\prec$ $0.05$ using two-tailed paired $t$-test between row-wise and column-wise methods.}\\
\end{tabular}
\begin{tablenotes}[para,flushleft]
      \footnotesize
      \item Adapted MWGAN: the graph-based architecture of the method introduced in \citep{Cao:2019}. Adapted MWGAN (clustering): a variant of the adapted method \citep{Cao:2019} with a clustering step. topoGAN+CC, topoGAN+BC, topoGAN+EC: the proposed method that includes closeness, betweenness and eigenvector.  MultiGraphGAN+CC, MultiGraphGAN+BC, MultiGraphGAN+EC: the state-of-the-art method \citep{bessadok:2020} that includes the reconstruction loss and the closeness, betweenness and eigenvector centralities. View 5: average curvature (RH). View 6: mean sulcal depth (RH). We highlight in bold the $p$-value $\prec$ $0.05$ using two-tailed paired $t$-test between row-wise and column-wise methods.
\end{tablenotes}
\end{threeparttable}
\label{p-value-3}
\end{table}

\begin{landscape}
\begin{table}
\begin{threeparttable}
\vspace{-50pt}
\centering
\captionsetup{justification=centering}
\caption{Prediction results for six source views derived from the left and right hemispheres using 3-fold cross-validation strategy.} %}
\footnotesize
\begin{tabular}{|c|c|c|c|c|c|c|c|}
\hline
\rowcolor[HTML]{F2F2F2} 
\cellcolor[HTML]{FFFFFF}\textbf{View 1} & \textbf{MAE}                            & \textbf{MAE(CC)}                        & \textbf{MAE(BC)}                        & \textbf{MAE(EC)}                        & \textbf{MAE(PC)}                        & \textbf{MAE(EFF)}                       & \textbf{MAE(Clst)}                      \\ \hline
\rowcolor[HTML]{FFF5CE} 
\textbf{Adapted MWGAN}                  & {\color[HTML]{0600FF} \textbf{0.19319}} & 0.27894                                 & 0.01212                                 & 0.01743                                 & {\color[HTML]{0600FF} \textbf{0.00375}} & 7.55474                                 & 0.40396                                 \\ \hline
\rowcolor[HTML]{DEE6EF} 
\textbf{Adapted MWGAN (clustering)}     & 0.20226                                 & 0.12416                                 & 0.00449                                 & 0.00936                                 & 0.00517                                 & 4.00669                                 & 0.14722                                 \\ \hline
\rowcolor[HTML]{DEDCE6} 
\textbf{MultiGraphGAN}                  & 0.20063                                 & 0.13705                                 & 0.00502                                 & 0.00973                                 & 0.00523                                 & 4.39254                                 & 0.16524                                 \\ \hline
\rowcolor[HTML]{FFD7D7} 
\textbf{topoGAN}                        & 0.19342                                 & {\color[HTML]{0600FF} \textbf{0.09916}} & {\color[HTML]{0600FF} \textbf{0.00348}} & {\color[HTML]{0600FF} \textbf{0.00783}} & 0.0046                                  & {\color[HTML]{0600FF} \textbf{3.26974}} & {\color[HTML]{0600FF} \textbf{0.11495}} \\ \hline
\rowcolor[HTML]{F2F2F2} 
\cellcolor[HTML]{FFFFFF}\textbf{View 2} & \textbf{MAE}                            & \textbf{MAE(CC)}                        & \textbf{MAE(BC)}                        & \textbf{MAE(EC)}                        & \textbf{MAE(PC)}                        & \textbf{MAE(EFF)}                       & \textbf{MAE(Clst)}                      \\ \hline
\rowcolor[HTML]{FFF5CE} 
\textbf{Adapted MWGAN}                  & {\color[HTML]{0600FF} \textbf{0.20901}} & 0.3032                                  & 0.01362                                 & 0.01755                                 & 0.00552                                 & 7.63256                                 & 0.44805                                 \\ \hline
\rowcolor[HTML]{DEE6EF} 
\textbf{Adapted MWGAN (clustering)}     & 0.21127                                 & {\color[HTML]{0600FF} \textbf{0.16951}} & {\color[HTML]{0600FF} \textbf{0.00648}} & 0.01323                                 & {\color[HTML]{0600FF} \textbf{0.00462}} & {\color[HTML]{0600FF} \textbf{5.1742}}  & {\color[HTML]{0600FF} \textbf{0.20651}} \\ \hline
\rowcolor[HTML]{DEDCE6} 
\textbf{MultiGraphGAN}                  & 0.21245                                 & 0.18279                                 & 0.00708                                 & 0.01242                                 & 0.0062                                  & 5.58634                                 & 0.2296                                  \\ \hline
\rowcolor[HTML]{FFD7D7} 
\textbf{topoGAN}                        & 0.21138                                 & 0.18863                                 & 0.00733                                 & {\color[HTML]{0600FF} \textbf{0.01236}} & 0.00545                                 & 5.78864                                 & 0.24022                                 \\ \hline
\rowcolor[HTML]{F2F2F2} 
\cellcolor[HTML]{FFFFFF}\textbf{View 3} & \textbf{MAE}                            & \textbf{MAE(CC)}                        & \textbf{MAE(BC)}                        & \textbf{MAE(EC)}                        & \textbf{MAE(PC)}                        & \textbf{MAE(EFF)}                       & \textbf{MAE(Clst)}                      \\ \hline
\rowcolor[HTML]{FFF5CE} 
\textbf{Adapted MWGAN}                  & {\color[HTML]{0600FF} \textbf{0.12093}} & 0.21652                                 & 0.00871                                 & 0.01469                                 & 0.0037                                  & 6.41082                                 & 0.28415                                 \\ \hline
\rowcolor[HTML]{DEE6EF} 
\textbf{Adapted MWGAN (clustering)}     & 0.15036                                 & 0.14922                                 & 0.00556                                 & 0.01116                                 & 0.00385                                 & 4.72162                                 & 0.18153                                 \\ \hline
\rowcolor[HTML]{DEDCE6} 
\textbf{MultiGraphGAN}                  & 0.14884                                 & {\color[HTML]{0600FF} \textbf{0.12242}} & {\color[HTML]{0600FF} \textbf{0.00442}} & {\color[HTML]{0600FF} \textbf{0.00936}} & 0.00405                                 & {\color[HTML]{0600FF} \textbf{3.96439}} & {\color[HTML]{0600FF} \textbf{0.14465}} \\ \hline
\rowcolor[HTML]{FFD7D7} 
\textbf{topoGAN}                        & 0.13851                                 & 0.13969                                 & 0.00515                                 & 0.01044                                 & {\color[HTML]{0600FF} \textbf{0.00358}} & 4.42449                                 & 0.16662                                 \\ \hline
\rowcolor[HTML]{F2F2F2} 
\cellcolor[HTML]{FFFFFF}\textbf{View 4} & \textbf{MAE}                            & \textbf{MAE(CC)}                        & \textbf{MAE(BC)}                        & \textbf{MAE(EC)}                        & \textbf{MAE(PC)}                        & \textbf{MAE(EFF)}                       & \textbf{MAE(Clst)}                      \\ \hline
\rowcolor[HTML]{FFF5CE} 
\textbf{Adapted MWGAN}                  & 0.18549                                 & 0.27491                                 & 0.01193                                 & 0.01761                                 & {\color[HTML]{0600FF} \textbf{0.00435}} & 7.32622                                 & 0.3872                                  \\ \hline
\rowcolor[HTML]{DEE6EF} 
\textbf{Adapted MWGAN (clustering)}     & 0.19355                                 & 0.10838                                 & 0.00385                                 & {\color[HTML]{0600FF} \textbf{0.00853}} & 0.00471                                 & 3.52877                                 & 0.12668                                 \\ \hline
\rowcolor[HTML]{DEDCE6} 
\textbf{MultiGraphGAN}                  & 0.19864                                 & 0.12713                                 & 0.0046                                  & 0.0092                                  & 0.00474                                 & 4.11026                                 & 0.15115                                 \\ \hline
\rowcolor[HTML]{FFD7D7} 
\textbf{topoGAN}                        & {\color[HTML]{0600FF} \textbf{0.19029}} & {\color[HTML]{0600FF} \textbf{0.09762}} & {\color[HTML]{0600FF} \textbf{0.00342}} & 0.00891                                 & 0.00478                                 & {\color[HTML]{0600FF} \textbf{3.2047}}  & {\color[HTML]{0600FF} \textbf{0.1125}}  \\ \hline
\rowcolor[HTML]{F2F2F2} 
\cellcolor[HTML]{FFFFFF}\textbf{View 5} & \textbf{MAE}                            & \textbf{MAE(CC)}                        & \textbf{MAE(BC)}                        & \textbf{MAE(EC)}                        & \textbf{MAE(PC)}                        & \textbf{MAE(EFF)}                       & \textbf{MAE(Clst)}                      \\ \hline
\rowcolor[HTML]{FFF5CE} 
\textbf{Adapted MWGAN}                  & {\color[HTML]{0600FF} \textbf{0.20877}} & 0.29294                                 & 0.01303                                 & 0.02137                                 & {\color[HTML]{0600FF} \textbf{0.00506}} & 7.52582                                 & 0.42447                                 \\ \hline
\rowcolor[HTML]{DEE6EF} 
\textbf{Adapted MWGAN (clustering)}     & 0.21394                                 & 0.18547                                 & 0.00721                                 & 0.0131                                  & 0.00565                                 & 5.66924                                 & 0.23389                                 \\ \hline
\rowcolor[HTML]{DEDCE6} 
\textbf{MultiGraphGAN}                  & 0.21372                                 & 0.22346                                 & 0.00916                                 & 0.01538                                 & 0.00536                                 & 6.45255                                 & 0.30052                                 \\ \hline
\rowcolor[HTML]{FFD7D7} 
\textbf{topoGAN}                        & 0.21143                                 & {\color[HTML]{0600FF} \textbf{0.1744}}  & {\color[HTML]{0600FF} \textbf{0.00669}} & {\color[HTML]{0600FF} \textbf{0.01261}} & 0.00521                                 & {\color[HTML]{0600FF} \textbf{5.4011}}  & {\color[HTML]{0600FF} \textbf{0.21785}} \\ \hline
\rowcolor[HTML]{F2F2F2} 
\cellcolor[HTML]{FFFFFF}\textbf{View 6} & \textbf{MAE}                            & \textbf{MAE(CC)}                        & \textbf{MAE(BC)}                        & \textbf{MAE(EC)}                        & \textbf{MAE(PC)}                        & \textbf{MAE(EFF)}                       & \textbf{MAE(Clst)}                      \\ \hline
\rowcolor[HTML]{FFF5CE} 
\textbf{Adapted MWGAN}                  & 0.09501                                 & 0.32472                                 & 0.01514                                 & 0.02613                                 & 0.00543                                 & 7.4837                                  & 0.48978                                 \\ \hline
\rowcolor[HTML]{DEE6EF} 
\textbf{Adapted MWGAN (clustering)}     & {\color[HTML]{0600FF} \textbf{0.05621}} & {\color[HTML]{0600FF} \textbf{0.16894}} & {\color[HTML]{0600FF} \textbf{0.00643}} & 0.01149                                 & 0.00436                                 & {\color[HTML]{0600FF} \textbf{5.30278}} & {\color[HTML]{0600FF} \textbf{0.21254}} \\ \hline
\rowcolor[HTML]{DEDCE6} 
\textbf{MultiGraphGAN}                  & 0.06627                                 & 0.18414                                 & 0.0071                                  & {\color[HTML]{0600FF} \textbf{0.01143}} & {\color[HTML]{0600FF} \textbf{0.00431}} & 5.71455                                 & 0.2348                                  \\ \hline
\rowcolor[HTML]{FFD7D7} 
\textbf{topoGAN}                        & 0.10415                                 & 0.17302                                 & 0.00664                                 & 0.01254                                 & 0.00456                                 & 5.32186                                 & 0.2146                                  \\ \cline{1-8}
%\multicolumn{8}{p{680}}{View 1: maximum principal curvature (LH). View 2: average curvature (LH). View 3: mean sulcal depth (LH). View 4: maximum principal curvature (RH). View 5: average curvature (RH). View 6: mean sulcal depth (RH). MAE: mean absolute error. CC: closeness centrality. BC: betweenness centrality. EC: eigenvector centrality. PC: PageRank centrality. EFF: effective size. Clst: clustering coefficient. We highlight in blue the lowest MAE resulting from a particular evaluation metric for the GCN-based comparaison methods and our topoGAN method learned using EC in our loss function.}
\end{tabular}
\begin{tablenotes}[para,flushleft]
      \footnotesize
      \item View 1: maximum principal curvature (LH). View 2: average curvature (LH). View 3: mean sulcal depth (LH). View 4: maximum principal curvature (RH). View 5: average curvature (RH). View 6: mean sulcal depth (RH). MAE: mean absolute error. CC: closeness centrality. BC: betweenness centrality. EC: eigenvector centrality. PC: PageRank centrality. EFF: effective size. Clst: clustering coefficient. We highlight in blue the lowest MAE resulting from a particular evaluation metric for the GCN-based comparaison methods and our topoGAN method learned using EC in our loss function.
\end{tablenotes}
\end{threeparttable}
\label{CV}
\end{table}
\end{landscape}

\newpage
\bibliography{Biblio3}

\begin{thebibliography}{81}
\expandafter\ifx\csname natexlab\endcsname\relax\def\natexlab#1{#1}\fi
\expandafter\ifx\csname url\endcsname\relax
  \def\url#1{\texttt{#1}}\fi
\expandafter\ifx\csname urlprefix\endcsname\relax\def\urlprefix{URL }\fi
\providecommand{\eprint}[2][]{\url{#2}}
\providecommand{\bibinfo}[2]{#2}
\ifx\xfnm\relax \def\xfnm[#1]{\unskip,\space#1}\fi
%Type = Article
\bibitem[{Albuquerque et~al.(2019)Albuquerque, Monteiro, Doan, Considine, Falk
  and Mitliagkas}]{Albuquerque:2019}
\bibinfo{author}{Albuquerque, I.}, \bibinfo{author}{Monteiro, J.},
  \bibinfo{author}{Doan, T.}, \bibinfo{author}{Considine, B.},
  \bibinfo{author}{Falk, T.}, \bibinfo{author}{Mitliagkas, I.},
  \bibinfo{year}{2019}.
\newblock \bibinfo{title}{Multi-objective training of generative adversarial
  networks with multiple discriminators}.
\newblock \bibinfo{journal}{arXiv preprint arXiv:1901.08680} .
%Type = Article
\bibitem[{Banka and Rekik(2019)}]{Banka:2019}
\bibinfo{author}{Banka, A.}, \bibinfo{author}{Rekik, I.}, \bibinfo{year}{2019}.
\newblock \bibinfo{title}{Adversarial connectome embedding for mild cognitive
  impairment identification using cortical morphological networks}.
\newblock \bibinfo{journal}{International Workshop on Connectomics in
  Neuroimaging} , \bibinfo{pages}{74--82}.
%Type = Article
\bibitem[{Bassett and Sporns(2017)}]{Bassett:2017}
\bibinfo{author}{Bassett, D.S.}, \bibinfo{author}{Sporns, O.},
  \bibinfo{year}{2017}.
\newblock \bibinfo{title}{Network neuroscience}.
\newblock \bibinfo{journal}{Nature neuroscience} \bibinfo{volume}{20},
  \bibinfo{pages}{353--364}.
%Type = Article
\bibitem[{Beauchamp(1965)}]{Beauchamp:1965}
\bibinfo{author}{Beauchamp, M.A.}, \bibinfo{year}{1965}.
\newblock \bibinfo{title}{An improved index of centrality}.
\newblock \bibinfo{journal}{Behavioral science} \bibinfo{volume}{10},
  \bibinfo{pages}{161--163}.
%Type = Article
\bibitem[{Belli and Kipf(2019)}]{belli:2019}
\bibinfo{author}{Belli, D.}, \bibinfo{author}{Kipf, T.}, \bibinfo{year}{2019}.
\newblock \bibinfo{title}{Image-conditioned graph generation for road network
  extraction}.
\newblock \bibinfo{journal}{arXiv preprint arXiv:1910.14388} .
%Type = Article
\bibitem[{Bessadok et~al.(2019a)Bessadok, Mahjoub and Rekik}]{Bessadok:2019b}
\bibinfo{author}{Bessadok, A.}, \bibinfo{author}{Mahjoub, M.A.},
  \bibinfo{author}{Rekik, I.}, \bibinfo{year}{2019}a.
\newblock \bibinfo{title}{Hierarchical adversarial connectomic domain alignment
  for target brain graph prediction and classification from a source graph}.
\newblock \bibinfo{journal}{International Workshop on PRedictive Intelligence
  In MEdicine} , \bibinfo{pages}{105--114}.
%Type = Article
\bibitem[{Bessadok et~al.(2019b)Bessadok, Mahjoub and Rekik}]{Bessadok:2019a}
\bibinfo{author}{Bessadok, A.}, \bibinfo{author}{Mahjoub, M.A.},
  \bibinfo{author}{Rekik, I.}, \bibinfo{year}{2019}b.
\newblock \bibinfo{title}{Symmetric dual adversarial connectomic domain
  alignment for predicting isomorphic brain graph from a baseline graph}.
\newblock \bibinfo{journal}{International Conference on Medical Image Computing
  and Computer-Assisted Intervention} , \bibinfo{pages}{465--474}.
%Type = Article
\bibitem[{Bessadok et~al.(2020a)Bessadok, Mahjoub and Rekik}]{lgdada}
\bibinfo{author}{Bessadok, A.}, \bibinfo{author}{Mahjoub, M.A.},
  \bibinfo{author}{Rekik, I.}, \bibinfo{year}{2020}a.
\newblock \bibinfo{title}{Brain graph synthesis by dual adversarial domain
  alignment and target graph prediction from a source graph}.
\newblock \bibinfo{journal}{Medical Image Analysis} , \bibinfo{pages}{101902}.
%Type = Inproceedings
\bibitem[{Bessadok et~al.(2020b)Bessadok, Mahjoub and Rekik}]{bessadok:2020}
\bibinfo{author}{Bessadok, A.}, \bibinfo{author}{Mahjoub, M.A.},
  \bibinfo{author}{Rekik, I.}, \bibinfo{year}{2020}b.
\newblock \bibinfo{title}{Topology-aware generative adversarial network for
  joint prediction of multiple brain graphs from a single brain graph}, in:
  \bibinfo{booktitle}{International Conference on Medical Image Computing and
  Computer-Assisted Intervention}, \bibinfo{organization}{Springer}. pp.
  \bibinfo{pages}{551--561}.
%Type = Inproceedings
\bibitem[{Bessadok and Rekik(2018)}]{bessadok2018intact}
\bibinfo{author}{Bessadok, A.}, \bibinfo{author}{Rekik, I.},
  \bibinfo{year}{2018}.
\newblock \bibinfo{title}{Intact connectional morphometricity learning using
  multi-view morphological brain networks with application to autism spectrum
  disorder}, in: \bibinfo{booktitle}{International Workshop on Connectomics in
  Neuroimaging}, \bibinfo{organization}{Springer}. pp. \bibinfo{pages}{38--46}.
%Type = Article
\bibitem[{Bonacich(2007)}]{Bonacich:2007}
\bibinfo{author}{Bonacich, P.}, \bibinfo{year}{2007}.
\newblock \bibinfo{title}{Some unique properties of eigenvector centrality}.
\newblock \bibinfo{journal}{Social networks} \bibinfo{volume}{29},
  \bibinfo{pages}{555--564}.
%Type = Article
\bibitem[{Borgatti and Everett(2006)}]{Borgatti:2006}
\bibinfo{author}{Borgatti, S.P.}, \bibinfo{author}{Everett, M.G.},
  \bibinfo{year}{2006}.
\newblock \bibinfo{title}{A graph-theoretic perspective on centrality}.
\newblock \bibinfo{journal}{Social networks} \bibinfo{volume}{28},
  \bibinfo{pages}{466--484}.
%Type = Article
\bibitem[{Bresson and Laurent(2019)}]{Bresson:2019}
\bibinfo{author}{Bresson, X.}, \bibinfo{author}{Laurent, T.},
  \bibinfo{year}{2019}.
\newblock \bibinfo{title}{A two-step graph convolutional decoder for molecule
  generation}.
\newblock \bibinfo{journal}{arXiv preprint arXiv:1906.03412} .
%Type = Article
\bibitem[{Brin(1998)}]{Brin:1998}
\bibinfo{author}{Brin, S.}, \bibinfo{year}{1998}.
\newblock \bibinfo{title}{The {PageRank} citation ranking: bringing order to
  the web}.
\newblock \bibinfo{journal}{Proceedings of ASIS, 1998} \bibinfo{volume}{98},
  \bibinfo{pages}{161--172}.
%Type = Article
\bibitem[{Bronstein et~al.(2017)Bronstein, Bruna, LeCun, Szlam and
  Vandergheynst}]{Bronstein:2017}
\bibinfo{author}{Bronstein, M.M.}, \bibinfo{author}{Bruna, J.},
  \bibinfo{author}{LeCun, Y.}, \bibinfo{author}{Szlam, A.},
  \bibinfo{author}{Vandergheynst, P.}, \bibinfo{year}{2017}.
\newblock \bibinfo{title}{Geometric deep learning: going beyond euclidean
  data}.
\newblock \bibinfo{journal}{IEEE Signal Processing Magazine}
  \bibinfo{volume}{34}, \bibinfo{pages}{18--42}.
%Type = Book
\bibitem[{Burt(2009)}]{Burt:2009}
\bibinfo{author}{Burt, R.S.}, \bibinfo{year}{2009}.
\newblock \bibinfo{title}{Structural holes: The social structure of
  competition}.
\newblock \bibinfo{publisher}{Harvard university press}.
%Type = Article
\bibitem[{Cao et~al.(2019)Cao, Mo, Zhang, Jia, Shen and Tan}]{Cao:2019}
\bibinfo{author}{Cao, J.}, \bibinfo{author}{Mo, L.}, \bibinfo{author}{Zhang,
  Y.}, \bibinfo{author}{Jia, K.}, \bibinfo{author}{Shen, C.},
  \bibinfo{author}{Tan, M.}, \bibinfo{year}{2019}.
\newblock \bibinfo{title}{Multi-marginal {Wasserstein GAN}}.
\newblock \bibinfo{journal}{Advances in Neural Information Processing Systems}
  , \bibinfo{pages}{1774--1784}.
%Type = Article
\bibitem[{Cengiz and Rekik(2019)}]{Cengiz:2019}
\bibinfo{author}{Cengiz, K.}, \bibinfo{author}{Rekik, I.},
  \bibinfo{year}{2019}.
\newblock \bibinfo{title}{Predicting high-resolution brain networks using
  hierarchically embedded and aligned multi-resolution neighborhoods}.
\newblock \bibinfo{journal}{International Workshop on PRedictive Intelligence
  In MEdicine} , \bibinfo{pages}{115--124}.
%Type = Article
\bibitem[{Cid et~al.(2019)Cid, Jimenez-del Toro, Poletti and
  M{\"u}ller}]{Cid:2019}
\bibinfo{author}{Cid, Y.D.}, \bibinfo{author}{Jimenez-del Toro, O.},
  \bibinfo{author}{Poletti, P.A.}, \bibinfo{author}{M{\"u}ller, H.},
  \bibinfo{year}{2019}.
\newblock \bibinfo{title}{A graph model of the lungs with morphology-based
  structure for tuberculosis type classification}.
\newblock \bibinfo{journal}{International Conference on Information Processing
  in Medical Imaging} , \bibinfo{pages}{372--383}.
%Type = Article
\bibitem[{De~Cao and Kipf(2018)}]{de:2018}
\bibinfo{author}{De~Cao, N.}, \bibinfo{author}{Kipf, T.}, \bibinfo{year}{2018}.
\newblock \bibinfo{title}{Molgan: An implicit generative model for small
  molecular graphs}.
\newblock \bibinfo{journal}{arXiv preprint arXiv:1805.11973} .
%Type = Article
\bibitem[{Dhifallah et~al.(2020)Dhifallah, Rekik and
  Initiative}]{Dhifallah:2020}
\bibinfo{author}{Dhifallah, S.}, \bibinfo{author}{Rekik, I.},
  \bibinfo{author}{Initiative, A.D.N.}, \bibinfo{year}{2020}.
\newblock \bibinfo{title}{Estimation of connectional brain templates using
  selective multi-view network normalization}.
\newblock \bibinfo{journal}{Medical Image Analysis} \bibinfo{volume}{59},
  \bibinfo{pages}{101567}.
%Type = Article
\bibitem[{Durugkar et~al.(2016)Durugkar, Gemp and Mahadevan}]{Durugkar:2016}
\bibinfo{author}{Durugkar, I.}, \bibinfo{author}{Gemp, I.},
  \bibinfo{author}{Mahadevan, S.}, \bibinfo{year}{2016}.
\newblock \bibinfo{title}{Generative multi-adversarial networks}.
\newblock \bibinfo{journal}{arXiv preprint arXiv:1611.01673} .
%Type = Article
\bibitem[{Fischl(2012)}]{Fischl:2012}
\bibinfo{author}{Fischl, B.}, \bibinfo{year}{2012}.
\newblock \bibinfo{title}{Freesurfer}.
\newblock \bibinfo{journal}{Neuroimage} \bibinfo{volume}{62},
  \bibinfo{pages}{774--781}.
%Type = Article
\bibitem[{Flam-Shepherd et~al.(2020)Flam-Shepherd, Wu and
  Aspuru-Guzik}]{Flam:2020}
\bibinfo{author}{Flam-Shepherd, D.}, \bibinfo{author}{Wu, T.},
  \bibinfo{author}{Aspuru-Guzik, A.}, \bibinfo{year}{2020}.
\newblock \bibinfo{title}{Graph deconvolutional generation}.
\newblock \bibinfo{journal}{arXiv preprint arXiv:2002.07087} .
%Type = Article
\bibitem[{Fornito et~al.(2015)Fornito, Zalesky and Breakspear}]{Fornito:2015}
\bibinfo{author}{Fornito, A.}, \bibinfo{author}{Zalesky, A.},
  \bibinfo{author}{Breakspear, M.}, \bibinfo{year}{2015}.
\newblock \bibinfo{title}{The connectomics of brain disorders}.
\newblock \bibinfo{journal}{Nature Reviews Neuroscience} \bibinfo{volume}{16},
  \bibinfo{pages}{159--172}.
%Type = Article
\bibitem[{Freeman(1977)}]{Freeman:1977}
\bibinfo{author}{Freeman, L.C.}, \bibinfo{year}{1977}.
\newblock \bibinfo{title}{A set of measures of centrality based on
  betweenness}.
\newblock \bibinfo{journal}{Sociometry} , \bibinfo{pages}{35--41}.
%Type = Article
\bibitem[{Ghribi et~al.(2019)Ghribi, Li, Lin, Shen and Rekik}]{Ghribi:2019}
\bibinfo{author}{Ghribi, O.}, \bibinfo{author}{Li, G.}, \bibinfo{author}{Lin,
  W.}, \bibinfo{author}{Shen, D.}, \bibinfo{author}{Rekik, I.},
  \bibinfo{year}{2019}.
\newblock \bibinfo{title}{Progressive infant brain connectivity evolution
  prediction from neonatal {MRI} using bidirectionally supervised sample
  selection}.
\newblock \bibinfo{journal}{International Workshop on PRedictive Intelligence
  In MEdicine} , \bibinfo{pages}{63--72}.
%Type = Article
\bibitem[{Goktas et~al.(2020)Goktas, Bessadok and Rekik}]{goktas2020residual}
\bibinfo{author}{Goktas, A.S.}, \bibinfo{author}{Bessadok, A.},
  \bibinfo{author}{Rekik, I.}, \bibinfo{year}{2020}.
\newblock \bibinfo{title}{Residual embedding similarity-based network selection
  for predicting brain network evolution trajectory from a single observation}.
\newblock \bibinfo{journal}{arXiv preprint arXiv:2009.11110} .
%Type = Article
\bibitem[{Goodfellow et~al.(2014)}]{Goodfellow:2014}
\bibinfo{author}{Goodfellow, I.}, et~al., \bibinfo{year}{2014}.
\newblock \bibinfo{title}{Generative adversarial nets}.
\newblock \bibinfo{journal}{Advances in neural information processing systems}
  , \bibinfo{pages}{2672--2680}.
%Type = Article
\bibitem[{Gulrajani et~al.(2017)Gulrajani, Ahmed, Arjovsky, Dumoulin and
  Courville}]{gulrajani:2017}
\bibinfo{author}{Gulrajani, I.}, \bibinfo{author}{Ahmed, F.},
  \bibinfo{author}{Arjovsky, M.}, \bibinfo{author}{Dumoulin, V.},
  \bibinfo{author}{Courville, A.C.}, \bibinfo{year}{2017}.
\newblock \bibinfo{title}{Improved training of wasserstein gans}.
\newblock \bibinfo{journal}{Advances in neural information processing systems}
  , \bibinfo{pages}{5767--5777}.
%Type = Inproceedings
\bibitem[{Gurbuz and Rekik(2020)}]{gurbuz2020deep}
\bibinfo{author}{Gurbuz, M.B.}, \bibinfo{author}{Rekik, I.},
  \bibinfo{year}{2020}.
\newblock \bibinfo{title}{Deep graph normalizer: A geometric deep learning
  approach for estimating connectional brain templates}, in:
  \bibinfo{booktitle}{International Conference on Medical Image Computing and
  Computer-Assisted Intervention}, \bibinfo{organization}{Springer}. pp.
  \bibinfo{pages}{155--165}.
%Type = Article
\bibitem[{Hamilton et~al.(2017)Hamilton, Ying and Leskovec}]{hamilton:2017}
\bibinfo{author}{Hamilton, W.L.}, \bibinfo{author}{Ying, R.},
  \bibinfo{author}{Leskovec, J.}, \bibinfo{year}{2017}.
\newblock \bibinfo{title}{Representation learning on graphs: Methods and
  applications}.
\newblock \bibinfo{journal}{arXiv preprint arXiv:1709.05584} .
%Type = Article
\bibitem[{van~den Heuvel and Sporns(2019)}]{van:2019}
\bibinfo{author}{van~den Heuvel, M.P.}, \bibinfo{author}{Sporns, O.},
  \bibinfo{year}{2019}.
\newblock \bibinfo{title}{A cross-disorder connectome landscape of brain
  dysconnectivity}.
\newblock \bibinfo{journal}{Nature reviews neuroscience} \bibinfo{volume}{20},
  \bibinfo{pages}{435--446}.
%Type = Article
\bibitem[{Huang et~al.(2019a)Huang, Wu, Lin, Pang and Chen}]{Huang:2019a}
\bibinfo{author}{Huang, L.C.}, \bibinfo{author}{Wu, P.A.},
  \bibinfo{author}{Lin, S.Z.}, \bibinfo{author}{Pang, C.Y.},
  \bibinfo{author}{Chen, S.Y.}, \bibinfo{year}{2019}a.
\newblock \bibinfo{title}{Graph theory and network topological metrics may be
  the potential biomarker in {Parkinson’s} disease}.
\newblock \bibinfo{journal}{Journal of Clinical Neuroscience}
  \bibinfo{volume}{68}, \bibinfo{pages}{235--242}.
%Type = Article
\bibitem[{Huang et~al.(2019b)}]{Huang:2019}
\bibinfo{author}{Huang, P.}, et~al., \bibinfo{year}{2019}b.
\newblock \bibinfo{title}{{CoCa-GAN}: Common-feature-learning-based
  context-aware generative adversarial network for glioma grading}.
\newblock \bibinfo{journal}{International Conference on Medical Image Computing
  and Computer-Assisted Intervention} , \bibinfo{pages}{155--163}.
%Type = Article
\bibitem[{Jain(2010)}]{Jain:2010}
\bibinfo{author}{Jain, A.K.}, \bibinfo{year}{2010}.
\newblock \bibinfo{title}{Data clustering: 50 years beyond k-means}.
\newblock \bibinfo{journal}{Pattern recognition letters} \bibinfo{volume}{31},
  \bibinfo{pages}{651--666}.
%Type = Article
\bibitem[{Jolliffe and Cadima(2016)}]{Jolliffe:2016}
\bibinfo{author}{Jolliffe, I.T.}, \bibinfo{author}{Cadima, J.},
  \bibinfo{year}{2016}.
\newblock \bibinfo{title}{Principal component analysis: a review and recent
  developments}.
\newblock \bibinfo{journal}{Philosophical Transactions of the Royal Society A:
  Mathematical, Physical and Engineering Sciences} \bibinfo{volume}{374},
  \bibinfo{pages}{20150202}.
%Type = Article
\bibitem[{Joyce et~al.(2010)Joyce, Laurienti, Burdette and
  Hayasaka}]{Joyce:2010}
\bibinfo{author}{Joyce, K.E.}, \bibinfo{author}{Laurienti, P.J.},
  \bibinfo{author}{Burdette, J.H.}, \bibinfo{author}{Hayasaka, S.},
  \bibinfo{year}{2010}.
\newblock \bibinfo{title}{A new measure of centrality for brain networks}.
\newblock \bibinfo{journal}{PloS one} \bibinfo{volume}{5},
  \bibinfo{pages}{e12200}.
%Type = Article
\bibitem[{Khelifa and Rekik(2019)}]{Khelifa:2019}
\bibinfo{author}{Khelifa, O.B.}, \bibinfo{author}{Rekik, I.},
  \bibinfo{year}{2019}.
\newblock \bibinfo{title}{Graph morphology-based genetic algorithm for
  classifying late dementia states}.
\newblock \bibinfo{journal}{International Workshop on Connectomics in
  Neuroimaging} , \bibinfo{pages}{21--31}.
%Type = Article
\bibitem[{Kipf and Welling(2016)}]{Kipf:2016}
\bibinfo{author}{Kipf, T.N.}, \bibinfo{author}{Welling, M.},
  \bibinfo{year}{2016}.
\newblock \bibinfo{title}{Semi-supervised classification with graph
  convolutional networks}.
\newblock \bibinfo{journal}{arXiv preprint arXiv:1609.02907} .
%Type = Article
\bibitem[{Kullback(1959)}]{Kullback:1959}
\bibinfo{author}{Kullback, S.}, \bibinfo{year}{1959}.
\newblock \bibinfo{title}{Information theory and statistics wiley}.
\newblock \bibinfo{journal}{New York} .
%Type = Article
\bibitem[{Li et~al.(2019)}]{Li:2019}
\bibinfo{author}{Li, H.}, et~al., \bibinfo{year}{2019}.
\newblock \bibinfo{title}{Diamondgan: Unified multi-modal generative
  adversarial networks for {MRI} sequences synthesis}.
\newblock \bibinfo{journal}{International Conference on Medical Image Computing
  and Computer-Assisted Intervention} , \bibinfo{pages}{795--803}.
%Type = Article
\bibitem[{Li et~al.(2013)Li, Andreasen, Nopoulos and Magnotta}]{Li:2013}
\bibinfo{author}{Li, W.}, \bibinfo{author}{Andreasen, N.C.},
  \bibinfo{author}{Nopoulos, P.}, \bibinfo{author}{Magnotta, V.A.},
  \bibinfo{year}{2013}.
\newblock \bibinfo{title}{Automated parcellation of the brain surface generated
  from magnetic resonance images}.
\newblock \bibinfo{journal}{Frontiers in neuroinformatics} \bibinfo{volume}{7},
  \bibinfo{pages}{23}.
%Type = Inproceedings
\bibitem[{Li et~al.(2020)Li, Zhou, Dvornek, Zhang, Zhuang, Ventola and
  Duncan}]{li2020pooling}
\bibinfo{author}{Li, X.}, \bibinfo{author}{Zhou, Y.}, \bibinfo{author}{Dvornek,
  N.C.}, \bibinfo{author}{Zhang, M.}, \bibinfo{author}{Zhuang, J.},
  \bibinfo{author}{Ventola, P.}, \bibinfo{author}{Duncan, J.S.},
  \bibinfo{year}{2020}.
\newblock \bibinfo{title}{Pooling regularized graph neural network for fmri
  biomarker analysis}, in: \bibinfo{booktitle}{International Conference on
  Medical Image Computing and Computer-Assisted Intervention},
  \bibinfo{organization}{Springer}. pp. \bibinfo{pages}{625--635}.
%Type = Article
\bibitem[{Liao et~al.(2019)}]{Liao:2019}
\bibinfo{author}{Liao, R.}, et~al., \bibinfo{year}{2019}.
\newblock \bibinfo{title}{Efficient graph generation with graph recurrent
  attention networks}.
\newblock \bibinfo{journal}{Advances in Neural Information Processing Systems}
  , \bibinfo{pages}{4257--4267}.
%Type = Article
\bibitem[{Lin et~al.(1990)Lin, Sellke and Coyle}]{Lin:1990}
\bibinfo{author}{Lin, J.H.}, \bibinfo{author}{Sellke, T.M.},
  \bibinfo{author}{Coyle, E.J.}, \bibinfo{year}{1990}.
\newblock \bibinfo{title}{Adaptive stack filtering under the mean absolute
  error criterion}.
\newblock \bibinfo{journal}{IEEE transactions on acoustics, speech, and signal
  processing} \bibinfo{volume}{38}, \bibinfo{pages}{938--954}.
%Type = Article
\bibitem[{Liu et~al.(2017)}]{Liu:2017}
\bibinfo{author}{Liu, J.}, et~al., \bibinfo{year}{2017}.
\newblock \bibinfo{title}{Complex brain network analysis and its applications
  to brain disorders: a survey}.
\newblock \bibinfo{journal}{Complexity} \bibinfo{volume}{2017}.
%Type = Article
\bibitem[{Maaten and Hinton(2008)}]{Maaten:2008}
\bibinfo{author}{Maaten, L.v.d.}, \bibinfo{author}{Hinton, G.},
  \bibinfo{year}{2008}.
\newblock \bibinfo{title}{Visualizing data using {t-SNE}}.
\newblock \bibinfo{journal}{Journal of machine learning research}
  \bibinfo{volume}{9}, \bibinfo{pages}{2579--2605}.
%Type = Article
\bibitem[{Mahjoub et~al.(2018)Mahjoub, Mahjoub and Rekik}]{Mahjoub:2018}
\bibinfo{author}{Mahjoub, I.}, \bibinfo{author}{Mahjoub, M.A.},
  \bibinfo{author}{Rekik, I.}, \bibinfo{year}{2018}.
\newblock \bibinfo{title}{Brain multiplexes reveal morphological connectional
  biomarkers fingerprinting late brain dementia states}.
\newblock \bibinfo{journal}{Scientific reports} \bibinfo{volume}{8},
  \bibinfo{pages}{4103}.
%Type = Article
\bibitem[{Mhiri et~al.(2020a)Mhiri, Khalifa, Mahjoub and Rekik}]{Mhiri:2020b}
\bibinfo{author}{Mhiri, I.}, \bibinfo{author}{Khalifa, A.B.},
  \bibinfo{author}{Mahjoub, M.A.}, \bibinfo{author}{Rekik, I.},
  \bibinfo{year}{2020}a.
\newblock \bibinfo{title}{Brain graph super-resolution for boosting
  neurological disorder diagnosis using unsupervised multi-topology
  connectional brain template learning}.
\newblock \bibinfo{journal}{Medical Image Analysis} , \bibinfo{pages}{101768}.
%Type = Inproceedings
\bibitem[{Mhiri et~al.(2020b)Mhiri, Mahjoub and Rekik}]{mhiri2020supervised}
\bibinfo{author}{Mhiri, I.}, \bibinfo{author}{Mahjoub, M.A.},
  \bibinfo{author}{Rekik, I.}, \bibinfo{year}{2020}b.
\newblock \bibinfo{title}{Supervised multi-topology network cross-diffusion for
  population-driven brain network atlas estimation}, in:
  \bibinfo{booktitle}{International Conference on Medical Image Computing and
  Computer-Assisted Intervention}, \bibinfo{organization}{Springer}. pp.
  \bibinfo{pages}{166--176}.
%Type = Article
\bibitem[{Mhiri and Rekik(2020)}]{Mhiri:2020}
\bibinfo{author}{Mhiri, I.}, \bibinfo{author}{Rekik, I.}, \bibinfo{year}{2020}.
\newblock \bibinfo{title}{Joint functional brain network atlas estimation and
  feature selection for neurological disorder diagnosis with application to
  autism}.
\newblock \bibinfo{journal}{Medical image analysis} \bibinfo{volume}{60},
  \bibinfo{pages}{101596}.
%Type = Article
\bibitem[{Mitton et~al.()Mitton, Senn, Wynne and Murray-Smith}]{mittongraph}
\bibinfo{author}{Mitton, J.}, \bibinfo{author}{Senn, H.M.},
  \bibinfo{author}{Wynne, K.}, \bibinfo{author}{Murray-Smith, R.}, .
\newblock \bibinfo{title}{A graph vae and graph transformer approach to
  generating molecular graphs} .
%Type = Inproceedings
\bibitem[{Nebli et~al.(2020)Nebli, Kaplan and Rekik}]{nebli2020deep}
\bibinfo{author}{Nebli, A.}, \bibinfo{author}{Kaplan, U.A.},
  \bibinfo{author}{Rekik, I.}, \bibinfo{year}{2020}.
\newblock \bibinfo{title}{Deep evographnet architecture for time-dependent
  brain graph data synthesis from a single timepoint}, in:
  \bibinfo{booktitle}{International Workshop on PRedictive Intelligence In
  MEdicine}, \bibinfo{organization}{Springer}. pp. \bibinfo{pages}{144--155}.
%Type = Article
\bibitem[{Nebli and Rekik(2019)}]{Nebli:2019}
\bibinfo{author}{Nebli, A.}, \bibinfo{author}{Rekik, I.}, \bibinfo{year}{2019}.
\newblock \bibinfo{title}{Gender differences in cortical morphological
  networks}.
\newblock \bibinfo{journal}{Brain Imaging and Behavior} ,
  \bibinfo{pages}{1--9}.
%Type = Article
\bibitem[{Neyshabur et~al.(2017)Neyshabur, Bhojanapalli and
  Chakrabarti}]{Neyshabur:2017}
\bibinfo{author}{Neyshabur, B.}, \bibinfo{author}{Bhojanapalli, S.},
  \bibinfo{author}{Chakrabarti, A.}, \bibinfo{year}{2017}.
\newblock \bibinfo{title}{Stabilizing {GAN} training with multiple random
  projections}.
\newblock \bibinfo{journal}{arXiv preprint arXiv:1705.07831} .
%Type = Article
\bibitem[{Pan et~al.(2018)Pan, Hu, Long, Jiang, Yao and Zhang}]{Pan:2018}
\bibinfo{author}{Pan, S.}, \bibinfo{author}{Hu, R.}, \bibinfo{author}{Long,
  G.}, \bibinfo{author}{Jiang, J.}, \bibinfo{author}{Yao, L.},
  \bibinfo{author}{Zhang, C.}, \bibinfo{year}{2018}.
\newblock \bibinfo{title}{Adversarially regularized graph autoencoder for graph
  embedding}.
\newblock \bibinfo{journal}{arXiv preprint arXiv:1802.04407} .
%Type = Article
\bibitem[{Pan et~al.(2019)Pan, Liu, Lian, Xia and Shen}]{Pan:2019}
\bibinfo{author}{Pan, Y.}, \bibinfo{author}{Liu, M.}, \bibinfo{author}{Lian,
  C.}, \bibinfo{author}{Xia, Y.}, \bibinfo{author}{Shen, D.},
  \bibinfo{year}{2019}.
\newblock \bibinfo{title}{Disease-image specific generative adversarial network
  for brain disease diagnosis with incomplete multi-modal neuroimages}.
\newblock \bibinfo{journal}{International Conference on Medical Image Computing
  and Computer-Assisted Intervention} , \bibinfo{pages}{137--145}.
%Type = Article
\bibitem[{Rice et~al.(2020)Rice, Wong and Kolter}]{rice2020overfitting}
\bibinfo{author}{Rice, L.}, \bibinfo{author}{Wong, E.},
  \bibinfo{author}{Kolter, J.Z.}, \bibinfo{year}{2020}.
\newblock \bibinfo{title}{Overfitting in adversarially robust deep learning}.
\newblock \bibinfo{journal}{arXiv preprint arXiv:2002.11569} .
%Type = Article
\bibitem[{Saram{\"a}ki et~al.(2007)Saram{\"a}ki, Kivel{\"a}, Onnela, Kaski and
  Kertesz}]{Saramaki:2007}
\bibinfo{author}{Saram{\"a}ki, J.}, \bibinfo{author}{Kivel{\"a}, M.},
  \bibinfo{author}{Onnela, J.P.}, \bibinfo{author}{Kaski, K.},
  \bibinfo{author}{Kertesz, J.}, \bibinfo{year}{2007}.
\newblock \bibinfo{title}{Generalizations of the clustering coefficient to
  weighted complex networks}.
\newblock \bibinfo{journal}{Physical Review E} \bibinfo{volume}{75},
  \bibinfo{pages}{027105}.
%Type = Inproceedings
\bibitem[{Song et~al.(2020)Song, Frangi, Xiao, Cao, Wang and
  Lei}]{song2020integrating}
\bibinfo{author}{Song, X.}, \bibinfo{author}{Frangi, A.},
  \bibinfo{author}{Xiao, X.}, \bibinfo{author}{Cao, J.}, \bibinfo{author}{Wang,
  T.}, \bibinfo{author}{Lei, B.}, \bibinfo{year}{2020}.
\newblock \bibinfo{title}{Integrating similarity awareness and adaptive
  calibration in graph convolution network to predict disease}, in:
  \bibinfo{booktitle}{International Conference on Medical Image Computing and
  Computer-Assisted Intervention}, \bibinfo{organization}{Springer}. pp.
  \bibinfo{pages}{124--133}.
%Type = Article
\bibitem[{Soussia and Rekik(2018)}]{Soussia:2018b}
\bibinfo{author}{Soussia, M.}, \bibinfo{author}{Rekik, I.},
  \bibinfo{year}{2018}.
\newblock \bibinfo{title}{Unsupervised manifold learning using high-order
  morphological brain networks derived from {T1-w MRI} for autism diagnosis}.
\newblock \bibinfo{journal}{Frontiers in Neuroinformatics}
  \bibinfo{volume}{12}.
%Type = Article
\bibitem[{Sserwadda and Rekik(2020)}]{sserwadda2020}
\bibinfo{author}{Sserwadda, A.}, \bibinfo{author}{Rekik, I.},
  \bibinfo{year}{2020}.
\newblock \bibinfo{title}{Topology-guided cyclic brain connectivity generation
  using geometric deep learning}.
\newblock \bibinfo{journal}{Journal of Neuroscience Methods} ,
  \bibinfo{pages}{108988}.
%Type = Article
\bibitem[{Su et~al.(2019)Su, Hajimirsadeghi and Mori}]{Su:2019}
\bibinfo{author}{Su, S.Y.}, \bibinfo{author}{Hajimirsadeghi, H.},
  \bibinfo{author}{Mori, G.}, \bibinfo{year}{2019}.
\newblock \bibinfo{title}{Graph generation with {Variational Recurrent Neural
  Network}}.
\newblock \bibinfo{journal}{arXiv preprint arXiv:1910.01743} .
%Type = Article
\bibitem[{Tiao et~al.()Tiao, Elinas, Nguyen and Bonilla}]{Tiao}
\bibinfo{author}{Tiao, L.}, \bibinfo{author}{Elinas, P.},
  \bibinfo{author}{Nguyen, H.}, \bibinfo{author}{Bonilla, E.V.}, .
\newblock \bibinfo{title}{{Variational Graph Convolutional Networks}} .
%Type = Article
\bibitem[{Veli{\v{c}}kovi{\'c} et~al.(2017)Veli{\v{c}}kovi{\'c}, Cucurull,
  Casanova, Romero, Lio and Bengio}]{Velivckovic:2017}
\bibinfo{author}{Veli{\v{c}}kovi{\'c}, P.}, \bibinfo{author}{Cucurull, G.},
  \bibinfo{author}{Casanova, A.}, \bibinfo{author}{Romero, A.},
  \bibinfo{author}{Lio, P.}, \bibinfo{author}{Bengio, Y.},
  \bibinfo{year}{2017}.
\newblock \bibinfo{title}{Graph attention networks}.
\newblock \bibinfo{journal}{arXiv preprint arXiv:1710.10903} .
%Type = Article
\bibitem[{Vohryzek et~al.(2020)}]{Vohryzek:2020}
\bibinfo{author}{Vohryzek, J.}, et~al., \bibinfo{year}{2020}.
\newblock \bibinfo{title}{Dynamic spatiotemporal patterns of brain connectivity
  reorganize across development}.
\newblock \bibinfo{journal}{Network neuroscience} \bibinfo{volume}{4},
  \bibinfo{pages}{115--133}.
%Type = Article
\bibitem[{Wang et~al.(2017)Wang, Ramazzotti, De~Sano, Zhu, Pierson and
  Batzoglou}]{Wang:2017}
\bibinfo{author}{Wang, B.}, \bibinfo{author}{Ramazzotti, D.},
  \bibinfo{author}{De~Sano, L.}, \bibinfo{author}{Zhu, J.},
  \bibinfo{author}{Pierson, E.}, \bibinfo{author}{Batzoglou, S.},
  \bibinfo{year}{2017}.
\newblock \bibinfo{title}{{SIMLR}: a tool for large-scale single-cell analysis
  by multi-kernel learning}.
\newblock \bibinfo{journal}{bioRxiv} , \bibinfo{pages}{118901}.
%Type = Article
\bibitem[{Wang et~al.(2020)Wang, Chen, Lin, Sigal and
  de~Silva}]{wang2020discriminative}
\bibinfo{author}{Wang, J.}, \bibinfo{author}{Chen, J.}, \bibinfo{author}{Lin,
  J.}, \bibinfo{author}{Sigal, L.}, \bibinfo{author}{de~Silva, C.W.},
  \bibinfo{year}{2020}.
\newblock \bibinfo{title}{Discriminative feature alignment:
  Improvingtransferability of unsupervised domainadaptation by gaussian-guided
  latentalignment}.
\newblock \bibinfo{journal}{arXiv preprint arXiv:2006.12770} .
%Type = Article
\bibitem[{Wen et~al.(2017)}]{Wen:2017}
\bibinfo{author}{Wen, H.}, et~al., \bibinfo{year}{2017}.
\newblock \bibinfo{title}{Disrupted topological organization of structural
  networks revealed by probabilistic diffusion tractography in tourette
  syndrome children}.
\newblock \bibinfo{journal}{Human Brain Mapping} \bibinfo{volume}{38},
  \bibinfo{pages}{3988--4008}.
%Type = Article
\bibitem[{Wu et~al.(2019)Wu, Lin, Chang, Chang and Liao}]{Wu:2019}
\bibinfo{author}{Wu, P.W.}, \bibinfo{author}{Lin, Y.J.},
  \bibinfo{author}{Chang, C.H.}, \bibinfo{author}{Chang, E.Y.},
  \bibinfo{author}{Liao, S.W.}, \bibinfo{year}{2019}.
\newblock \bibinfo{title}{Relgan: Multi-domain image-to-image translation via
  relative attributes}.
\newblock \bibinfo{journal}{Proceedings of the IEEE International Conference on
  Computer Vision} , \bibinfo{pages}{5914--5922}.
%Type = Article
\bibitem[{Yang et~al.(2018)Yang, Lu, Lee, Batra and Parikh}]{yang:2018}
\bibinfo{author}{Yang, J.}, \bibinfo{author}{Lu, J.}, \bibinfo{author}{Lee,
  S.}, \bibinfo{author}{Batra, D.}, \bibinfo{author}{Parikh, D.},
  \bibinfo{year}{2018}.
\newblock \bibinfo{title}{Graph r-cnn for scene graph generation}.
\newblock \bibinfo{journal}{Proceedings of the European conference on computer
  vision (ECCV)} , \bibinfo{pages}{670--685}.
%Type = Inproceedings
\bibitem[{Yang et~al.(2020)Yang, Zhu, Zhang, Huang and Zhang}]{yang2020unified}
\bibinfo{author}{Yang, J.}, \bibinfo{author}{Zhu, Q.}, \bibinfo{author}{Zhang,
  R.}, \bibinfo{author}{Huang, J.}, \bibinfo{author}{Zhang, D.},
  \bibinfo{year}{2020}.
\newblock \bibinfo{title}{Unified brain network with functional and structural
  data}, in: \bibinfo{booktitle}{International Conference on Medical Image
  Computing and Computer-Assisted Intervention},
  \bibinfo{organization}{Springer}. pp. \bibinfo{pages}{114--123}.
%Type = Article
\bibitem[{You et~al.(2018)You, Liu, Ying, Pande and Leskovec}]{you:2018}
\bibinfo{author}{You, J.}, \bibinfo{author}{Liu, B.}, \bibinfo{author}{Ying,
  Z.}, \bibinfo{author}{Pande, V.}, \bibinfo{author}{Leskovec, J.},
  \bibinfo{year}{2018}.
\newblock \bibinfo{title}{Graph convolutional policy network for goal-directed
  molecular graph generation}.
\newblock \bibinfo{journal}{Advances in neural information processing systems}
  , \bibinfo{pages}{6410--6421}.
%Type = Article
\bibitem[{Zeng and Zheng(2019)}]{Zeng:2019}
\bibinfo{author}{Zeng, G.}, \bibinfo{author}{Zheng, G.}, \bibinfo{year}{2019}.
\newblock \bibinfo{title}{Hybrid generative adversarial networks for deep {MR}
  to {CT} synthesis using unpaired data}.
\newblock \bibinfo{journal}{International Conference on Medical Image Computing
  and Computer-Assisted Intervention} , \bibinfo{pages}{759--767}.
%Type = Inproceedings
\bibitem[{Zhang et~al.(2020a)Zhang, Wang and Zhu}]{zhang2020}
\bibinfo{author}{Zhang, L.}, \bibinfo{author}{Wang, L.}, \bibinfo{author}{Zhu,
  D.}, \bibinfo{year}{2020}a.
\newblock \bibinfo{title}{Recovering brain structural connectivity from
  functional connectivity via multi-gcn based generative adversarial network},
  in: \bibinfo{booktitle}{International Conference on Medical Image Computing
  and Computer-Assisted Intervention}, \bibinfo{organization}{Springer}. pp.
  \bibinfo{pages}{53--61}.
%Type = Article
\bibitem[{Zhang et~al.(2020b)Zhang, Wang, Liu and Zhang}]{Zhang:2020}
\bibinfo{author}{Zhang, L.}, \bibinfo{author}{Wang, M.}, \bibinfo{author}{Liu,
  M.}, \bibinfo{author}{Zhang, D.}, \bibinfo{year}{2020}b.
\newblock \bibinfo{title}{A survey on deep learning for neuroimaging-based
  brain disorder analysis}.
\newblock \bibinfo{journal}{arXiv preprint arXiv:2005.04573} .
%Type = Article
\bibitem[{Zhang et~al.(2019)Zhang, Tong, Xu and Maciejewski}]{Zhang:2019}
\bibinfo{author}{Zhang, S.}, \bibinfo{author}{Tong, H.}, \bibinfo{author}{Xu,
  J.}, \bibinfo{author}{Maciejewski, R.}, \bibinfo{year}{2019}.
\newblock \bibinfo{title}{Graph convolutional networks: a comprehensive
  review}.
\newblock \bibinfo{journal}{Computational Social Networks} \bibinfo{volume}{6},
  \bibinfo{pages}{11}.
%Type = Article
\bibitem[{Zhang et~al.(2020c)Zhang, Cui and Zhu}]{Zhang:2020a}
\bibinfo{author}{Zhang, Z.}, \bibinfo{author}{Cui, P.}, \bibinfo{author}{Zhu,
  W.}, \bibinfo{year}{2020}c.
\newblock \bibinfo{title}{Deep learning on graphs: A survey}.
\newblock \bibinfo{journal}{IEEE Transactions on Knowledge and Data
  Engineering} .
%Type = Article
\bibitem[{Zhou et~al.(2018)}]{Zhou:2018}
\bibinfo{author}{Zhou, J.}, et~al., \bibinfo{year}{2018}.
\newblock \bibinfo{title}{Graph neural networks: A review of methods and
  applications}.
\newblock \bibinfo{journal}{arXiv preprint arXiv:1812.08434} .
%Type = Article
\bibitem[{Zhu et~al.(2017)Zhu, Park, Isola and Efros}]{Zhu:2017}
\bibinfo{author}{Zhu, J.Y.}, \bibinfo{author}{Park, T.},
  \bibinfo{author}{Isola, P.}, \bibinfo{author}{Efros, A.A.},
  \bibinfo{year}{2017}.
\newblock \bibinfo{title}{Unpaired image-to-image translation using
  cycle-consistent adversarial networks}.
\newblock \bibinfo{journal}{Proceedings of the IEEE international conference on
  computer vision} , \bibinfo{pages}{2223--2232}.

\end{thebibliography}
\bibliographystyle{model2-names}

\end{document}